\documentclass[a4paper,fleqn]{cas-sc}

\usepackage{subcaption}
\usepackage{amsthm} 
\usepackage{siunitx}

\usepackage[finalnew]{trackchanges}  % inline finalnew

\usepackage[authoryear]{natbib}
\def\tsc#1{\csdef{#1}{\textsc{\lowercase{#1}}\xspace}}
\tsc{WGM}
\tsc{QE}
\usepackage{lineno}
\renewcommand{\theequation}{Eq.~\arabic{equation}}

\usepackage{setspace}
\usepackage{tabularx}
\usepackage{booktabs}
\usepackage{multirow}
\usepackage{rotating}   % For sideways table

\begin{document}
\let\WriteBookmarks\relax
\def\floatpagepagefraction{1}
\def\textpagefraction{.001}

% For editing/track changes
\addeditor{YJ}

\shorttitle{Definition and validation of 3D ocean clusters}    

\shortauthors{Jenniges et al.}  

\title [mode = title]{Unveiling 3D Ocean Biogeochemical Provinces in the North Atlantic: A Systematic Comparison and Validation of Clustering Methods}
%Unveiling 3D Ocean Biogeochemical Provinces in the North Atlantic: A Systematic Comparison of Objective Clustering Methods
%Unveiling 3D Ocean Biogeochemical Provinces: A Machine Learning Approach for Systematic Clustering and Validation}  




\author[1, 2]{Yvonne Jenniges}[orcid=0009-0007-7933-8002]

\cormark[1]


\ead{yvonne.jenniges@awi.de}


\credit{Conceptualization, data curation, funding acquisition, investigation, methodology, project administration, software, visualization, writing – original draft}

\author[3, 4, 5]{Maike Sonnewald}[orcid=0000-0002-5136-9604]
%\ead{sonnewald@ucdavis.edu}
\credit{Conceptualization, methodology, supervision, writing – review and editing}

\author[2]{Sebastian Maneth}[orcid=0000-0001-8667-5436]
%\fnmark[3]
\cormark[1]
\ead{maneth@uni-bremen.de}
\credit{Conceptualization, data curation, funding acquisition, methodology, supervision, writing – review and editing}

\author[6]{Are Olsen}[orcid=0000-0003-1696-9142]
%\fnmark[4]
%\ead{are.olsen@uib.no}
\credit{Data curation, writing – review and editing}

\author[1, 7]{Boris P. Koch}[orcid=0000-0002-8453-731X]
%\fnmark[5]
%\ead{boris.koch@awi.de}
\credit{Conceptualization, funding acquisition, methodology, supervision, visualization, writing – review and editing}

% Affiliations
\affiliation[1]{organization={Ecological Chemistry, Alfred-Wegener Institut Helmholtz-Zentrum f\"ur Polar- und Meeresforschung},
            %addressline={Am Handelshafen 12}, 
            city={27570 Bremerhaven},
            %citysep={}, 
            %postcode={}, 
            %state={Bremen},
            country={Germany}}
            
\affiliation[2]{organization={Faculty of Informatics, University of Bremen},
            %addressline={Bibliothekstraße 5},
            city={28539 Bremen},
            %citysep={},
            %postcode={},  
            %state={Bremen},
            country={Germany}}
            
\affiliation[3]{organization={Department of Computer Science, University of California},
            %addressline={One Shields Avenue}, 
            city={Davis},
            %citysep={}, 
            %postcode={95616}, 
            state={CA},
            country={USA}}

\affiliation[4]{organization={University of Washington},
            %addressline={}, 
            city={Seattle},
            %citysep={}, 
            %postcode={}, 
            state={WA},
            country={USA}}
            
\affiliation[5]{organization={NOAA/Geophysical Fluid Dynamics Laboratory},
            %addressline={}, 
            city={Princeton},
            %citysep={}, 
            %postcode={}, 
            state={NJ},
            country={USA}}

\affiliation[6]{organization={Geophysical Institute, University of Bergen},
            %addressline={Allégaten 70}, 
            city={5020 Bergen},
            %citysep={},
            %postcode={}, 
            country={Norway}}

\affiliation[7]{organization={Department of Technology, University of Applied Sciences},
            %addressline={An der Karlstadt 8}, 
            city={27568 Bremerhaven},
            %citysep={},
            %postcode={}, 
            country={Germany}}

\nonumnote{}

\begin{abstract}
Defining ocean regions and water masses helps to understand marine processes and can serve downstream tasks such as defining marine protected areas. However, such definitions often result from subjective decisions potentially producing misleading, unreproducible outcomes. Here, the aim was to objectively define regions of the North Atlantic through systematic comparison of clustering methods within the Native Emergent Manifold Interrogation (NEMI) framework (Sonnewald, 2023).
About 300 million measured salinity, temperature, and oxygen, nitrate, phosphate and silicate concentration values served as input for various clustering methods (k-Means, agglomerative Ward, and Density-Based Spatial Clustering of Applications with Noise (DBSCAN)). Uniform Manifold Approximation and Projection (UMAP) emphasised (dis-)similarities in the data while reducing dimensionality. Based on systematic validation of clustering methods and their hyperparameters using internal, external and relative validation techniques, results showed that UMAP-DBSCAN best represented the data. Strikingly, internal validation metrics proved systematically unreliable for comparing clustering methods.
To address stochastic variability, 100 UMAP-DBSCAN clustering runs were conducted and aggregated following NEMI, yielding a final set of 321 clusters. Reproducibility was evaluated via ensemble overlap (88.81±1.8\%) and mean grid cell-wise uncertainty (15.49±20\%). Case studies of the Mediterranean Sea, deep Atlantic waters and Labrador Sea showed strong agreement with common water mass definitions.
This study revealed a more detailed regionalisation compared to previous concepts such as the Longhurst provinces through systematic clustering method comparison. The applied method is objective, efficient and reproducible and will support future research on biogeochemical differences and changes in oceanic regions.
\end{abstract}


% Research highlights (3-5)
% \begin{highlights}
% \item Objective, reproducible 3D clustering of North Atlantic physics and biogeochemistry
% \item DBSCAN on embedded data outperformed k-Means and agglomerative Ward clustering
% \item Nonlinear dimensionality reduction substantially improved clustering performance
% \item Grid-cells were assigned to clusters with a mean uncertainty of about \SI{15}{\percent}
% \item Good agreement with previous ecological and biogeochemical subdivisions
% \end{highlights}

\begin{highlights}
\item NEMI enabled objective, reproducible clustering of physics and biogeochemistry
\item DBSCAN best captured natural associations within North Atlantic data
\item UMAP embedding strengthened data associations, improving clustering of all methods
%\item Cluster validity indices proved unreliable for comparing different clustering methods
\item Grid-cells were assigned to clusters with a mean uncertainty of about \SI{15}{\percent}
\item Good agreement with known subdivisions with potential for novel insights
\end{highlights}

% Keywords
% Each keyword is separated by \sep

\maketitle

\doublespacing

% Introduction ----------------------------------------------
\section{Introduction}
The definition of ocean regions \change{is fundamental for advancing}{has offered fundamental advances in} our understanding of marine (eco)systems, biodiversity distributions and their variability. Since the 19th century, efforts to delineate biogeographic patterns have evolved from early taxonomic classifications \citep{forbes1856} to more comprehensive ecological and physical ocean regionalisations \citep{ekman1935, hedgpeth1957, briggs1974, hayden1984, briggs1995, bailey1998}. 
One of the most influential partitioning schemes, the ecological provinces by \cite{longhurst2007}, has provided a foundational framework for investigating large-scale oceanographic patterns and processes, hotspots of biodiversity and ecological relationships. For example, \change{they}{the Longhurst regimes} helped quantify primary production \citep{longhurst1995}, characterise tuna movements \citep{logan2020} and trophic dynamics and food web structure \citep{arnoldi2023}.

Ocean regionalisations have been used in studies addressing fields such as biogeographic realms \citep{costello2017}, carbon flux \citep{gloege2017} or patterns of marine viruses \citep{brum2015}. 
They also serve economic decisions in fisheries \citep{jorda2022} and policy such as the designation of marine protected areas legislation \citep{spalding2007, sonnewald2020, zhao2020a, reisinger2022}.

While previous approaches often rely on predefined thresholds and subjective criteria, emerging data-driven techniques, particularly clustering algorithms, provide a more objective framework for ocean region delineation \citep{devred2007, hardman2008, oliver2008, kavanaugh2014, sayre2017, sonnewald2019, sonnewald2020}. 
\add{The Native Emergent Manifold Interrogation (NEMI) method } \citep{sonnewald2023}
\add{ addresses key challenges in analysing complex geophysical data by integrating manifold learning, clustering, stochastic ensemble methods, uncertainty quantification and intuitive validation to ensure robust and interpretable clustering outcomes. This study builds upon and enhances the NEMI method through systematic comparison of clustering algorithms applied to biogeochemical data.}
\add{Fundamentally,} clustering results are influenced by algorithmic biases \citep{thrun2021} and, for some algorithms, the possibility of variable outcomes raising concerns about suitability, comparability and reproducibility. Two main validation strategies exist \citep{rui2005, ullmann2022}: 
(i) Internal validation, which assesses cohesion within clusters and separation between clusters using indices, such as the Calinski-Harabasz index \citep{calinski1974}, Davies-Bouldin index \citep{davies1979} and silhouette score \citep{rousseeuw1987} and 
(ii) external validation, which utilises knowledge not seen during model training, like comparing clustering results to established ecological classifications or visual analysis in different data spaces. For example, dimensionality reduction techniques like Uniform Manifold Approximation and Projection (UMAP, \cite{mcinnes2018}) \change{enhance}{improve} the interpretability of clustering outcomes through\add{ enhancing associations between data structures} and visualisation \citep{allaoui2020}.

A key challenge for clustering is quantifying uncertainty. Some clustering and dimensionality reduction algorithms are non-deterministic and a globally optimal solution is not guaranteed. For example, UMAP uses stochasticity \citep{mcinnes2018}, Density-Based Spatial Clustering of Applications with Noise (DBSCAN) results may change with permuted input data \citep{schubert2017} and k-Means is sensitive to cluster centroid initialisation \citep{franti2019}. Quantifying clustering variability is crucial to assess reproducibility and reliability. One  approach involves multiple runs and single-number similarity metrics, such as overlap \citep{manning2008}. The NEMI method, which combines multiple clustering runs to form a final cluster set, has been proposed as a novel solution to represent statistical variability and quantify uncertainty.

Most prior studies on ocean partitioning focus on surface waters (e.g. \cite{devred2007, longhurst2007, hardman2008, oliver2008, vichi2011, reygondeau2013, fay2014, kavanaugh2014,  reygondeau2020, sonnewald2020}) despite the fact that critical processes such as particle export, upwelling or deep-water formation extend into deeper layers. Also, vertical mixing affects biodiversity leading to depth-variable community trends \citep{delong2006, hoerstmann2022}. Recent efforts to develop 3D ocean regionalisations have incorporated clustering approaches such as k-Means \citep{sayre2017} and hybrid methods combining k-Means, CMeans, agglomerative Ward and agglomerative full linkage \citep{reygondeau2017}. 
In physical oceanography, the definition of 3D water masses, mainly defined by temperature and salinity but also e.g. oxygen, has always played a central role (e.g. \cite{emery2001, tomczak2003}). More recent definitions of water masses are based either on regional \citep{liu2021} or model data \citep{zika2021}.  

This study aims to develop an objective and reproducible 3D regionalisation of the North Atlantic ocean and its marginal seas using a data-driven clustering approach. Geographic coordinates and depth were excluded from the clustering to ensure that the regions are purely based on water properties.
\add{ The analysis is based on a mostly post-industrialisation time-aggregate representing a long-term observational baseline, maximising spatial data coverage.}
A key novelty of this work is the systematic definition and evaluation of a marine clustering using both internal and external validation criteria, bridging statistical rigor with oceanographic knowledge. Furthermore, we increase statistical representativeness by combining multiple clustering runs \add{within the NEMI framework} that also allows for quantification of uncertainty. To contextualize the results, clustering outputs are compared to established definitions of ecological provinces and ocean regions \citep{longhurst2007, sayre2017} in three distinct areas: the deep Atlantic, the Mediterranean, and the Labrador Sea. This work specifically addressed the following research questions:
(i) \change{Which clustering method is most appropriate for the given data?}{Which clustering method works best within the NEMI framework for the given oceanographic data?}
(ii) How reproducible are the clustering results? 
(iii) How do the results compare to existing definitions of ecological provinces and ocean regions?

The insights gained from this study have broad potential for ecological and environmental research. A thoroughly validated, data-driven ocean partitioning may refine or challenge existing frameworks, influencing our understanding of oceanic systems, climate dynamics, and marine resource management. By integrating ecological informatics methodologies, we contribute to the advancement of reproducible and objective ocean classifications, ultimately supporting more robust marine ecosystem assessments and policy applications.

The final gridded 3D set of clusters of the North Atlantic Ocean is publicly available (\url{https://doi.org/10.5281/zenodo.15201767}). It also contains auxiliary information, like the gridded oceanographic parameters. 
\add{For interactive exploration of the clusters, a dashboard is available }(\url{https://doi.org/10.5281/zenodo.16742244}).

\section{Material and Methods}
% Opening blurb -------------------------------------
\subsection{Native Emergent Manifold Interrogation (NEMI) Framework}
\add{
This study implements the Native Emergent Manifold Interrogation (NEMI) framework, developed by }
\cite{sonnewald2023}\add{, as the methodological foundation for objective ocean regionalisation. NEMI integrates manifold learning, systematic clustering comparison, ensemble methods, and uncertainty quantification to ensure robust clustering outcomes. Specifically, we use NEMI to 
1) aggregate results from multiple stochastic clustering runs (e.g. here UMAP-DBSCAN), reducing variability to yield robust final clusters, 
2) quantify uncertainty by combining ensemble runs, where NEMI directly quantifies the uncertainty of cluster assignments, enhancing reproducibility and reliability, and 
3) utilising a systematic framework for validation through structured approach to integrate external and relative validation strategies.
The following sections detail our implementation of the NEMI protocol and systematic algorithm comparison in the context of the COMFORT dataset, described below.
}

% Data ----------------------------------------------
\subsection{Data}
The dataset for this study \citep{comfortDataset2022} was assembled in the framework of the EU project “Our Common Future Ocean in the Earth System” (COMFORT) and combines ten observational datasets, including the World Ocean Database 2018 (WOD18) and Argo floats (for more information refer to \cite{comfortReport2021}). It contains data on 47 parameters measured globally from the year 1772 to 2020 amounting to 458,724,734 values.

The focus area was the North Atlantic from \SI{-77}{} to \SI{30}{\degree} longitude, from 0 to \SI{70}{\degree} latitude and from 0 to \SI{5000}{\meter} depth. 
From the COMFORT dataset, the parameters temperature, salinity, as well as oxygen, nitrate, silicate and phosphate concentration were selected, which had comparatively good spatial coverage. 
After quality filtering, unit conversions and averaging over all times (years 1772 -- 2020),  the data was mapped on a grid with a spatial resolution of \SI{1}{\degree} and 12 water depth intervals:
0 -- \SI{50}{\meter}, 
50 -- \SI{100}{\meter}, 
100 -- \SI{200}{\meter}, 
200 -- \SI{300}{\meter}, 
300 -- \SI{400}{\meter}, 
400 -- \SI{500}{\meter}, 
500 -- \SI{1000}{\meter}, 
1000 -- \SI{1500}{\meter}, 
1500 -- \SI{2000}{\meter}, 
2000 -- \SI{3000}{\meter}, 
3000 -- \SI{4000}{\meter}, 
4000 -- \SI{5000}{\meter}. 
Missing values were imputed using the K-Nearest Neighbours (KNN) algorithm (Python library scikit-learn, version 1.5; \cite{pedregosa2011}). Details on data preparation and parameter distributions can be found in Supplementary Material A.

% Methods ----------------------------------------------
\subsection{Clustering}
\label{sec:methods:clustering}
\begin{figure}[ht]
    \centering
    \includegraphics[width=0.8\textwidth]{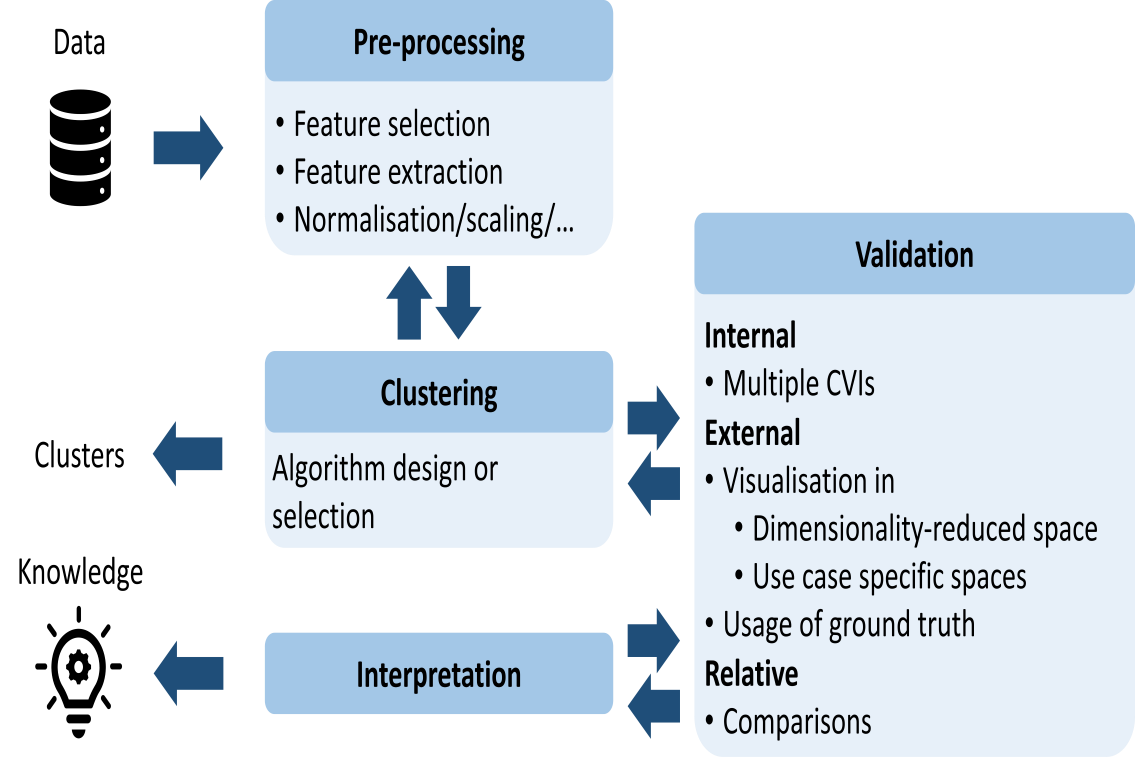}
    \caption{Clustering procedure adapted from \cite{rui2005}. The workflow was iterative and comprised (i) pre-processing,  (ii) clustering,  (iii) external, internal and relative validation that helped refine the selection of pre-processings, clustering methods and hyperparameter settings, and (iv) interpretation within the respective context leading to (new) knowledge.}
    \label{fig:method}
\end{figure}

Following a systematic clustering procedure (Fig.~\ref{fig:method}), three clustering methods were selected (Python library scikit-learn, version 1.5; \cite{pedregosa2011}) for comparison: K-Means (the baseline), agglomerative Ward clustering (a hierarchical method) and DBSCAN (a density-based algorithm). 
All six parameters were scaled to a range from zero to one (MinMaxScaler, Python library scikit-learn, version 1.5)\change{. This made}{ making} the values comparable and usable with distance-variant methods like k-Means clustering.

To assess the influence of pre-processing, each algorithm was applied to the scaled data and to a dimensionality-reduced version of the scaled data. For the \change{dimensionality reduction}{projection} from six parameters (6D) to 3D, the non-linear method UMAP (Python library umap-learn, version 0.5.5; \cite{umap_software}) was applied. The resulting 3D space will be called embedding in the following. 
This study focused on clustering methods that map a data point to exactly one cluster/label and assumed that the number of clusters must be $K>1$. The terms cluster and label will be used interchangeably. For methodological details see Supplementary Material B.

\remove{We defined \textit{confetti} as clusters too small to be scientifically interesting for the current, basin-wide application (e.g. Fig.~S3). Each cluster, for which $|cluster|/|grid|\cdot100 \le 0.1$
(Eq. 1) holds is considered \textit{confetti}. This means that if a cluster contains less data points than 0.1\% of the overall data, it is \textit{confetti}. This is an assessment to describe the clustering and does not imply that \textit{confetti} clusters would not contain meaningful information for other research questions.}

To analyse the importance of each input parameter, cluster assignments were replicated with a more interpretable, predictive model by using the six scaled parameters as input and the cluster labels as output. The idea is similar to translating the clustering function into a neural network \citep{kauffmann2024}, though without explicitly transferring the formulas and thus additionally requiring a second training step. 
For cluster assignment replication, a random forest classifier (RandomForestClassifier, Python library scikit-learn, 1.4) was trained for each clustering model and its inherent feature importance was leveraged.
The hyperparameters were configured as follows: The number of trees was set to 1,000, weights were balanced and the random state was fixed. 

\subsection{Validation}
\label{sec:methods:validation}
Validation can generally be categorized into internal and external approaches.
Sometimes relative validation is listed as a third option referring to the comparison of different models \citep{rui2005}. Internal validation exploits information available during the modelling process. In particular for clustering, Cluster Validity Indices (CVIs) or “scores” provide information on how cohesive a cluster is within itself and how separate it is to other clusters. 
\add{Here, distance, density, and neighborhood based CVIs were used. The applied distance based CVIs are }the Calinski-Harabasz (CH, \cite{calinski1974}), Davies-Bouldin (DB, \cite{davies1979}) and Silhouette (SH, \cite{rousseeuw1987}) scores (Python library scikit-learn, version 1.5).
\add{ The applied density and neighborhood based CVIs are  k-Density-Based Cluster Validation (k-DBCV, }\cite{hammer2024}\add{; adaption of Python library kdbcv, version 1.0.0; a more efficient implementation of DBCV, }\cite{moulavi2014}\add{), 
Clustering Validation Index based on Nearest Neighbours by }\cite{halkidi2015}\add{ (CVNNH, Python library ascvi, version 0.1.0, }\cite{schlake2024}\add{; an adaption of CVNN, }\cite{liu2013}\add{) and Contiguous Density Region (CDR, }\cite{rojasthomas2021}\add{; Python library ascvi, version 0.1.0)}.
\add{ The CVIs }were computed to determine hyperparameters and compare performance\add{ and are described in detail in  Supplementary Material B.3}. 
Desirable high within-cluster cohesiveness and between-cluster separation is indicated by low DB\change{ and SH and}{, CDR and CVNNH and by high SH and k-DBCV as well as} by a local or global maximum of CH.

External validation is achieved by additional knowledge, such as ground truth labels or domain expertise. For biogeochemical and physical clustering, basic principles can be used to evaluate the cluster sets in their 3D geographic space as well as in their temperature-salinity (TS) space. Visual cluster examination can also be conducted in a dimensionality-reduced feature space to assess compliance with feature topology.

\add{For dimensionality reduction methods like UMAP that enhance associations between data structures, several internal validation metrics have been proposed to e.g. assess embedding quality and guide hyperparameter tuning, including reconstruction error }
\citep{zhang2021}\add{. Trustworthiness and continuity evaluate how well local neighbourhoods are preserved during the projection }\citep{venna06}\add{. Trustworthiness measures the extent to which neighbours in the embedding were also neighbours in the original space, while continuity measures whether neighbours in the original space remain close in the embedding. Qlocal and Qglobal provide an alternative approach to quantifying neighbourhood preservation at different scales }\citep{lee2010}\add{. Unlike trustworthiness and continuity, which are based on binary neighbour relationships, these metrics consider relative ranking of all pairwise distances. Trustworthiness, continuity, Qglobal and Qlocal range between zero and one, with higher values indicating better structure preservation. They were computed using the Python library pydrmetrics, version 0.0.7 }\citep{zhang2021}\add{ on a random subset of 2,000 data points for computational efficiency.}

\subsection{Uncertainty Assessment}
\add{Uncertainty was systematically examined in four parts of the analysis. First,}\change{As the selected clustering pipeline (UMAP-DBSCAN) contains non-deterministic components, 100 clustering runs were performed and combined into a final cluster set following the NEMI framework}{100 clustering runs were performed and combined into a final cluster set following the NEMI framework since the selected clustering pipeline (UMAP-DBSCAN) contains non-deterministic components}. 
Variability of the complete process was assessed by\add{ the Adjusted Rand Index (ARI), Normalised Mutual Information (NMI),} cluster overlap (see Section~\ref{sec:cluster_similarity}) and by NEMIs inherent grid-cell-wise uncertainty. 
\change{Moreover}{Second}, this work addressed uncertainty of the dimensionality reduction alone by computing the UMAP embedding 100 times and by comparing the similarity between runs using the root mean-squared error (RMSE). 
\change{F}{Third, f}or hyperparameter tuning, uncertainty was taken into account by computing each hyperparameter combination ten times for each experiment (see Section~\ref{sec:results:clustering_results}). 
\change{T}{Fourth, t}o quantify uncertainty of DBSCAN, it was applied 100 times to a fixed embedding. 

Additional sources of uncertainty of the presented approach were data quality and coverage, as well as missing value imputation. Data was filtered using existing quality flags, wherefore data quality was not further investigated in this study. It should be noted though that the COMFORT dataset comprises data measured at different times and with different instruments\change{, which decreases}{decreasing} accuracy and precision. 
Another potential source of uncertainty was the imputation of non-existent measurement values in the defined geospatial grid, which was considered by flagging each imputed data point. 

% \subsubsection{Uncertainty Quantification}
\change{The uncertainty in the final cluster assignments is quantified from an ensemble of partitions.}{Following the NEMI framework, an ensemble of clustering runs was performed and aggregated, from which NEMI infers uncertainty to assess variability.} The algorithm requires manual selection of a base cluster set (\textit{base\_id}), against which all other cluster sets are compared. The process begins by matching labels across ensemble members. For this, each cluster set is reassigned new labels sorted by cluster size (with label zero corresponding to the largest cluster). For every member cluster set (except the base), the algorithm visits every possible pair of a base label $a_i$ and a member label $b_j$ and determines the overlap as

\begin{equation}
    \text{NEMI\_overlap}(a_i, b_j) = \frac{\text{volume}(a_i \cap b_j)}{\text{volume}(a_i \cup b_j)}
\label{eq:nemi_overlap}
\end{equation}

i.e., the volume of geographically shared data points divided by the joint volume of data points. The original NEMI implementation works with cell count instead of volume.

The ensemble approach also allows uncertainty quantification per grid cell, which was computed here as the number of times the cell was assigned to different clusters:

\begin{equation}
    \text{NEMI\_uncertainty(grid\_cell)} = 100 - \frac{ |\text{same\_cluster\_assignment}|}{|\text{ensemble}|}
\label{eq:nemi_uncertainty}
\end{equation}

\subsection{Cluster Similarity Metrics}
\label{sec:cluster_similarity}
\add{When clusters are generated, comparisons to other sets of clusters or regionalisations, i.e. relative validation, is an important step for putting results into context. Various  metrics are available to assess similarity of cluster sets (cf. e.g.  }\cite{vinh2010}\add{) such as overlap }\citep{manning2008}\add{, Adjusted Rand Index (ARI, }\cite{hubert1985}\add{) and Normalised Mutual Information (NMI, }\cite{strehl2002}\add{).}
\remove{The overlap measure (Manning et al., 2008) served as a similarity metric.}

\subsubsection{Adjusted Rand Index}
The Adjusted Rand Index (ARI, Python library scikit-learn, version 1.5, \cite{hubert1985}) is a symmetric measure to assess similarity between two cluster sets. It is based on the Rand Index (RI), which measures the proportion of agreeing pairs of points between two cluster sets

\begin{equation}
    \mathrm{RI} = \frac{a + b}{\binom{n}{2}},
\end{equation}

where \(a\) is the number of pairs of points that are in the same cluster in both partitions, \(b\) is the number of pairs that are in different clusters in both partitions, and \(n\) is the total number of points. The ARI adjusts this measure for chance agreement 

\begin{equation}
   \mathrm{ARI} = \frac{\mathrm{RI} - \mathbb{E}[\mathrm{RI}]}{max(\mathrm{RI}) - \mathbb{E}[\mathrm{RI}]}
\label{eq:ari} 
\end{equation}

where \(\mathbb{E}[\mathrm{RI}]\) denotes the expected RI of random labelings. The index is bound between \SI{-0.5}{} and one, where identical cluster sets receive a score of one, a random label matching a score of zero and complete disagreement a negative score. 

\subsubsection{Normalised Mutual Information}
Normalised Mutual Information (NMI, \cite{strehl2002}) quantifies the amount of shared information between two cluster sets. In this study, the normalisation proposed by \cite{kvalseth1987} was applied yielding 

\begin{equation}
   \mathrm{NMI} = \frac{2I(A, B)}{H(A) + H(B)} = \frac{2 \big(H(A) + H(B) - H(A,B)\big)}{H(A) + H(B)}
\label{eq:nmi} 
\end{equation}

where \(H(A)\) and \(H(B)\) denote the entropies of the two cluster sets \(A\) and \(B\), \(H(A,B)\) their joint entropy, and \(I(A, B)\) the mutual information. 

NMI ranges from zero to one, where one indicates perfect agreement (identical cluster sets) and zero denotes no shared information.

\subsubsection{Cluster Overlap}
An asymmetric measure to compare two cluster sets $A$ and $B$ and hence quantify reproducibility is overlap or purity \citep{manning2008}, which ranges between 0 and 1. 
It assesses how much the clusters of two cluster sets overlap by calculating the maximum overlap for each cluster/label in $A$ with labels in $B$ and vice versa. The average overlap is used as a cluster set similarity measure.

Let there be $N$ objects that are grouped by clustering $A$ into clusters $a_0, ..., a_I$ and by clustering $B$ into clusters $b_0, ..., b_J$. The overlap of $A$ with $B$ is then defined as

\begin{equation}
    \text{overlap}(A, B) = \frac{1}{N}\sum_i \max_j|a_i \cap b_j|
\label{eq:overlap}
\end{equation}

To obtain a symmetric measure, we computed the overlap as 

\begin{equation}
    \text{overlap} = \frac{\text{overlap}(A, B) + \text{overlap}(B, A)}{2}
\label{eq:symmetric_overlap}
\end{equation}

% Results ----------------------------------------------
 
\section{Results}

\subsection{Embedding using UMAP}
\label{sec:results:umap}
\begin{figure}[ht]
    \centering
    \includegraphics[width=\textwidth]{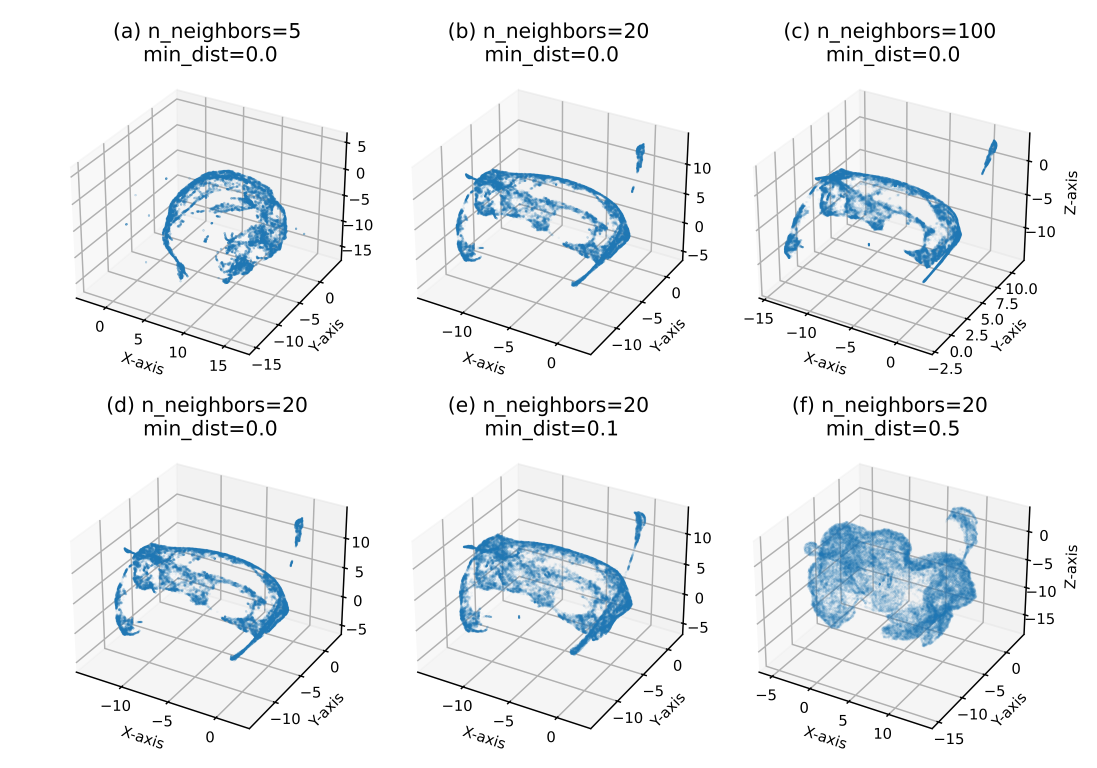}
    \caption{Influence of the UMAP hyperparameters number of neighbours (\textit{n\_neighbors}, top) and minimum distance (\textit{min\_dist}, bottom) on the embedded space. With an increasing number of neighbours, the topology unfolded but did not substantially change for more than 20 neighbours. With a growing minimum distance, the data structure lost structural detail.}
    \label{fig:umap_tuning}
\end{figure}

As an initial validation step, the influence of UMAP hyperparameters was tested. \add{The parameters control if global or local associations within the data are emphasised by minimising the cross-entropy (Supplementary Material B.1).}
A lower number of neighbours led to a more curled structure, while a higher number unfolded it (Fig.~\ref{fig:umap_tuning}, a-c). Increasing the number of neighbours to more than 20 did not further change the embedding (Fig.~\ref{fig:umap_tuning}, c). As expected, a higher minimum distance caused the data points to spread apart further, blurring finer topologies (Fig.~\ref{fig:umap_tuning}, e-f). For the final cluster runs, the number of neighbours was set to 20 and the minimum distance to 0 (Fig.~\ref{fig:umap_tuning}, b and d) so that the embedding-space, with three dimensions, exhibited a distinct topology. 
\add{Over 20 iterations, the final embedding achieved a mean trustworthiness of $0.97 \pm 0.004$ and a mean continuity of $0.97 \pm 0.007$. Global neighbourhood was better preserved than local ones ($Qglobal=0.87 \pm 0.009$, $Qlocal=0.66 \pm 0.01$, Fig.~S4).} \add{A Shepard plot of a UMAP embedding is illustrated in Supplementary Material Fig.~S4. }

\subsection{Clustering Results}
\label{sec:results:clustering_results}
Six experiments were carried out during which three clustering methods were applied to original and embedded data. Each experiment was run ten times for every hyperparameter combination.\add{ A striking result was that little agreement between the CVIs was found }(Table~\ref{tab:experiments}).

\begin{table}[ht]
\caption{Overview of clustering results. For external validation, the clustering was visually analysed in geographical and embedding spaces (last two columns, cf. Section~\ref{sec:results:clustering_results}). Note that the Cluster Validity Indexes (CVIs), i.e. Calinski-Harabasz Score (CH), Davies-Bouldin Score (DB), Silhouette Score (SH), k-Density Based Clustering Validation (k-DBCV), Clustering Validation Index Based on Nearest Neighbours Halkidi (CVNNH) and Contiguous Density Region (CDR), are not suitable to compare across clustering methods (see Section~\ref{sec:discussion:cvis}). High CH, SH and k-DBCV as well as low DB, CVNNH and CDR indicate a well separated clustering.}
\label{tab:experiments}
\setlength{\tabcolsep}{4pt}
\begin{tabularx}{\textwidth}{@{} p{1.3cm} p{1.7cm} p{2.5cm} 
>{\centering\arraybackslash}p{0.9cm} 
>{\centering\arraybackslash}p{0.9cm} 
>{\centering\arraybackslash}p{0.9cm} 
>{\centering\arraybackslash}p{0.9cm} 
>{\centering\arraybackslash}p{0.9cm} 
>{\centering\arraybackslash}p{0.9cm} 
X X@{}}
\toprule
\multirow{2}{*}{\textbf{Data}} & 
\multirow{2}{*}{\textbf{Algorithm}} & 
\multirow{2}{*}{\textbf{Hyperparameters}} & 
\multicolumn{6}{c}{\textbf{Internal Validation}} & 
\multicolumn{2}{c}{\textbf{External Validation}} \\
\cmidrule(lr){4-9} \cmidrule(l){10-11}
& & & \textbf{CH $\uparrow$} & \textbf{DB $\downarrow$} & \textbf{SH $\uparrow$} & \textbf{k-DBCV $\uparrow$} & \textbf{CVNNH $\downarrow$} & \textbf{CDR $\downarrow$} & \textbf{Geo} & \textbf{UMAP} \\
\midrule

Original & K-Means & $n_{\text{clusters}} = 2$ 
& 57,358 & 0.78 & 0.50 & \SI{-1.00}{} & 0.24 & 0.57 
& Warm-cold separation & Straight cut \\
Embedded & K-Means & $n_{\text{clusters}} = 8$ 
& 49,703 & 0.71 & 0.47 & \SI{-0.78}{} & 3.55 & 0.60 
& Non-separation of deep areas & Non-separations \\
Original & Agg. Ward & $n_{\text{clusters}} = 2$ 
& 51,825 & 0.84 & 0.48 & \SI{-1.00}{} & 0.24 & 0.56
& Hot-cold like & Intrusions \\
Embedded & Agg. Ward & $n_{\text{clusters}} = 25$ 
& 65,901 & 0.72 & 0.46 & \SI{-0.61}{} & 1.72 & 0.59 
& Geo. connected & Intrusions, non-separations \\
Original & DBSCAN & $\epsilon = 0.11949153$\newline$min_{\text{samples}} = 11$ 
& 800 & 1.44 & 0.55 & \SI{-1.00}{} & 0.98 & 0.54
& Focus on Baltic & One big cluster \\
Embedded & DBSCAN & $\epsilon = 0.10661017$\newline$min_{\text{samples}} = 4$ 
& 1,754 & 1.5 & \SI{-0.35}{} & \SI{-0.41}{} & 3.31 & 0.56 
& Density-oriented & Many small clusters \\

\bottomrule
\end{tabularx}
\end{table}

\subsubsection{K-Means}
\begin{figure}[ht]
    \centering
    \begin{subfigure}[b]{0.45\textwidth}
        \centering
        \includegraphics[width=\textwidth]{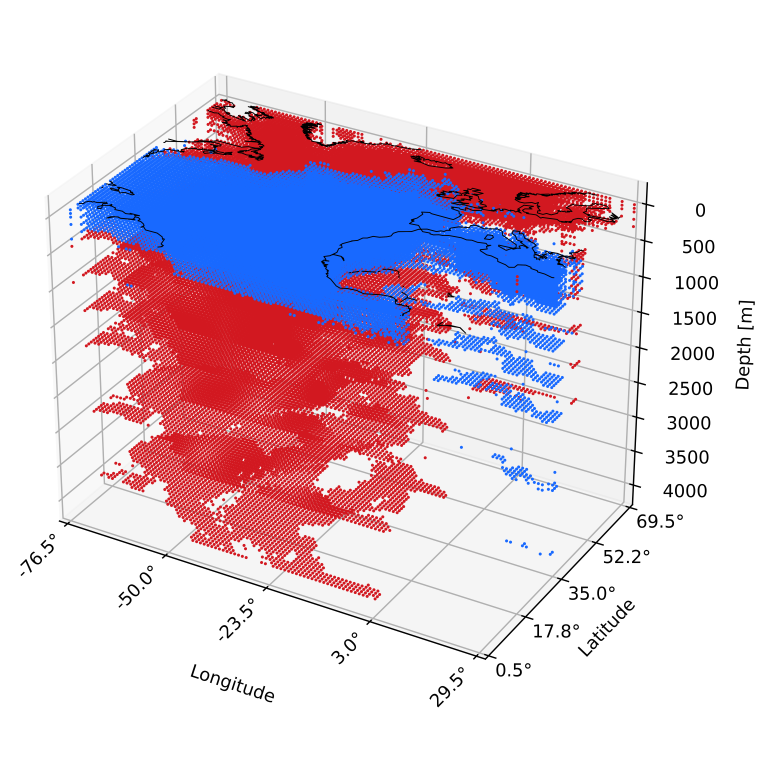}
        \label{fig:kmeans_original_geo}
    \end{subfigure}
    \raisebox{7\height}{ % This adjusts the vertical alignment of the middle figure (the arrow)
        \begin{subfigure}[b]{0.05\textwidth} 
            \centering
            \includegraphics[width=\textwidth]{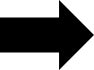}
        \end{subfigure}
    }
    \begin{subfigure}[b]{0.45\textwidth} 
        \centering
        \includegraphics[width=\textwidth]{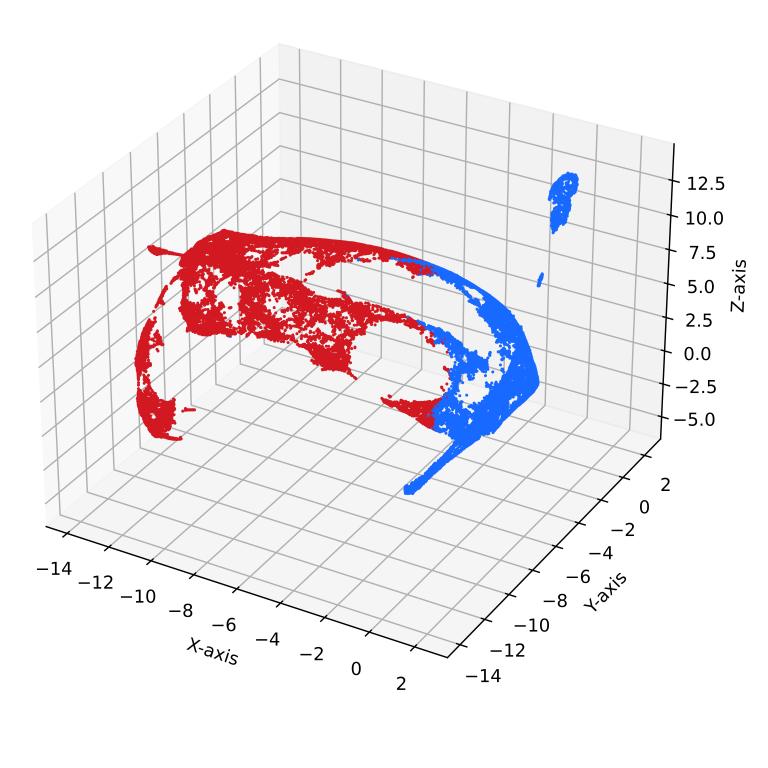}
        \label{fig:kmeans_original_umap}
    \end{subfigure}
\caption{K-Means applied to the original data with two clusters in geographic space (left) and its projection into the embedded space (right). The geographic separation of the clusters resembled the temperature distribution. In embedding space, the division approximated a straight cut.}
    \label{fig:kmeans_original}
\end{figure}

\textbf{Clustering original data.}
For k-Means applied to the original 6D data, all three\add{ classical distance-based} scores\add{ - CH, DB and SH -} reached peak performance when setting the number of clusters to two (Fig.~S5). Plotting the result of k-Means with two clusters in geographic space, two coherent regions emerged (Fig.~\ref{fig:kmeans_original}). 
They most strongly resembled the input temperature distribution, supported by temperature having the highest feature importance (\SI{47}{\percent}). \add{Structural validity scores, in contrast, showed different preferences: CVNNH reached its optimum at 12 clusters, while CDR and k-DBCV favored higher values, continuing to improve up to 60 clusters, though k-DBCV remained negative throughout.}
Note that clustering here is not performed on the embedding. The embedding is merely used to visualise how the clusters look like in this space. 
The visualisation in embedding space showed clear errors: Structures that are separate were not subdivided, other clusters intruded into foreign clusters.

Higher numbers of clusters resulted in more globular clusters in embedding space (Fig.~S6). Across most clustering levels, temperature and oxygen emerge as the most important features, while the lower-ranked feature of phosphate slightly gained in importance for larger cluster numbers. 

\begin{figure}[ht]
    \centering
    \begin{subfigure}[b]{0.45\textwidth}
        \centering
        \includegraphics[width=\textwidth]{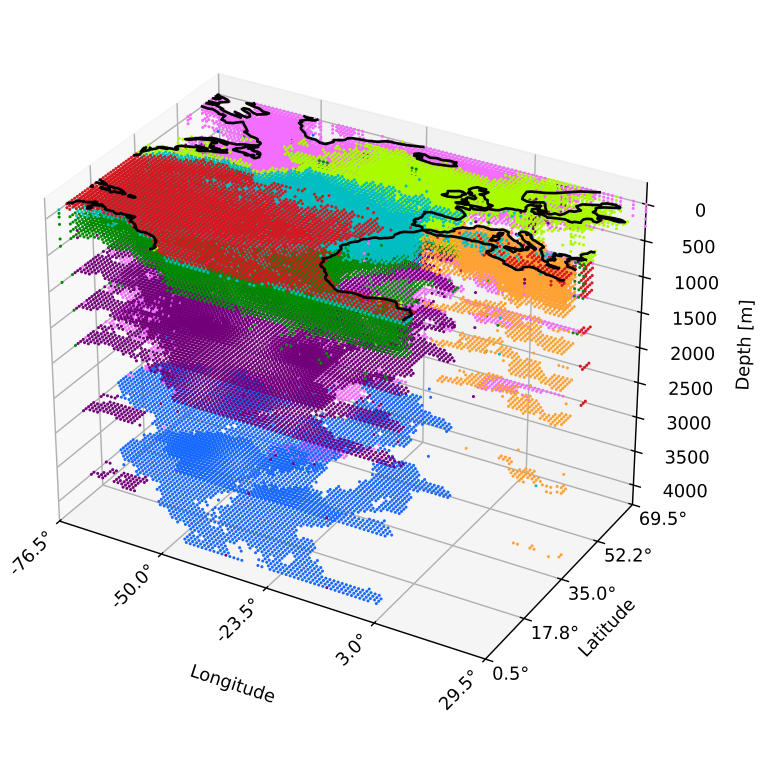}
    \end{subfigure}
    \raisebox{7\height}{ % This adjusts the vertical alignment of the middle figure (the arrow)
        \begin{subfigure}[b]{0.05\textwidth} 
            \centering
            \includegraphics[width=\textwidth]{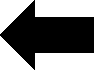}
        \end{subfigure}
    }
    \begin{subfigure}[b]{0.45\textwidth} 
        \centering
        \includegraphics[width=\textwidth]{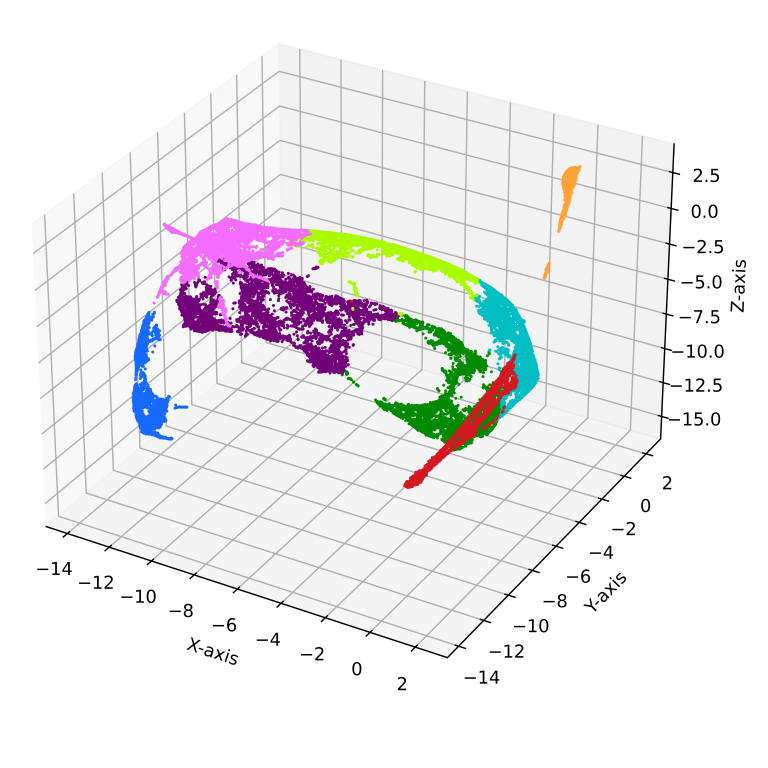}
    \end{subfigure}
    \caption{K-Means applied to the embedded data with \change{ten}{eight} clusters (right) and its projection to geographic space (left). Insufficient performance is reflected in clusters that are not separated despite their clear segregation in embedding space, e.g. the blue cluster.}
    \label{fig:kmeans_embedding}
\end{figure}

\medskip
\noindent
\textbf{Clustering embedded data.}
For k-Means applied to the data in the previously computed embedded space (Section~\ref{sec:results:umap}), internal validation (the scores) suggested numbers of clusters greater than two (Fig.~S5). 
\add{SH and DB both favoured eight clusters, while the other scores rose/fell beyond the selected value range with CH peaking locally at 14 clusters.}
\change{The subsequent external validation revealed a more representative clustering as compared to k-Means on original data.- However, there were still non-separated regions visible in UMAP space}{Despite some improvement over clustering on original data, the resulting clusters still showed non-separated regions in embedding space} (Fig. \ref{fig:kmeans_embedding}).

\change{In case that the scores did not detect the proper number of clusters, the cluster sets, generated by varying the number of clusters, were examined by external validation.}{To compensate for potential inaccuracies in internal validation, cluster sets across different numbers of clusters were further evaluated externally.} A higher number of clusters resulted in the embedded space adapting a chessboard pattern (Fig.~S7), that was clearly not representative of the underlying co-variance space given by the data.  

\subsubsection{Agglomerative Ward}
\label{sec:results:ward}
\begin{figure}[ht]
    \centering
    \begin{subfigure}[b]{0.45\textwidth}
        \centering
        \includegraphics[width=\textwidth]{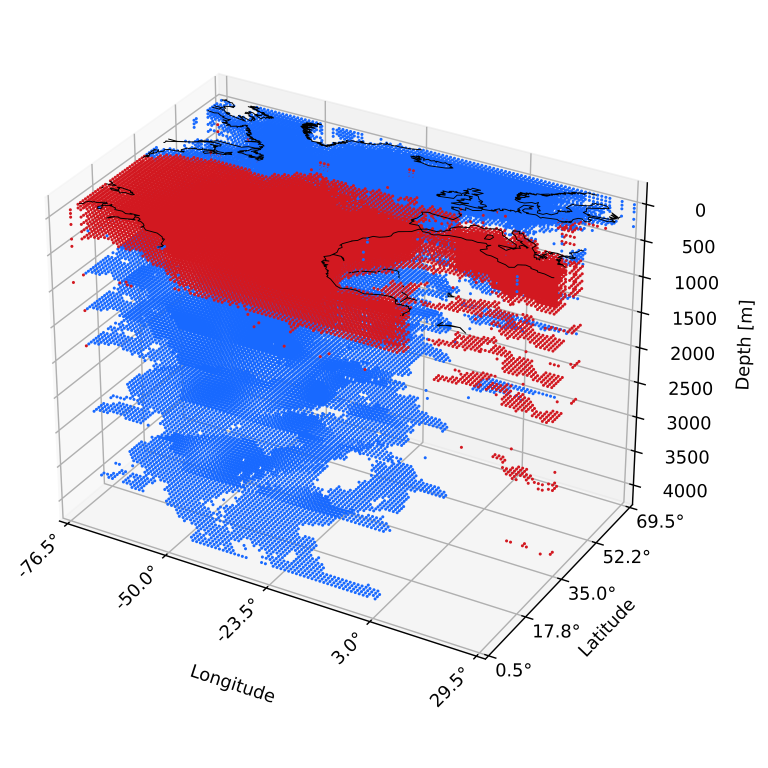}
    \end{subfigure}
    \raisebox{7\height}{ % This adjusts the vertical alignment of the middle figure (the arrow)
        \begin{subfigure}[b]{0.05\textwidth} 
            \centering
            \includegraphics[width=\textwidth]{images_medium/fig_right_arrow.png}
        \end{subfigure}
    }
    \begin{subfigure}[b]{0.45\textwidth} 
        \centering
        \includegraphics[width=\textwidth]{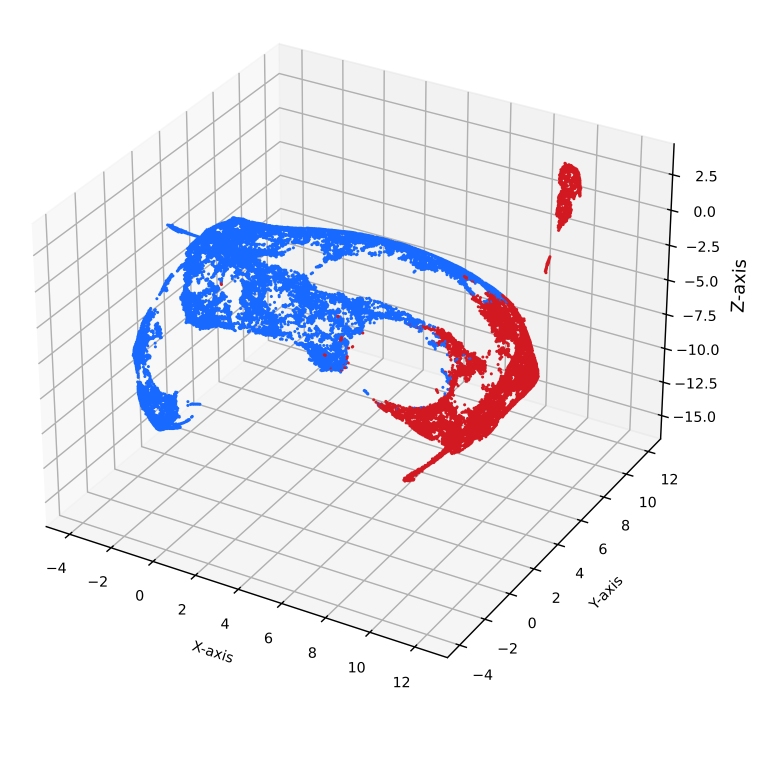}
    \end{subfigure}
    \caption{Agglomerative Ward clustering applied to the original data with two clusters (left) and its projection into the embedded space (right).}
    \label{fig:ward_original}
\end{figure}

\textbf{Clustering original data.}
Similar to k-Means on the original data, all three \add{classical} CVIs suggested two clusters as the best split using agglomerative Ward clustering (Fig.~S8). The visualisation in the embedding revealed a straight cut (Fig.~\ref{fig:ward_original}) and temperature had the highest importance (\SI{35}{\percent}) followed by oxygen (\SI{34}{\percent}). \add{Structural scores (k-DBCV, CVNNH and CDR) indicated more fine-grained cluster sets, with optimal values of 27, 12 and 30 clusters, respectively. }
Increasing the number of clusters up to 30, regardless of the scores, resulted in unseparated clusters and intrusion issues (visible in the embedded space), though less than using k-Means. Temperature contributed most to separation followed by oxygen and salinity. 

\begin{figure}[ht]
    \centering
    \begin{subfigure}[b]{0.45\textwidth}
        \centering
        \includegraphics[width=\textwidth]{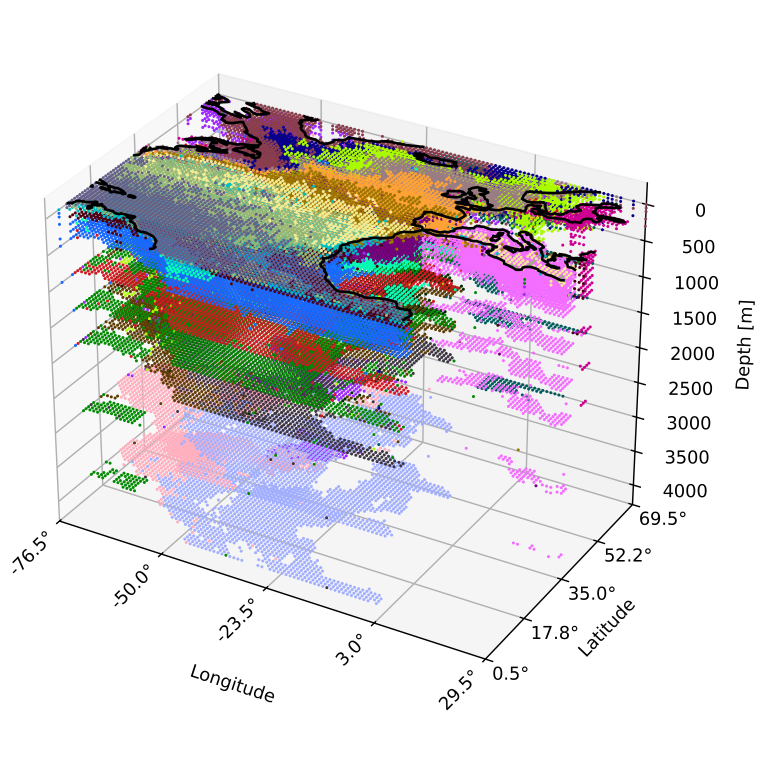}
    \end{subfigure}
    \raisebox{7\height}{ % This adjusts the vertical alignment of the middle figure (the arrow)
        \begin{subfigure}[b]{0.05\textwidth} 
            \centering
            \includegraphics[width=\textwidth]{images_medium/fig_left_arrow.png}
        \end{subfigure}
    }
    \begin{subfigure}[b]{0.45\textwidth} 
        \centering
        \includegraphics[width=\textwidth]{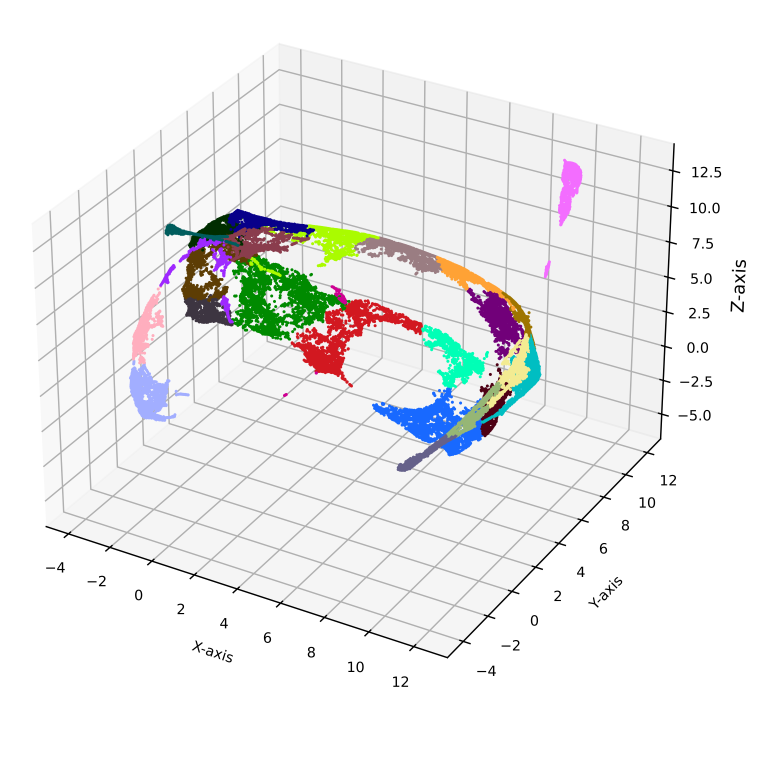}
    \end{subfigure}
    \caption{Agglomerative Ward applied to embedded data with \change{23}{25} clusters (right) and its projection to geographic space (left).}
    \label{fig:ward_embedding}
\end{figure}

\medskip
\noindent
\textbf{Clustering embedded data.}
The scores for agglomerative Ward clustering applied to the embedded data preferred a number of clusters larger than two (Fig.~S8). While CH and CVNNH did not show an extreme value that could be used to optimise the number of clusters, SH suggested 24 clusters as the optimum. DB and k-DBCV preferred 24 clusters and CDR was minimal for 23 and 28 clusters. 
Selecting \change{23 clusters for less complexity}{25} reasonably divided the embedded data at areas of low data point density, except for\add{ the Labrador Sea and} some smaller clusters (Fig.~\ref{fig:ward_embedding}). Deeper ocean regions (two bulbous structures on the left side of embedding space) and the Mediterranean water masses were well-separated while clustering of original data was not able to find these differences. 

\subsubsection{DBSCAN}
\label{sec:results:dbscan}
In contrast to k-Means and Agglomerative Ward clustering, DBSCAN requires tuning of two main hyperparameters: \textit{epsilon}, i.e., the search radius, and \textit{min\_samples}, i.e., the minimum cluster size. DBSCAN labels data points that it could not assign to a cluster as noise. Therefore, the amount of noise was taken into account for the final choice of hyperparameter settings.
For hyperparameter tuning of \textit{min\_points}, values between 2 and 11 and for \textit{epsilon} values between 0.01 and 0.2 (20 steps with step size 0.00322) were tested for original and embedded data. 
The number of clusters was considered for hyperparameter selection to prevent an excessive number of regions. 
For this, the elbow method was applied in the 2d space given by the two hyperparameters, i.e., by selecting the steepest gradient in cluster number in the heatmap \citep{sonnewald2020}.

\begin{figure}[ht]
    \centering
    \begin{subfigure}[b]{0.45\textwidth}
        \centering
        \includegraphics[width=\textwidth]{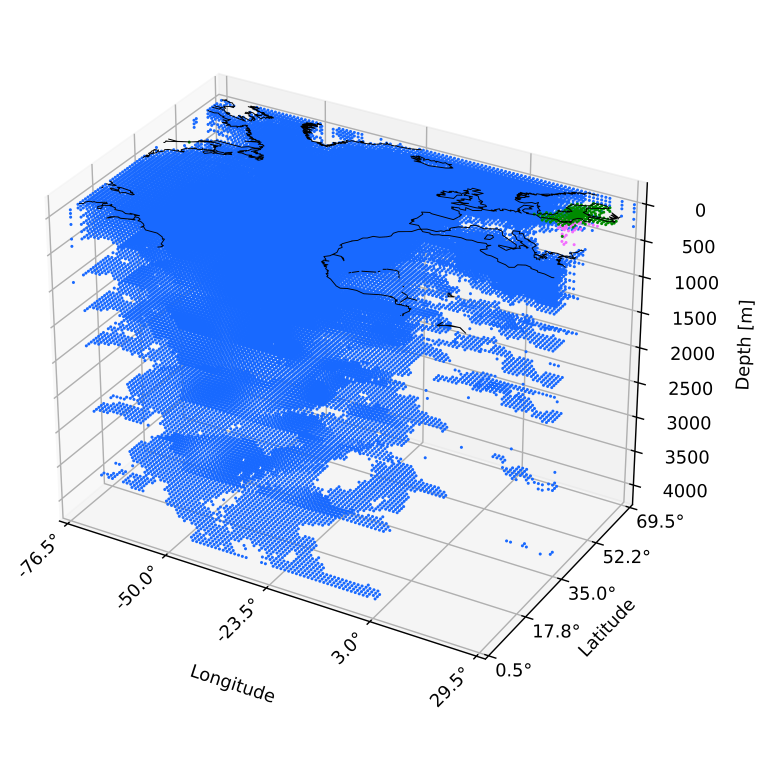}
    \end{subfigure}
    \raisebox{7\height}{ % This adjusts the vertical alignment of the middle figure (the arrow)
        \begin{subfigure}[b]{0.05\textwidth} 
            \centering
            \includegraphics[width=\textwidth]{images_medium/fig_right_arrow.png}
        \end{subfigure}
    }
    \begin{subfigure}[b]{0.45\textwidth} 
        \centering
        \includegraphics[width=\textwidth]{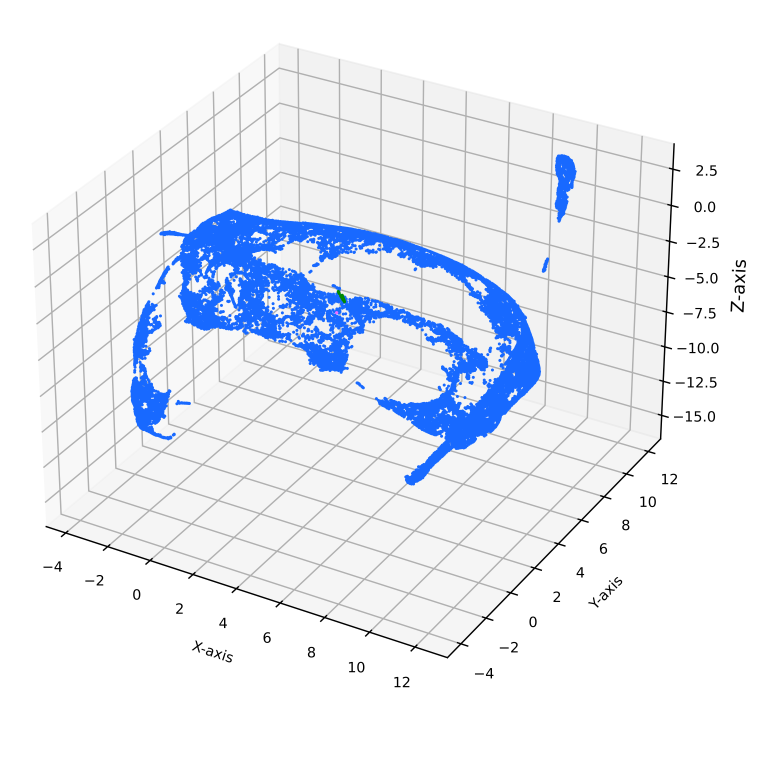}
    \end{subfigure}
    \caption{DBSCAN applied to original data (left) and its projection to embedding space (right). The hyperparameters were selected through the Calinski-Harabasz Score (CH). The clustering was heavily focused on the Baltic.}
    \label{fig:dbscan_original}
\end{figure}

\medskip
\noindent
\textbf{Clustering original data.}
All tested hyperparameter combinations for DBSCAN applied to the original 6D data resulted in similar clusterings that heavily focused on the Baltic, Black and/or Mediterranean Sea (Fig.~\ref{fig:dbscan_original}) independent of the tuning criterion (Table~ST1). For low \textit{epsilon}, some surface clusters were identified along with 40 - \SI{60}{\percent} DBSCAN noise. 
The number of small clusters \remove{or \textit{confetti} (Section~2.2)} decreased with increasing \textit{epsilon} while higher \textit{min\_samples} led to less \change{\textit{confetti}}{small clusters} and more coherent, but few (mostly below 10) regions.

CH favoured a high \textit{min\_samples} producing a low-noise clustering with coherent regions in the Baltic Sea. DB preferred fewer \textit{min\_samples} and generated a clustering with low noise but spatially less coherent clusters. SH \add{and CVNNH }suggested high values for \change{both hyperparameters}{\textit{epsilon}}\add{, SH also requiring high \textit{min\_samples}. K-DBCV quantified all points as noise in }\SI{99}{\percent} \add{of the tested combinations. CDR suggested low \textit{epsilon} producing cluster sets with around }\SI{10}{\percent}\add{ noise}. \remove{Using the number of clusters for the hyperparameter selection resulted in about 34 noise, a lot of confetti and only three larger clusters.}

\medskip
\noindent
\textbf{Clustering embedded data.}
The distribution of the number of clusters and the noise fraction was similar for DBSCAN applied to the embedded data compared to original data. However, the number of clusters rose above 6,500 and the noise reached proportions of \SI{99}{\percent}. 
Low \textit{epsilon} and high \textit{min\_samples} generally resulted in highest noise. 
Low \textit{epsilon} and low \textit{min\_samples} led to a large number of \add{small} clusters \remove{(\textit{confetti})}, while an \textit{epsilon} value above 0.1 generated larger, coherent structures. 

The CH (Fig.~S9) formed a clear ridge that rose linearly with increasing \textit{min\_samples}. It separated very coarse cluster sets (above the ridge, i.e., larger \textit{epsilon} values) from cluster sets consisting of smaller regions \remove{and increasing \textit{confetti} }(below the ridge, i.e., smaller \textit{epsilon}). 
The score hence delineated a compromise between coarse and fine cluster sets. 
Its optimal/maximal value (at $epsilon=0.15491525$, $min\_samples=11$) produced a cluster set that incorporated clearly delineated regions in geographic and embedded space. In the embedded space, some larger structures were not subdivided despite only thin connections, such as a geographic region in the north. 
The SH resembled the CH only favouring lower \textit{epsilon} and \textit{min\_samples} resulting in more \change{\textit{confetti}}{small clusters}. In contrast, the DB preferred hyperparameter combinations producing much noise. \add{The structural CVIs (k-DBCV, CVNNH and CDR) generally favoured configurations that produced many small clusters with small differences in the few larger clusters and noise proportion.}

Final hyperparameters were selected based on CH as well as external validation, i.e., the visual inspection of embedded and geographic space. CH was used as orientation since it reflected the trade-off between \change{confetti}{small clusters} versus larger coherent regions. The other scores were ignored due to their lack of agreement with visual clustering quality, i.e., they favoured cluster sets with more noise and/or \change{\textit{confetti}}{many small clusters}. The visual inspection led to the decision of $epsilon=0.10661017$ and $min\_samples=4$ (Fig.~S9, red square). 
\textit{Epsilon} smaller than the selected value resulted in more \change{confetti}{small clusters} while for larger \textit{epsilon}, clusters tended to merge. 
A \textit{min\_samples} smaller than four at the selected \textit{epsilon} formed a larger cluster in the Labrador Sea and more \change{confetti}{small clusters}. The clustering for a minimum of five samples was similar, but had more noise (\SI{5.17}{\percent} versus \SI{3.82}{\percent}). 
Increasing the \textit{min\_samples} up to 11 resulted in smaller regions of smaller size.

\subsection{UMAP-DBSCAN: Uncertainty and Post-Processing}
Based on the above analysis, DBSCAN applied to the data embedded by UMAP best represented the given data structure.
Due to stochasticity of UMAP-DBSCAN, each of the 100 runs produced slightly different results. To assess reproducibility, three sources of uncertainty were examined:

First, UMAP was applied 100 times with the same $n\_neighbors$ and $min\_dist$ to the scaled input data. 
It was not possible to directly compute the Root Mean Squared Error (RMSE) between any two embeddings since they were not always spatially aligned. 
Therefore, the point clouds of the embeddings were aligned through manual translation, rotation and/or mirroring followed by the application of the Iterative Closest Point (ICP, Python library simpleicp, version 2.0.14) algorithm to maximise alignment. 
ICP calculates a translation and rotation matrix for one point cloud trying to minimise distances to another point cloud. 
The RMSE was computed for the embeddings before and after applying ICP. The minimal deviation, i.e. the error when two embeddings aligned best, was stored, resulting in an average RMSE of $0.22\pm0.06$ (or about $1.3\pm$\SI{0.36}{\percent} of the value range)

Second, variability of DBSCAN was assessed by applying it 100 times to a fixed, pre-computed embedding while keeping hyperparameters fixed. \change{For each run, the training data was shuffled}{For each run, the rows of the training data were shuffled as DBSCAN is sensitive to the sequence of input data}. The mean overlap between the resulting cluster sets was $99.99\pm0.003\%$. 

Third, variability of the combined UMAP-DBSCAN pipeline was assessed by running it 100 times, i.e., in each iteration both, the embedding and the clustering, were recomputed. \change{On average, the overlap between the resulting cluster sets}{On average, the ARI between the resulting cluster sets was }$0.78\pm0.05$\add{, the NMI was }$0.91\pm0.01$\add{ and the overlap} was $88.81\pm1.8\%$ (Fig.~S10) .

\begin{figure}[h]
    \centering
    % Top subfigure spanning the full width
    \begin{subfigure}[b]{0.7\textwidth}
        \centering
        \includegraphics[width=\textwidth]{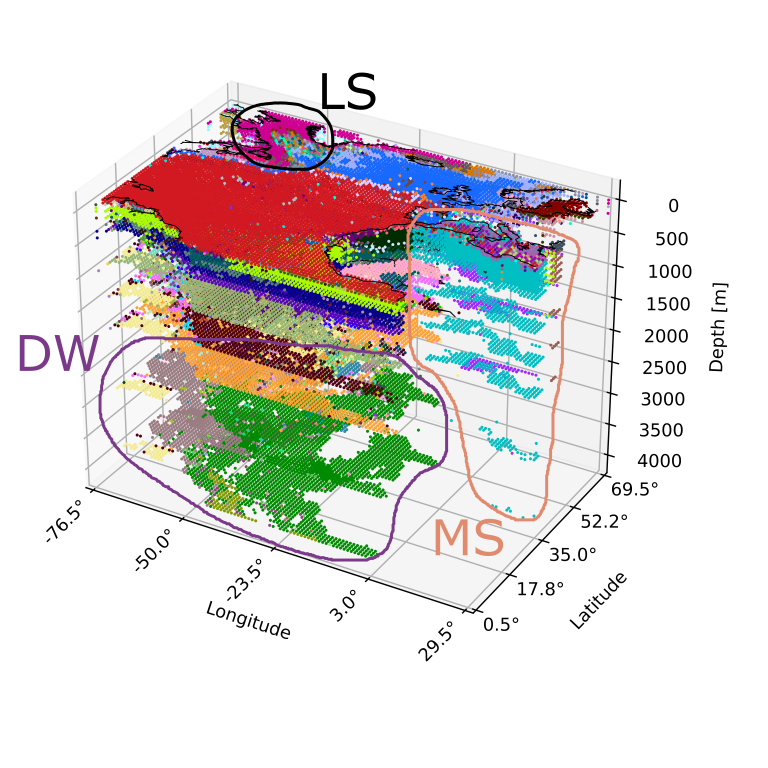}
    \end{subfigure} 
    
    \begin{subfigure}[b]{0.45\textwidth}
        \centering
        \includegraphics[width=\textwidth]{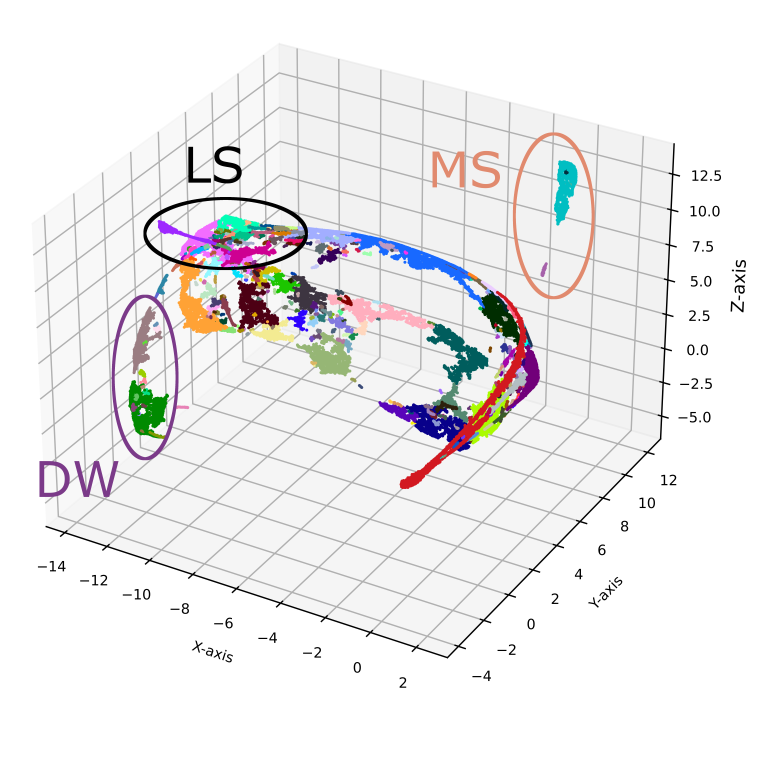}
    \end{subfigure}
    \hspace{0.05\textwidth} 
    \begin{subfigure}[b]{0.45\textwidth}
        \centering
        \includegraphics[width=\textwidth]{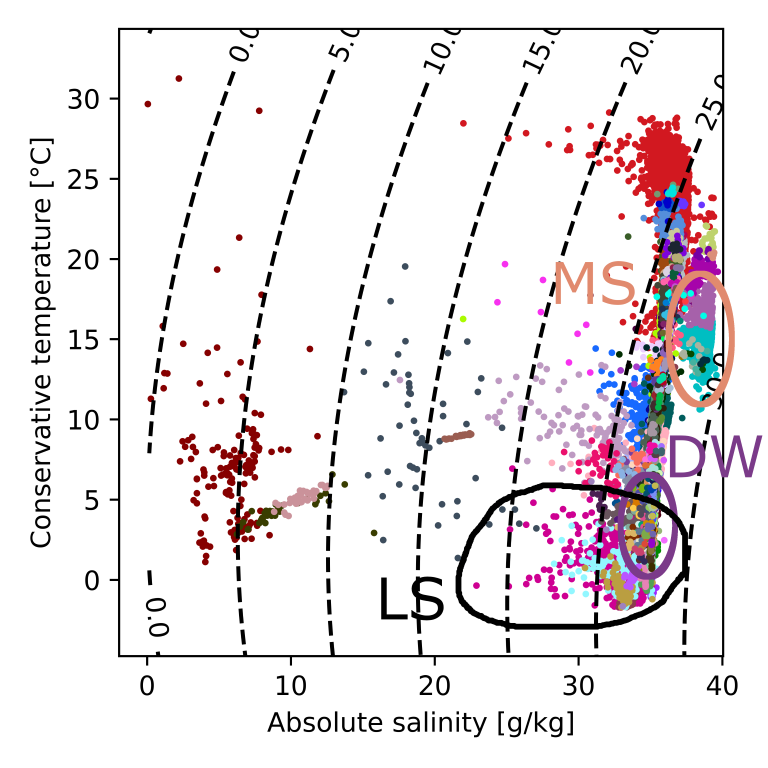}
    \end{subfigure}
    
    \caption{Final UMAP-DBSCAN clustering melted from 100 runs using Native Emergent Manifold Interrogation (NEMI) in geographic (top), embedded (bottom left) and temperature-salinity (TS) space (bottom right). The three use cases (discussed in Section~\ref{sec:results:case_studies}) are highlighted: Deep Atlantic waters (DW), Labrador Sea (LS) and Mediterranean Sea (MS).}
    \label{fig:dbscan_embedding}
\end{figure}

\medskip
\noindent
\textbf{Final cluster assignments.}
To obtain a final cluster set, following the NEMI framework, the individual members of the 100 UMAP-DBSCAN runs were combined.
As \textit{base\_id}, the cluster set with lowest mean uncertainty was chosen.
The final clustering (Fig.~\ref{fig:dbscan_embedding}) had 321 clusters and \SI{3.92}{\percent} ($n=1,920$) of the grid cells were assigned as noise and were subsequently excluded.

\begin{figure}[h]
    \centering
    \begin{subfigure}[c]{0.3\textwidth}
        \centering
        \includegraphics[width=\textwidth]{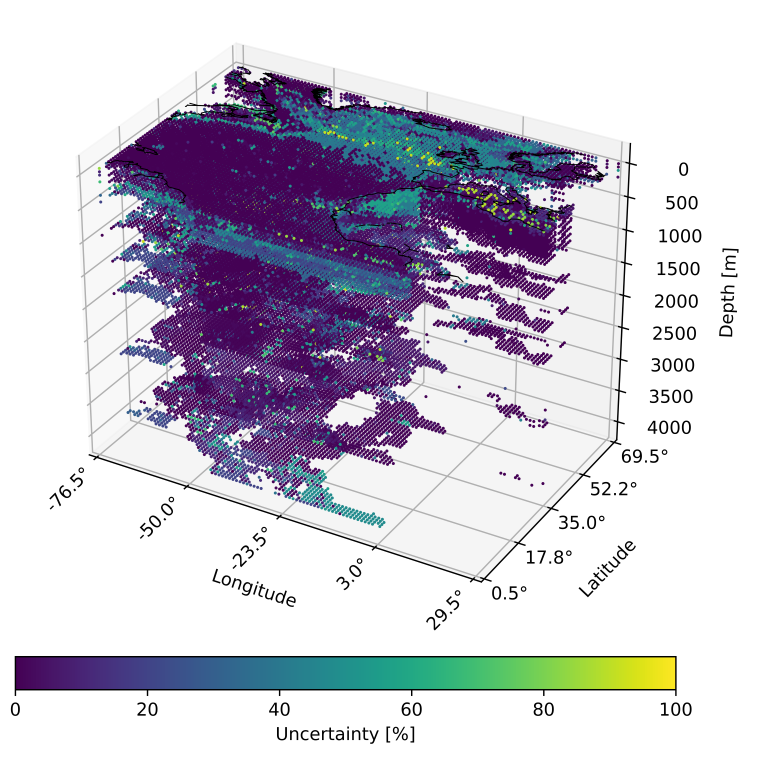}
    \end{subfigure}
    \begin{subfigure}[c]{0.3\textwidth} 
        \centering
        \includegraphics[width=\textwidth]{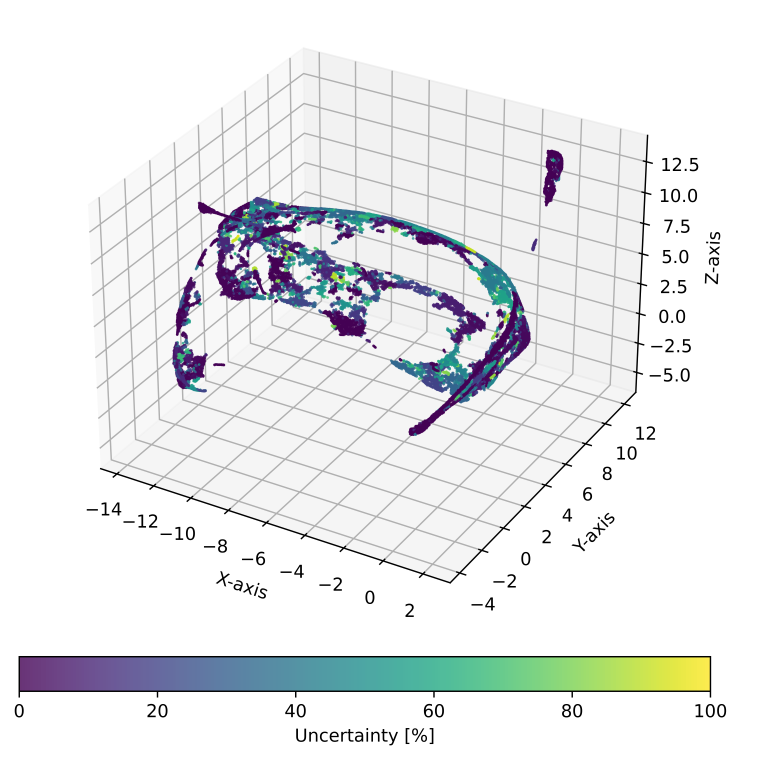}
    \end{subfigure}
    \begin{subfigure}[c]{0.3\textwidth} 
        \centering
        \includegraphics[width=\textwidth]{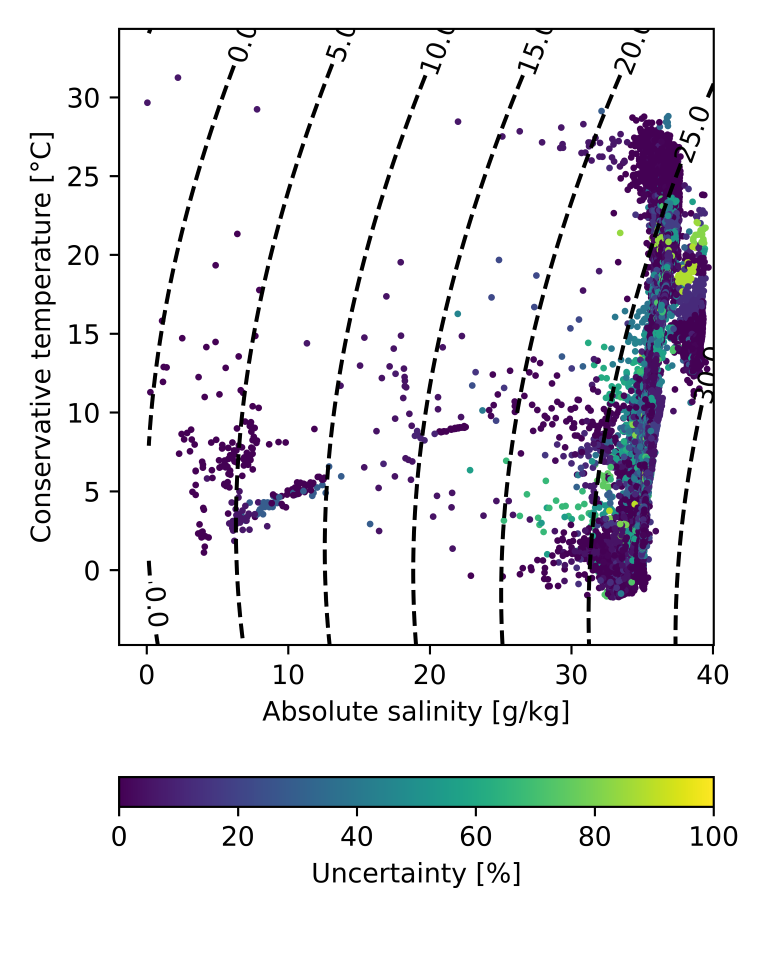}
    \end{subfigure}
    
    \caption{NEMI uncertainty of a UMAP-DBSCAN clustering melted from 100 runs using NEMI in geographic (left), embedded (middle) and temperature-salinity (TS) space (right). The mean uncertainty was $15.49\pm20\%$ with a median of \SI{5}{\percent}. The high standard deviation and low median indicate a few highly uncertain grid cells skewing the mean.}
    \label{fig:nemi_uncertainty}
\end{figure}

The average uncertainty (Fig.~\ref{fig:nemi_uncertainty}) was $15.49\pm20\%$ (min: \SI{0}{\percent}, max: \SI{96}{\percent}) and \SI{50}{\percent} of the uncertainties were $\leq5\%$. 
Lowest mean uncertainties (\SI{0}{\percent}) were found in 6 clusters, three of which were geographically scattered but well separated in embedding space (labels 209, 244, 288). The three certain, geographically cohesive clusters were located in the Black Sea (100 -- \SI{2,000}{\meter}; label 93), in the southern Baltic Sea (100 -- \SI{400}{\meter}; label 102) and in the Gulf of Saint Lawrence (200 -- \SI{1,000}{\meter}; label 112). Uncertainties $\geq50\%$ were geographically scattered with the exception of a cluster stretching from the Strait of Gibraltar until \SI{-64.5}{\degree}~West (0 -- \SI{300}{\meter}; label 28) and a cluster off the coast of North West Africa (200 -- \SI{1,500}{\meter}; label 53).

\clearpage

\clearpage

\subsection{Case Studies}
\label{sec:results:case_studies}
Three ocean regions from the final clustering, which were determined based on the combination of 100 UMAP-DBSCAN runs using the NEMI framework (Fig.~\ref{fig:dbscan_embedding}), were inspected. 
\add{Overall, the final cluster set shows higher similarity to the EMUs than to Longhurst provinces (Tab.~}\ref{tab:literature_comparison}\add{). In both cases, overlap and NMI are moderate, while ARI values are relatively low. } 

\begin{table}[ht]
\caption{Similarity of this study's final cluster set with marine provinces \citep{longhurst2007} and Ecological Marine Units (EMUs, \cite{sayre2017}).}
\label{tab:literature_comparison}
\setlength{\tabcolsep}{4pt}
\begin{tabularx}{\textwidth}{@{} X X X @{}}
\toprule
\textbf{Similarity Score} & \textbf{Longhurst Provinces} & \textbf{EMUs} \\
\midrule
Overlap & 0.56 & 0.62 \\
Normalised Mutual Information & 0.43 &  0.51\\
Adjusted Rand Index & 0.16 &  0.18 \\
\bottomrule
\end{tabularx}
\end{table}

\subsubsection{Mediterranean Sea}
\begin{figure}[ht]
    \centering
    \begin{subfigure}[c]{0.3\textwidth}
        \centering
        \includegraphics[width=\textwidth]{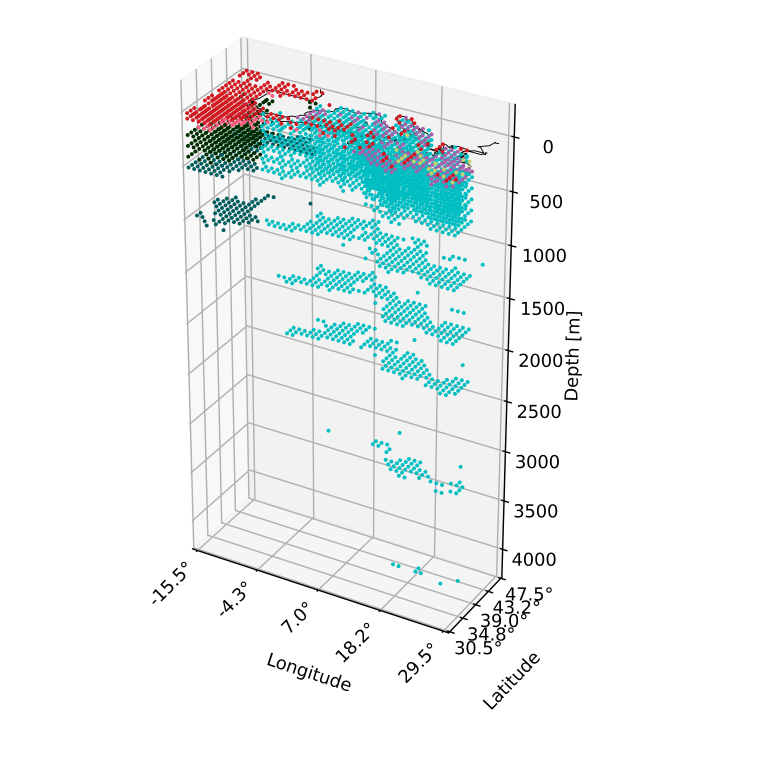} 
    \end{subfigure} 
    \begin{subfigure}[c]{0.3\textwidth}
        \centering
        \includegraphics[width=\textwidth]{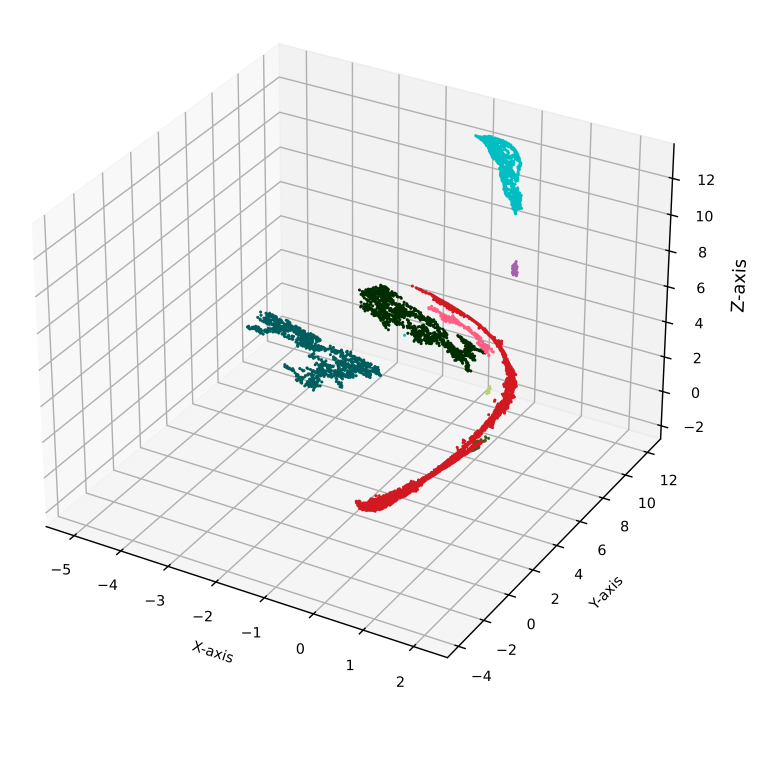} 
    \end{subfigure} 
    \begin{subfigure}[c]{0.3\textwidth}
        \centering
        \includegraphics[width=\textwidth]{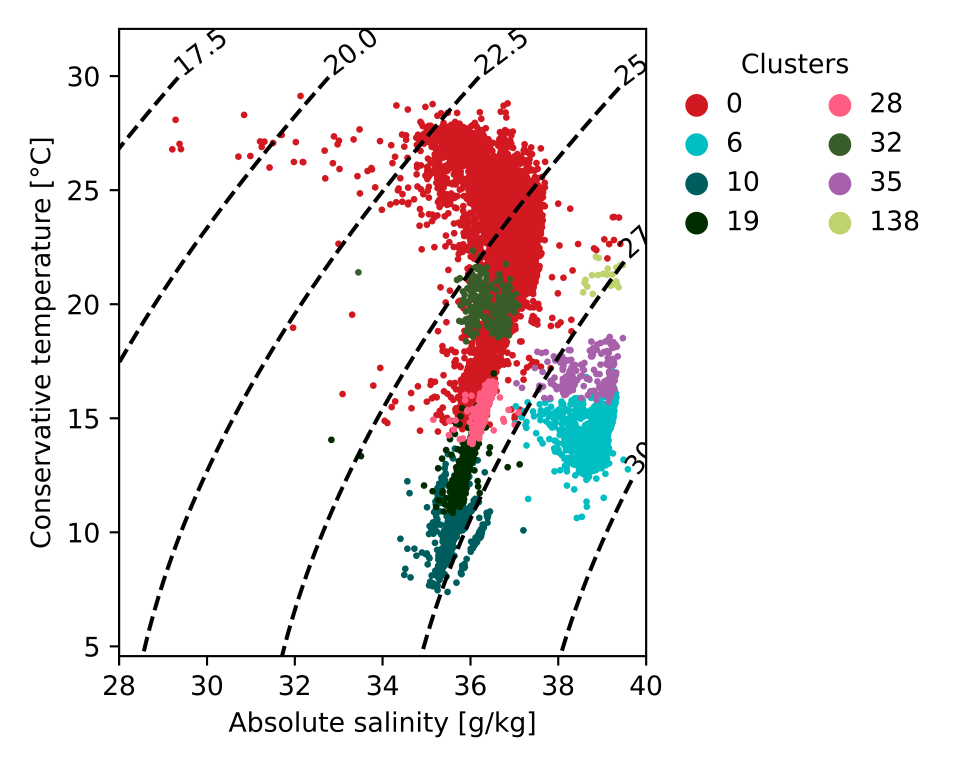} 
    \end{subfigure}

    \caption{Selected Mediterranean Sea clusters (labels 6, 32, 35, 138) and related in-/outflow regions in the North Atlantic (labels 0, 10, 19, 28) in geographic (left, zoomed into the geographic area of the Mediterranean Sea), embedded (middle) and temperature-salinity (TS) space (right).}
    \label{fig:medi}
\end{figure}

In the presented regionalisation, the Mediterranean Sea, delineated by longitude $\in [-6, 30]$ and latitude $\in [30, 48]$ omitting the Black Sea and Bay of Biscay, had a total of 2,509 grid cells (Fig.~\ref{fig:medi}). The region was subdivided into one large cluster occupying \SI{83}{\percent} of all cells (label 6), the second largest cluster covering \SI{7}{\percent} of the grid cells (label 35) and 25 smaller clusters ($<80$ cells) that were mainly found at the surface. Only one larger clusters extended downwards until \SI{100}{\meter} (label 35, eastern basin).

The main Mediterranean clusters (labels 6, 32, 35, 138) were partly related to the North Atlantic representing the Mediterranean in- and outflow (labels 0, 10, 19, 28) at the Strait of Gibraltar. In TS space, the division between the Mediterranean Sea and North Atlantic waters was clearly visible around the 27-isopycnal. In embedding space, it was even more pronounced by the large distance between the main Mediterranean cluster (label 6) and the remaining data structure. The clusters representing the Mediterranean outflow were sorted by increasing depth in embedding space. Highest uncertainties were found in this outflow area (Fig.~S11), while the remaining clusters of the Mediterranean Sea had low uncertainty.

\subsubsection{Deep Atlantic Waters}
\begin{figure}[ht]
    \centering
    \begin{subfigure}[c]{0.3\textwidth}
        \centering
        \includegraphics[width=\textwidth]{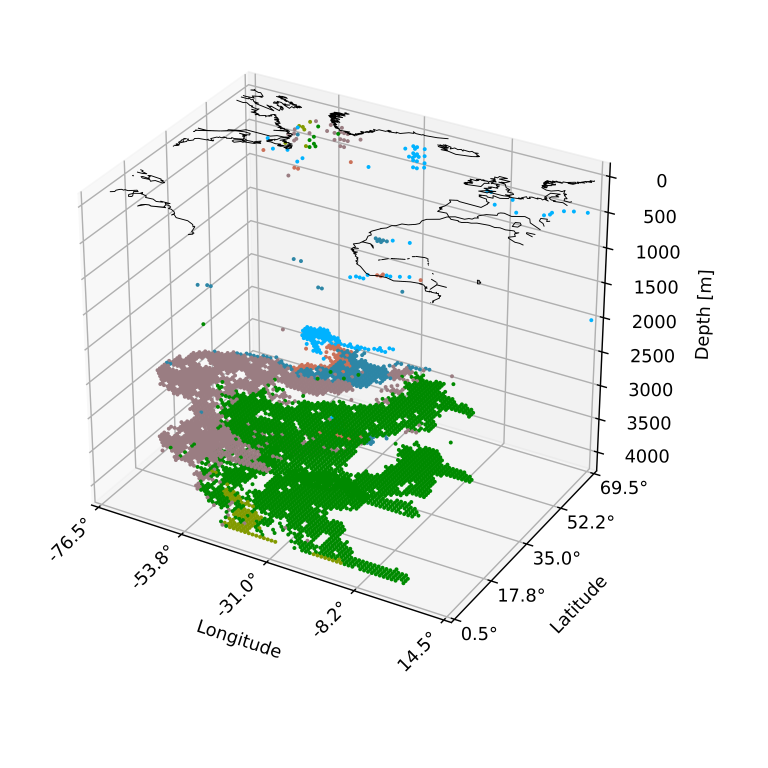}
    \end{subfigure} 
    \begin{subfigure}[c]{0.3\textwidth}
        \centering
        \includegraphics[width=\textwidth]{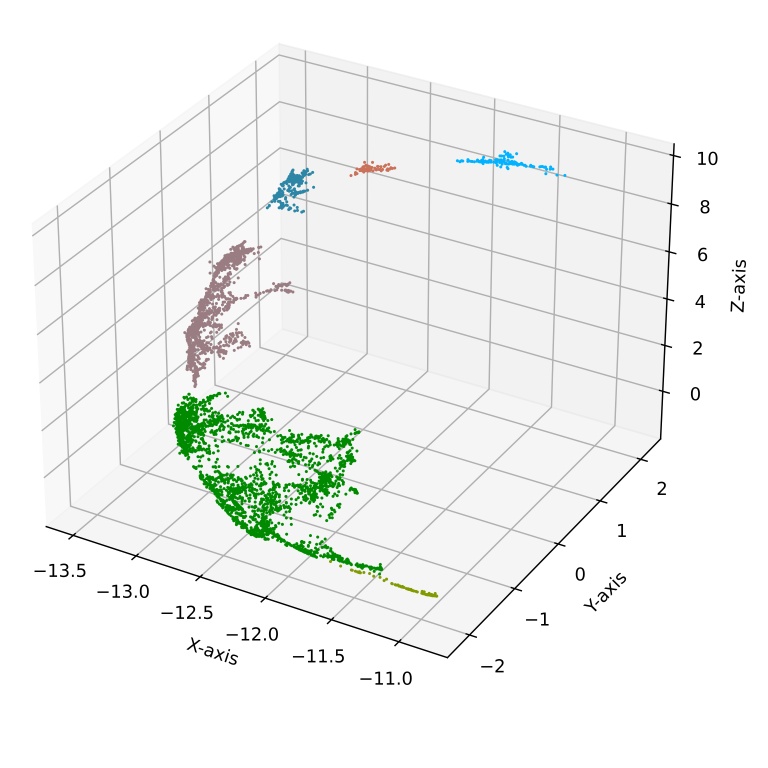}
    \end{subfigure} 
    \begin{subfigure}[c]{0.3\textwidth}
        \centering
        \includegraphics[width=\textwidth]{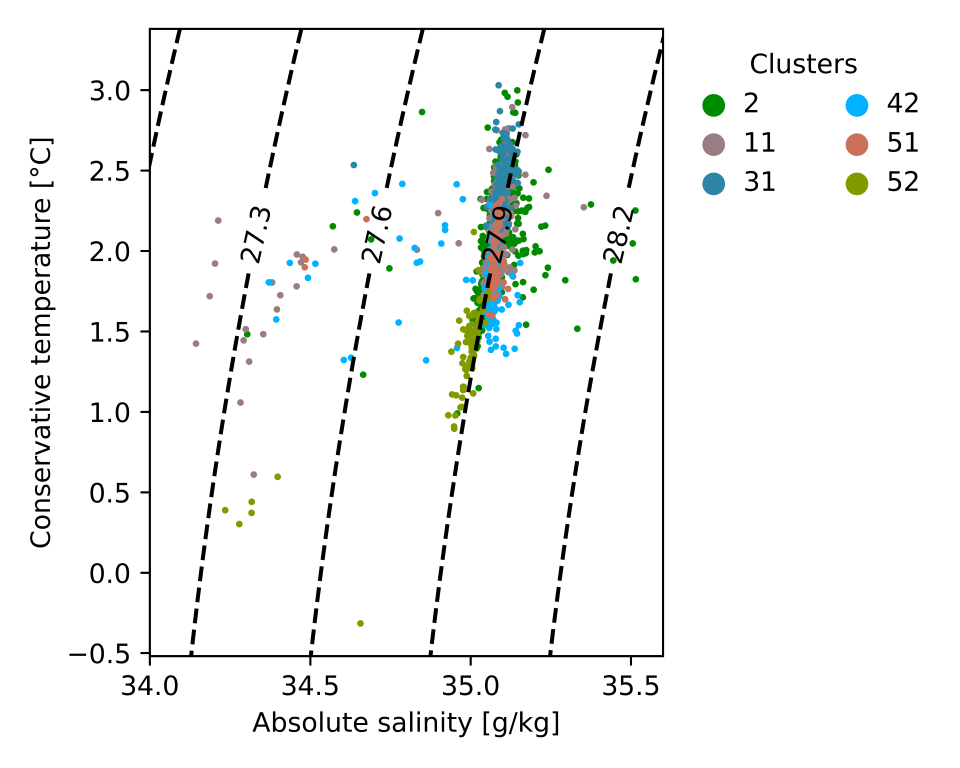}
    \end{subfigure}

    \caption{Selected clusters in the deep North Atlantic (labels 2, 11, 31, 42, 51, 52) in geographic (left), embedded (middle) and temperature-salinity (TS) space (right). The clusters were indistinguishable in TS space (except label 50 in the south), but well separable in embedding space. In geographic space, a division along the Mid-Atlantic ridge was visible.}
    \label{fig:deep}
\end{figure}

The deep waters of the North Atlantic extended from \SI{3,000}{\meter} to \SI{5,000}{\meter} stretching over the complete range of considered latitudes and longitudes excluding the Mediterranean Sea. The presented clustering detected 43 clusters (Fig.~\ref{fig:deep}), two of which accounted for the assignment of \SI{79}{\percent} of the grid cells in that area (labels 2 and 11). 38 clusters had less than 100 cells. Despite having been generally separated by the North Atlantic Ridge, the eastern water mass was also present at the west side of the ridge between 10 and \SI{30}{\degree}. In TS space, these water masses were not distinguishable. 
For further analysis, six clusters (labels 2, 11, 31, 42, 51, 52) were chosen that had uncertainties below \SI{10}{\percent}, except one in the south-west (\SI{27}{\percent}, label 52, Fig.~S12).
 
\begin{figure}[ht]
    \centering
    \includegraphics[width=0.75\textwidth]{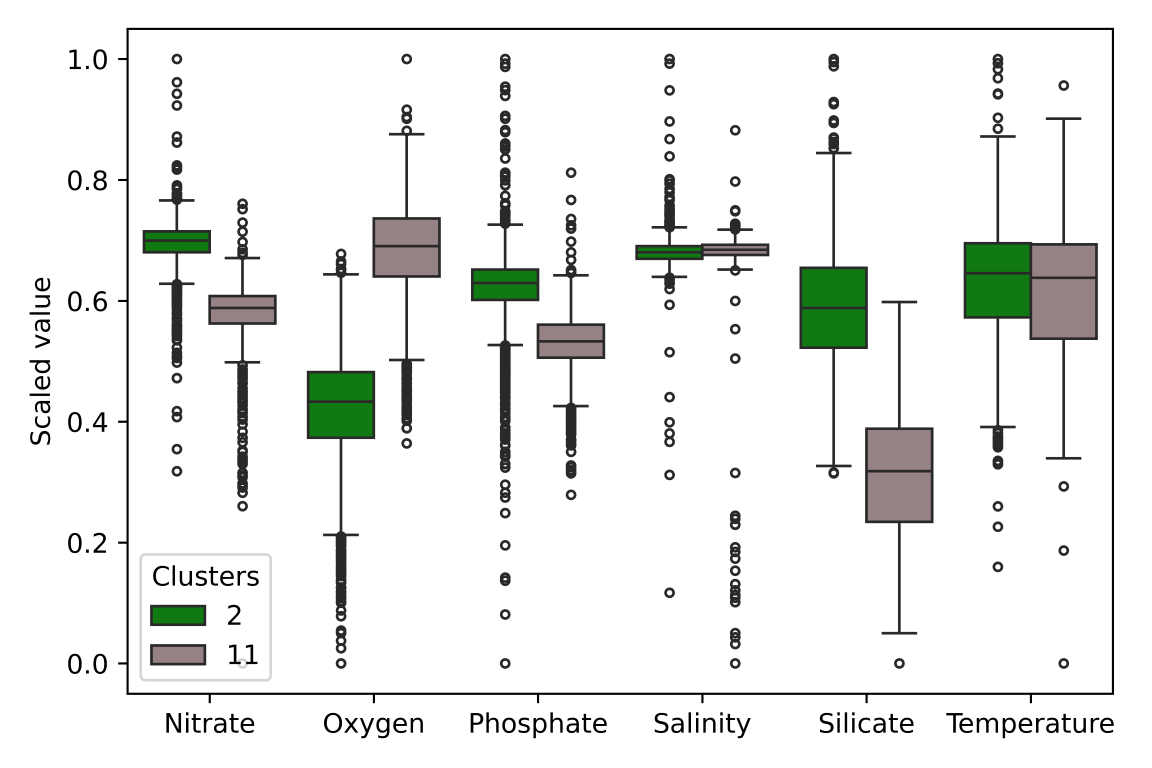}
    \caption{Comparison of parameter statistics of the eastern (label 2) and western deep Atlantic (label 11). Note that the western region had lower nutrient and higher oxygen concentration. The per-parameter differences were statistically significant (p\_value$=0$, Mann-Whitney U test, Python library scipy, version 1.11.4).}
    \label{fig:deep_params}
\end{figure}

The east-west separation (label 2 and 11 in Fig.~\ref{fig:deep}) was not clear in TS space but very pronounced in embedding space (which also considered oxygen and nutrients). Taking a closer look at parameter distributions (Fig.~\ref{fig:deep_params}) revealed that the western region has notably less silicate and more oxygen as well as lower nitrate and phosphate values. 

\subsubsection{Labrador Sea and Davis Strait}
\begin{figure}[ht]
    \centering
    \begin{subfigure}[c]{0.3\textwidth}
        \centering
        \includegraphics[width=\textwidth]{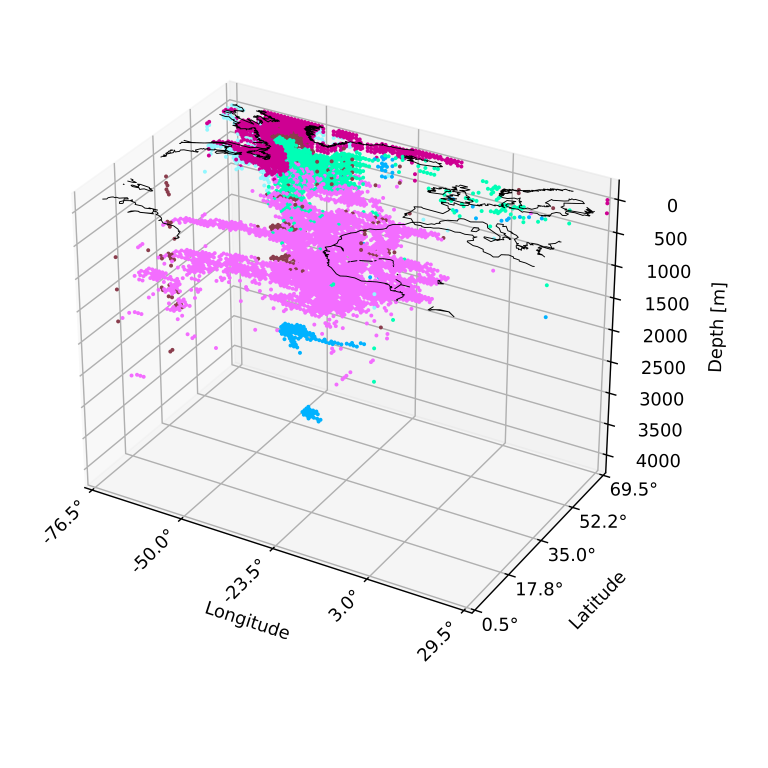} 
    \end{subfigure} 
    \begin{subfigure}[c]{0.3\textwidth}
        \centering
        \includegraphics[width=\textwidth]{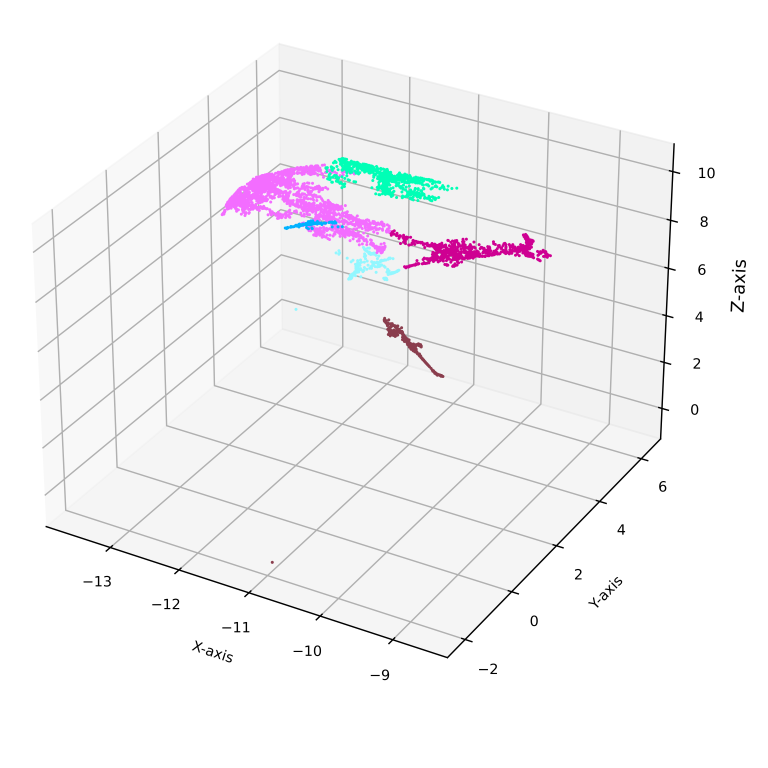} 
    \end{subfigure} 
    \begin{subfigure}[c]{0.3\textwidth}
        \centering
        \includegraphics[width=\textwidth]{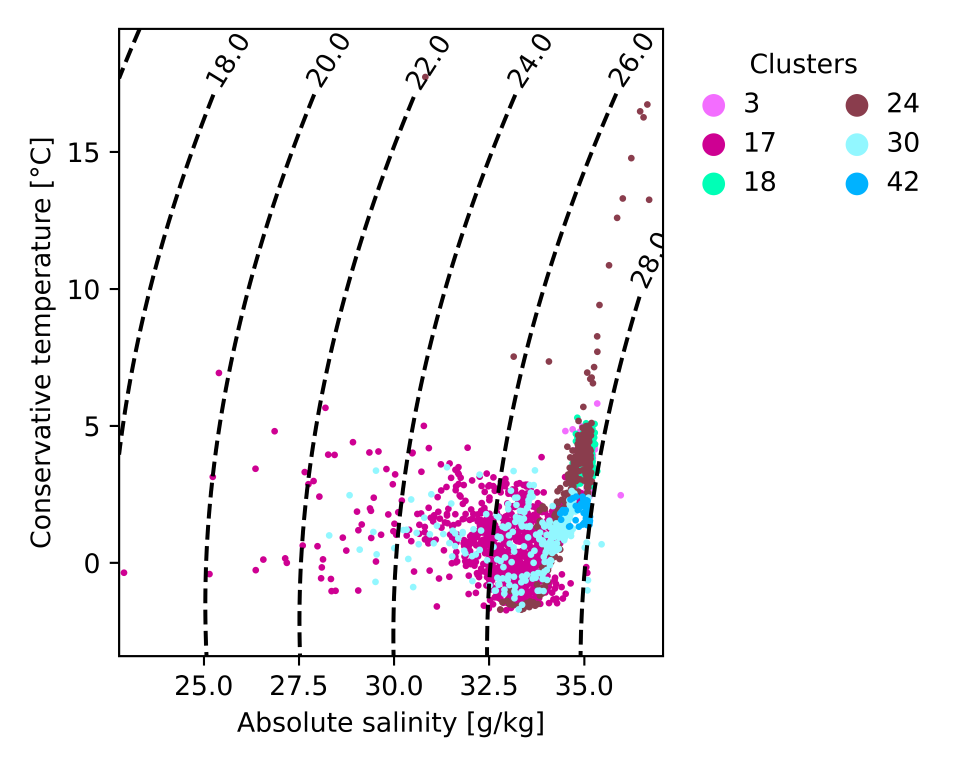} 
    \end{subfigure}

    \caption{Selected clusters in the Labrador Sea and Davis Strait (labels 3, 17, 18, 24, 30, 42) in geographic (left), embedded (middle) and temperature-salinity (TS) space (right).}
    \label{fig:lab}
\end{figure}

The Labrador Sea is located between the Labrador Peninsula and Greenland. The Davis Strait between 60 and \SI{70}{\degree}N connects it with Baffin Bay forming a shallower water pathway. 4,170 grid cells were located between 45 and \SI{70}{\degree}N and 40 and \SI{77}{\degree}W, which were assigned to 124 different clusters. The six largest clusters (labels 3, 17, 18, 24, 30, 42, Fig.~\ref{fig:lab}) were analysed, each comprising more than 100 grid cells and the largest having 780 cells (label 17).

The clusters exhibited a strong vertical structure. One cluster (label 17) formed a wide stretch around the surface coasts vanishing at about \SI{300}{\meter}, another one appeared centrally in the Labrador Sea at \SI{50}{\meter} depth stretching down to \SI{1,000}{\meter} (label 18). It borders the largest cluster (label 3), which reaches until \SI{3,000}{\meter} where it is replaced until the bottom by another cluster (label 42). Cluster label 30 filled Baffin Bay from 100 to \SI{400}{\meter} and was separated from the central cluster (label 18) by an intermediate region (label 24). The latter reached form the surface until \SI{3,000}{\meter}.

The cluster assignment uncertainty (Fig.~S13) in this region was on average $10.73\pm15.4\%$, the median was \SI{4}{\percent} indicating that uncertainty was inclined towards the lower end. Uncertainties mainly increased along the edges of the embedded data structure, which corresponded geographically to the central area (0 -- \SI{500}{\meter}) and an area along the American east coast (\SI{1,500}{\meter}).

% Discussion ----------------------------------------------
\section{Discussion}
\subsection{Comparison of Clustering Algorithm Performance}
\label{sec:discussion:clustering_methods}
% Why DBSCAN on UMAP was the best solution
DBSCAN applied to embedded space created through UMAP best suited the input data. It outperformed k-Means, agglomerative Ward clustering and DBSCAN on original data, as seen in external validation, specifically the visualisation in geographic and embedding spaces. K-Means and (to a smaller extent) Ward clustering were not able to distinguish small data structures in the  embedded space, where clusters were non-separated or merged (Section~\ref{sec:results:clustering_results}). \add{We assess that t}he main reason for the superiority of DBSCAN is its ability to detect clusters of any shape since it operates on data density \citep{ahmad2015}, i.e., it identifies areas where points are concentrated. Also, DBSCAN proved to work well in a similar use case, where it is applied to a dimensionality-reduced data space using t-Stochastic Neighbour Embedding (t-SNE) \citep{sonnewald2020}. Clusters with varying densities pose challenges for DBSCAN \citep{ahmad2015}, which may explain the occurrence of \change{\textit{confetti}}{small clusters}. Erroneous data could also contribute to this issue. 
\add{A key result was that CVIs based on similarity, density and neigbourhood structures were inconsistent and thus not helpful. This has wide implication for studies relying on only one or a few CVIs as is common in the geosciences.}

% DBSCAN on original data failed
DBSCAN on original data led to unsatisfactory results, as none of the tested hyperparameter combinations yielded a clustering that reflected the actual data structure, neither in embedded nor in geographic space. The focus on the Baltic Sea, whose salinity levels are well below oceanic average \citep{tomczak2003}, is consistent with the feature importances revealing salinity (\SI{35}{\percent}) as the most important feature followed by silicate (\SI{19}{\percent}).

% K-Means on original data
K-Means is a common and fast clustering method and therefore used as a baseline in this work. However, k-Means was not able to reflect the data structure. When applied to the original data, the CVIs agreed on two as the optimal number of clusters. However, the scores cannot be computed for less than two clusters and there is no clustering for less than two clusters. Hence, this result indicated that either (i) two clusters was indeed the best number of clusters or (ii) k-Means could not separate the data structure into relevant clusters or (iii) the scores were not meaningful. For only two clusters, the clustering is mainly a trivial hot-cold separation as seen by the high feature importance of temperature. Mapping the clusters into the embedded space highlighted how k-Means operates, i.e., drawing straight cuts through data structures. Further visual investigation with higher numbers of clusters revealed a tendency to form globular clusters and no increase in clustering quality, i.e., better representation of the embedded data structure.

% K-Means on UMAP
When applying k-Means to the embedding, the CVIs reached their optimum beyond two clusters. Compared to clustering on the original data, this indicated that k-Means was now better able to detect structures that were also recognized by the scores. 
However, the scores did not agree: While SH reached its optimum for eight and DB for ten clusters, the DB rose beyond the selected value range. 
This disagreement emphasised the need for selecting scores carefully and consulting not only one but multiple scores. Still, the clustering exhibited obvious flaws as it could e.g. not distinguish clearly separate structures in embedding space, such as the deep water. Again, a likely explanation is that k-Means detects spherical clusters best \citep{ahmad2015, jain2010, harris2022}, which were not observed in the embedded space. This clearly indicated that k-Means was not appropriate for the data and might not be for other non-linear use cases either, which are frequent in environmental sciences. Moreover, k-Means is sensitive to the initialisation of cluster centroids \citep{jain2010, harris2022} introducing a variance not investigated in this study. 

% Ward clustering
In contrast to k-Means and DBSCAN, hierarchical clustering provides a complete hierarchy of clusters. In this work, agglomerative Ward detected the overall data structures in embedding space well (Section~\ref{sec:results:ward}) but neglected smaller data point collections. Ward linkage is mathematically close to a k-Means algorithm in a hierarchical context since both methods try to minimise the same objective function, namely the within-cluster-sum-of-squares \citep{murtagh2014}. It is therefore reasonable that they also had similar score curves. 

% NEMI variability
\add{Within the NEMI framework, the combination of 100 ensemble runs showed sensitivity to the $base\_id$ parameter.} 
However, initial experiments showed that the standard deviation of mean uncertainties across runs with different $base\_id$s was only \SI{1}{\percent}. 
An alternative to NEMI for combining cluster sets from an ensemble are averages over the proximity matrices (whose $ij$-entry is one if $i$-th and $j$-th point are in the same cluster, else zero) of a UMAP-clustering pipeline \citep{bollon2022}. This average is then partitioned using spectral clustering. 

% Spatial coherence
\add{In the final DBSCAN clustering, some clusters were not geographically contiguous and appear as geographically disjoint regions with similar water mass properties. While spatial coherence can be desirable for certain applications, first experiments with a spatially constrained Ward clustering (using a 52-nearest-neighbour graph) showed that enforcing geographic continuity led to lower similarity with established oceanic classifications such as the EMUs} \citep{sayre2017}\add{ and Longhurst provinces }\citep{longhurst2007}\add{, quantified by NMI and ARI. This reveals the trade-off between feature homogeneity and geospatial contiguity (cf. e.g. }\cite{yuan2015, wang2024}\add{). Spatial constraints may obscure meaningful biogeochemical patterns. Not enforcing spatial constraints, on the other hand, may reveal physically or biogeochemically similar water masses across distant regions or alternatively, highlight limitations in the feature set’s ability to distinguish regional differences. Future work may investigate this aspect in more detail, e.g. by a binarised spatially-constrained spectral clustering as suggested by }\cite{yuan2015}.

% Scaling
\add{In this study, min-max scaling was applied prior to embedding and clustering to ensure equal contributions of features with varying ranges. Since the distribution of data used in this work did contain skew, a robust scaler that is less affected by outliers may be preferable. Downstream evaluation using UMAP quality metrics (Qlocal, Qglobal, trustworthiness and continuity) revealed that min-max scaling consistently outperformed robust scaling (RobustScaler, Python library scikit-learn, version 1.5) across multiple hyperparameter settings. 
This effect may be explained by UMAP relying by default on a distance metric to construct its neighbourhood graph }\citep{mcinnes2018}\add{ rendering the method sensitive to how features are scaled. When extreme values represent meaningful physical or biogeochemical conditions rather than noise, scaling methods that preserve the full value range, such as min-max scaling, may therefore help maintain relevant relationships.} 

% Provisional choices
\change{For data preparation, further research is required especially for the imputation.}{A focus of future research is the data preparation, especially with regards to the imputation.} Also, density was set to a constant value of \SI{1.025}{\kilo\gram\per\cubic\meter} for unit conversions as suggested previously \citep{comfortReport2021} when temperature or salinity values are unavailable. Feature importances computed by random forests were only used to get a first impression on the influence of parameters on cluster sets. For a thorough analysis, the models require further tuning and validation.

\subsection{The Importance of UMAP Embedding for Clustering Performance}
% DR and clustering
\change{A}{A key result was that a}ll clustering algorithms performed better when applied to the embedding despite the relatively low dimensional original data space, confirming previous findings \citep{allaoui2020, herrmann2023, sonnewald2020}. Additionally, the embedded data shows a more balanced feature contribution across all parameters (Fig.~S14), reflecting a potentially less biased clustering result. This may lead to clusters that are influenced by multiple factors rather than being driven by a single dominant feature. A possible reason for the performance gain using UMAP is that it enhances separability and thus the ability to cluster data \citep{herrmann2023}. Moreover, working on a space with fewer dimensions accelerates computation and optimises memory consumption of the given down-stream clustering tasks, and supports external validation by visualisation in the 3D space, where the six original dimensions could not have been used. 
\add{As noted above, relying on CVIs would have resulted in misleading clusters that did not fairly represent the data}. Working on the embedding helps to overcome the “curse of dimensionality” \citep{ayesha2020} that encompasses all phenomena related to higher dimensional data resulting in challenges for learning algorithms. For example, the amount of necessary training data grows exponentially with the number of dimensions to prevent overfitting. Also, Euclidean distances, which are used in all three clustering algorithms, become less discriminative in higher dimensional spaces \citep{verleysen2005}. 

% DR tuning
\add{Similar to clustering, dimensionality reduction is an unsupervised task with hyperparameters that need to be tuned using internal and/or external validation. In this study, embedding quality was externally evaluated based on visual clusterability, i.e. how clearly distinct and compact the clusters appear in the 3D embedded space and by assessing if the uncertainty of UMAP over multiple runs was within acceptable bounds. 
Alternatively, hyperparameters can be optimised using various internal metrics. Here, optimising for Qlocal resulted in a more dispersed embedding and notably impaired subsequent cluster sets. This suggests a potential trade-off between faithful Euclidean structure preservation and clear cluster formation. The discrepancy may also be explained by misaligned objectives: While the Q-metrics optimise distance-based rank preservation, UMAP optimises information preservation using cross-entropy. Another approach to tune UMAP hyperparameters would be to use CVIs as proxies for clusterability (e.g. }
\cite{jouilili2024}
\add{). However, performing a joint grid search over UMAP and clustering parameters increases computational cost substantially and CVIs are not inherently reliable; they may not consistently reflect meaningful structure or lead to improved embedding or clustering outcomes} (see Section~\ref{sec:discussion:cvis}). 
\add{The final embedding preserved global structure well, as indicated by a high Qglobal, and also achieved high trustworthiness and continuity, suggesting good preservation of local neighbourhood membership. A lower Qlocal, however, indicated that the fine-grained local neighbourhood rank ordering was more distorted. For the given clustering tasks, this level of distortion is acceptable, as precise ranks are typically less critical than maintaining broader neighbourhood consistency.}
\add{This study used UMAP's default Euclidean distance. Comparative tests with cosine, Mahalanobis, Chebyshev and Manhattan distances based on Qlocal and Qglobal supported this choice. This is likely because Euclidean distances preserve meaningful absolute differences across the scaled oceanographic features and align well with UMAP’s local neighborhood assumptions.}

% DR method choice
Due to the non-linear nature of the used environmental data, linear dimensionality reduction techniques were discarded such as Principal Component Analysis (PCA), Linear Discriminant Analysis (LDA), Singular Value Decomposition (SVD), Latent Semantic Analysis (LSA), Locality Preserving Projections (LPP), Independent Component Analysis (ICA) and Project Pursuit (PP) \citep{nanga2021}. 
Popular non-linear methods include Kernel Principal Component Analysis (KPCA), Multi-Dimensional Scaling (MDS), Isomap, Locally Linear Embedding (LLE), Self-Organizing Map (SOM), Latent Vector Quantization (LVQ), t-Stochastic Neighbour Embedding (t-SNE) and Uniform Manifold Approximation and Projection (UMAP) \citep{nanga2021}. 
With the perspective to apply the method to a larger dataset in the future (e.g. by considering a larger geographic area or time as an additional dimension), slow dimensionality reduction methods that do not scale well were omitted (KPCA, ISOMAP, LVQ, t-SNE) \citep{nanga2021}. 
\add{Further informing our choice, }
SOMs are computationally demanding \citep{vesanto2000} and MDS and LLE are sensitive to noise \citep{nanga2021} that could be present in the input data. Therefore, UMAP was favoured because it is able to preserve non-linear structures and to scale well.\add{ As noted in Section~}\ref{sec:results:umap}\add{ and detailed in Supplementary Material B.1, the cross-entropy method UMAP uses for optimisation was also seen as highly advantageous through its ability to strengthen associations between the data, which facilitated subsequent clustering.} 

\subsection{Choice and Interpretability of Cluster Validity Indices (CVIs)} 
\label{sec:discussion:cvis}
% CVIs to compare clustering algorithms
CVIs for comparing clustering methods \change{did not agree  amongst themselves or}{showed limited agreement} with external validation. 
Despite clearly being the best choice, DBSCAN received \change{the worst CVIs for five out of six experiments. The exception was}{worst ranks according to the classical CVIs (CH, DB and SH) and CVNNH, except for} SH, which assigned DBSCAN on original data highest rank despite it being the visually poorest subdivision. \change{In four out of six cases, k-Means was classified as the best solution.}{These scores clearly favoured k-Means and Ward on the original data, reflecting their bias towards globular, convex clusters (details on convexity in Supplementary Material B.3).}
\add{Conversely, CDR preferred DBSCAN for clustering original and embedded data, likely due to its different notion of high clustering quality, defined by small local density variations better captured by DBSCAN. }
\remove{All three CVIs used in this study favour clusters of convex shape, a precondition shared by k-Means. It tries to subdivide the data structure into Gaussian regions corresponding to spheres in 3D. Agglomerative Ward is based on an objective function closely related to k-Means, which may explain their similar performance on the clustering and on the scores.}
\remove{In the used dataset, few convex and many connected clusters with varying shapes were visually found in embedded space questioning the expressiveness of the CVIs for hyperparameter tuning. }Since \change{the scores favour a specific type of cluster structure}{each score has its own bias, i.e. imposes different assumptions on data and clusters}, they are not useful for\add{ performance} comparison \change{of}{across} clustering algorithms. \cite{thrun2021} support this by arguing that instead of selecting a clustering algorithm based on a CVI, the same result would be achieved by directly optimising for that CVI \citep{thrun2021}. \change{In practice}{Nonetheless}, the CVIs \change{have nevertheless proven useful in some cases, e.g. CH for DBSCAN tuning}{are useful for tuning hyperparameters since their bias is constant throughout the experiments.}\remove{ This suggested that the three CVIs can be useful for tuning hyperparameters in non-convex clusters, but should be selected and used with caution.}

% CVIs meaningfulness (did clustering benefit from UMAP?)
\add{Evaluating the impact of embedding with CVIs requires caution, as results did not always align with external validation. Most CVIs, including }SH, CH applied to k-Means, DB applied to DBSCAN\add{, CVNNH and CDR for all clustering methods,} \change{indicated that the clustering was}{scored} worse on the embedding than on original data, which conflicted with the previous \add{visual }finding that the clusterings benefited from the preceding dimensionality reduction. 
\add{A potential explanation may be the scale sensitivity of the CVIs (except SH and k-DBCV): UMAP inflated the maximum pairwise distanced from about 1.75 to around 25 (Fig.~S4) increasing absolute distances. This could be further evaluated in future work by scaling embedded dimensions before score computation. }
\add{K-DBCV, a scale-independent score, could not be computed for clustering original data suggesting that the original feature space lacked a clear density-based structure. After UMAP, k-DBCV scores slightly increased but remained negative indicating weak density separability. Visual inspection supported this, revealing few distinct, arbitrary-shaped clusters embedded in a more continuous structure with irregular boundaries. This aligns with SH deteriorating post-embedding since it is incompatible with non-convex geometry.}

 % CVI selection and hyperparameter tuning
\add{These results highlight the need for careful CVI selection and interpretation. }
The choice of CVIs \remove{is not trivial and }depends on the context and structure of the data\add{ and the nature of clusters, as each index offers a unique perspective on the clustering. For example, despite assuming convexity, CH did reflect DBSCAN cluster quality to some extent, indicating some flexibility of the scores.}
Generally, \cite{arbelaitz2013} found that \add{presence of }overlapping clusters or \remove{presence of }noise significantly \change{deteriorated}{impaired} performance of the 30 CVIs they investigated.\add{ Another impact factor is cluster shape: Some CVIs are more suitable for globular clusters, while others are better equipped to handle arbitrarily shaped clusters, like DBCV or CDR }\citep{schlake2024}. Other scores could be tested to tune hyperparameters of the clustering methods, like the Dunn index \citep{dunn1973}, WB index (a weighted ratio of sum-of-squares within and sum-of-squares between clusters) \citep{zhao2014}, I index \citep{maulik2002}, Cluster Validity index based on Density-involved Distance (CVDD) \citep{hu2019} or Distance-based Separability Index (DSI) \citep{guan2020}. Each has its own mathematical assumptions about the data \citep{thrun2021}. CVDD e.g. claims that it can deal with both spherical and non-spherical clusters. To assess ecological similarity, indices such as the Jaccard and Bray-Curtis indices have been utilised (e.g. \cite{carteron2012, sonnewald2020}). 

\subsection{Uncertainty Quantification}
The uncertainty and reproducibility of the best-performing clustering method (UMAP-DBSCAN) was evaluated using overlap (between all UMAP-DBSCAN runs) and RMSE as a measure.
DBSCAN is sensitive to the sequence of input samples \citep{tran2013}, which was here determined to be negligible (overlap: $99.99\pm0.003\%$). 
UMAP uses randomness as it implements stochastic gradient descent for an efficient optimization \citep{mcinnes2018}. 
With an average RMSE between the data points of the 100 embeddings of 0.22, or \SI{1.3}{\percent} of the value range, the procedure was assessed reproducible on the given data. \add{Consistency across multiple runs is supported by low standard deviations of the four computed dimensionality reduction scores.}
The combination of UMAP followed by DBSCAN had a mean overlap of $88.81\pm1.8\%$, corresponding to about \SI{11}{\percent} uncertainty. \add{Besides this high point-level compliance, the cluster sets also exhibited strong consistency in information content, as reflected by the high NMI (}$0.91\pm0.01$\add{). The ARI (}$0.78\pm0.05$\add{) indicates some variability, confirmed by the grid cell-wise uncertainty (}
\remove{Hence, the clusterings were very similar with minor differences. The grid cell-wise uncertainty amounted to a slightly worse value of}$15.49\pm20\%$\add{), which is likely caused by the sensitivity of the pipeline to UMAP}. 
\change{In summary, both, UMAP and DBSCAN, yielded robust and reproducible results alone and in combination.}{In summary, both, UMAP and DBSCAN, yielded robust and reproducible results, both individually and in combination, with only minor variations in the latter.}

Uncertainty can be a factor for deciding on a clustering method since a reproducible clustering is often desired. 
Variance of k-Means and agglomerative Ward was not further investigated, though both methods can differ over multiple runs \citep{harris2022, gordon1987}. 

\subsection{Relevance for Ecological Interpretations}
\begin{figure}[ht]
    \centering
    \begin{subfigure}[t]{0.3\textwidth}
        \centering
        \includegraphics[width=\textwidth]{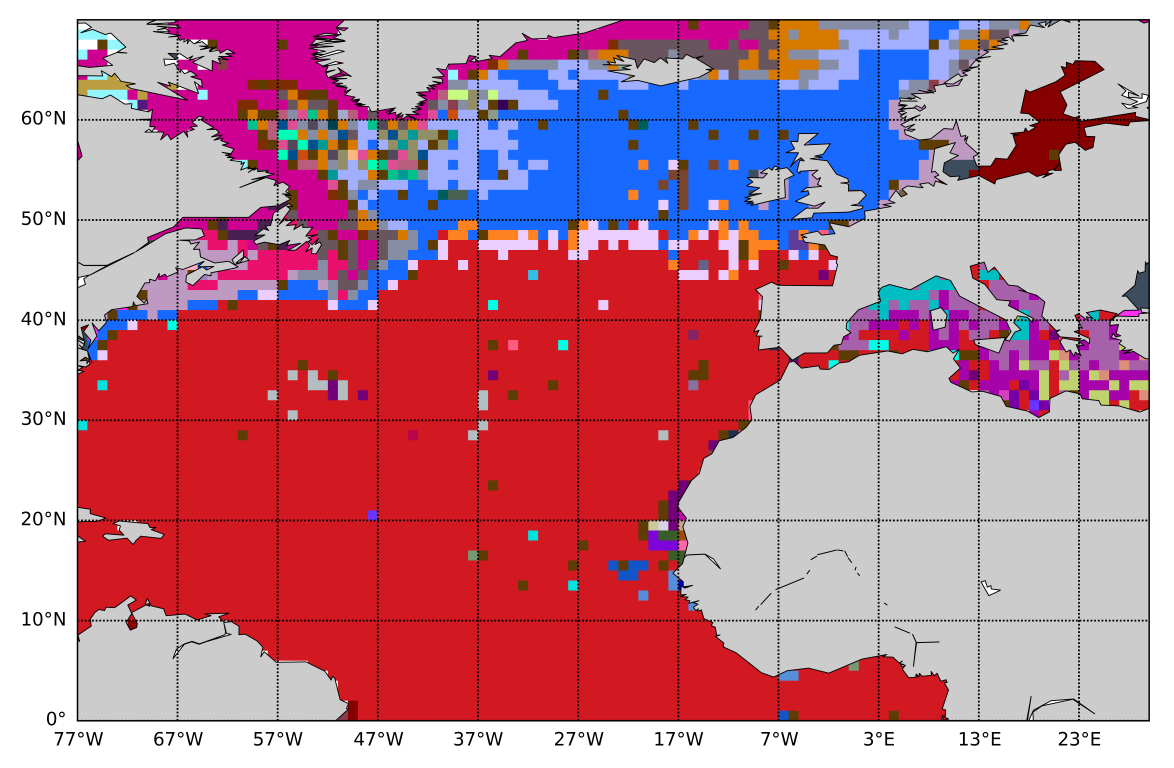}
        \label{fig:surface_this}
    \end{subfigure}
    %\hspace{0.05\textwidth}
    \begin{subfigure}[t]{0.3\textwidth}
        \centering
        \includegraphics[width=\textwidth]{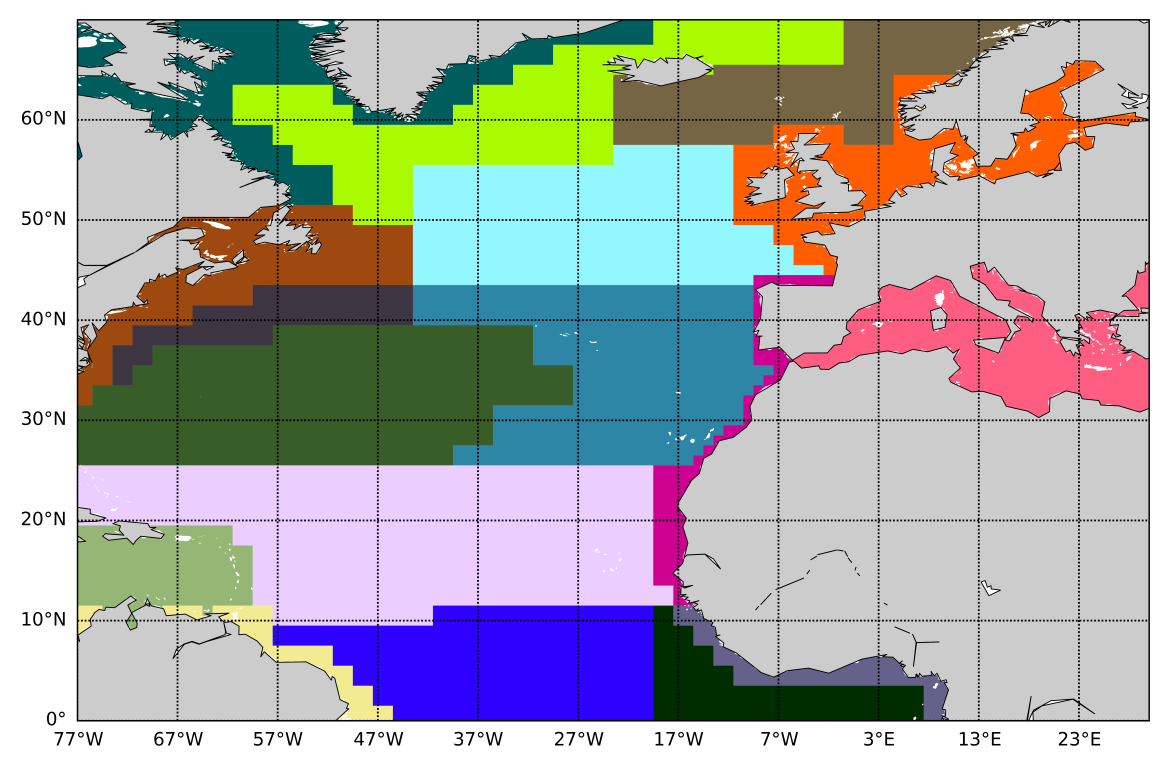}
        \label{fig:surface_longhurst}
    \end{subfigure}
    %\hspace{0.05\textwidth}
    \begin{subfigure}[t]{0.3\textwidth}
        \centering
        \includegraphics[width=\textwidth]{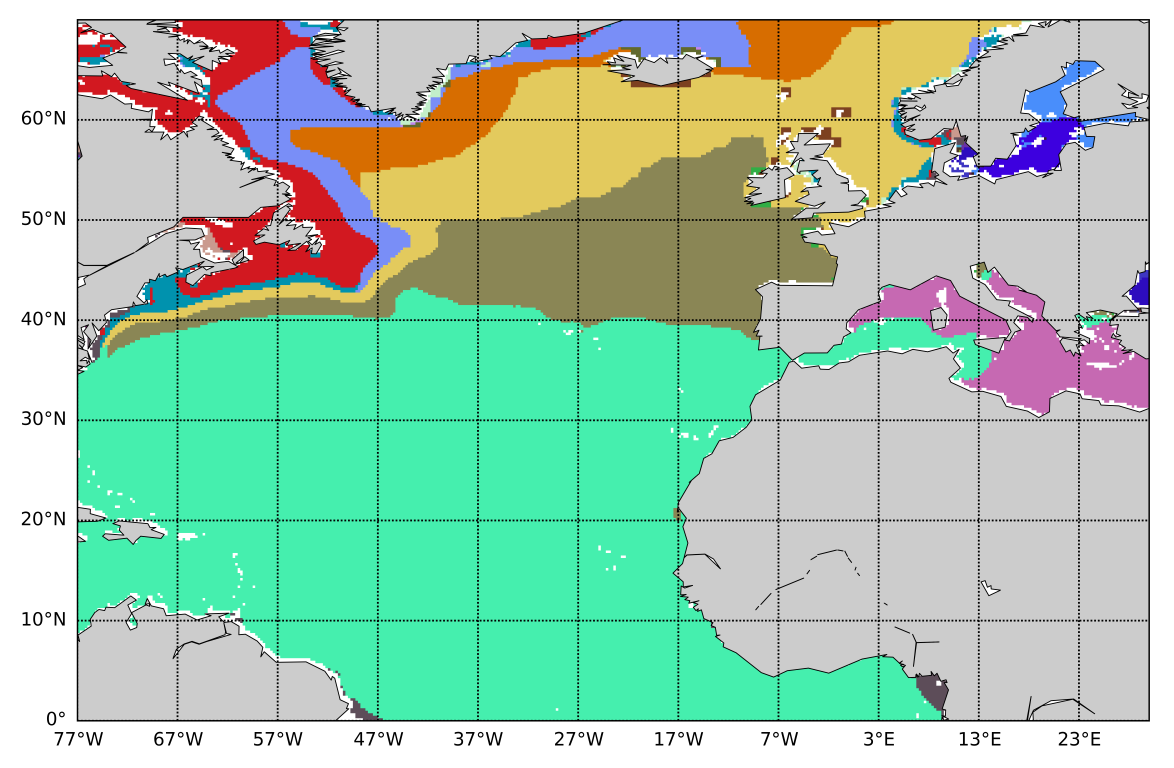}
        \label{fig:surface_sayre}
    \end{subfigure}
    
    % Overall caption for the figure
    \caption{Surface water masses as defined in this study (UMAP-DBSCAN, left), by \cite{longhurst2007} (middle) and \cite{sayre2017} (right). While similarities e.g. in the Labrador Sea stand out, there were also differences, such as \cite{longhurst2007}’s subdivision of the central North Atlantic.}
    \label{fig:surfaces}
\end{figure}

The biogeochemical clustering approach in our study has strong connections to previous approaches.
In comparison to the ecological and biogeochemical regionalisations by \cite{longhurst2007} and \cite{sayre2017}, the clustering of this study resulted in similar but more detailed clusters with some variation in the spatial extents (Figs.~\ref{fig:surfaces}\add{, S15, S16}) \add{with stronger similarities to EMUs (Tab.~}\ref{tab:literature_comparison}\add{). While moderate NMI values for both subdivisions indicate shared information content and a degree of structural correspondence, relatively low ARI values suggest larger differences in exact partitioning.}
It is obvious that the physical and biogeochemical conditions are closely connected to the characteristics of marine biomes, such as primary production (e.g. \cite{taylor2011}), microbial diversity (e.g. \cite{friedline2012}) or cycling of organic matter (e.g. \cite{koch2012, schmitt2012, hertkorn2013}). For example, the Labrador Sea showed strong similarities across the three clustering sets, though \cite{longhurst2007} suggested a coarser subdivision. 
Similarly, \cite{longhurst2007} and \cite{sayre2017} identified only one or few regions in the Mediterranean Sea, whereas this study resulted in a total of 27 clusters. 
\change{This is a result of our strictly data-driven approach, which avoids an arbitrary delineation guided by expert knowledge. }
{This higher number of regions may arise from the presented method being data-driven in contrast to Longhurst's knowledge-guided approach, more sensitive to local structure and not imposing constraints on cluster number, shape, or size. Moreover, the COMFORT dataset includes the data used in }\cite{sayre2017}\add{ and incorporates additional measurements.} 
Previous studies are based on \textit{a priori} decisions on the total number of clusters - based on the data or the parameters applied for the clustering method. For example, k-Means (as used by \cite{sayre2017}) uses a predefined number of clusters and optimises for the entire dataset. This can result in smoothing over local variations, with consequences for the ecological interpretations of the regions. DBSCAN in out study does not enfore a specific number of clusters but prescribes the connectivity conditions, i.e., how far apart data points in feature space may be to form a cluster, which promotes fine-grained subdivisions. In global ecological studies, the Mediterranean is often described as one region (e.g. \cite{longhurst2007, costello2017}) though works like \cite{sayre2017, zhao2020b} and this study suggest a higher diversity especially at the surface in comparison, for example, to the North Atlantic. This is likely caused by complex upper ocean currents \citep{tomczak2003} and small-scale patterns of seasonal primary production (as represented in \url{https://www.grida.no/resources/5937}).
%\textcolor{red}{TODO BETTER REFERENCE}). 
A short analysis of the proportion of endemic species per region using OBIS data (\cite{obis2025}, data not shown) revealed \SI{12}{\percent} of occuring species in the largest and in the second largest Mediterranean clusters (labels 6, 35) to be endemic suggesting the ecological uniqueness of the main biogeochemical water masses in the Mediterranean Sea. 

An example for varying region extent in different clustering approaches is the western deep North Atlantic (cluster 11). In this study using DBSCAN and UMAP, the region extended further south along the American coast compared to the Ecological Marine Units by \cite{sayre2017}. By using k-Means, we were able to reproduce the spatial extend in the previous work. 
The visualisation in embedding space revealed that k-Means struggled to adequately separate this area. This might be attributable to the complexity of the data, visible in embedding space as an irregularly connected, curved structure, not resembling normal distributions for which k-Means is optimised.
Despite the intrinsic bias of k-Means, the clustering from \cite{sayre2017} approach and the DBSCAN clustering presented here exhibited many similarities (e.g. surface regions, Fig.~\ref{fig:surfaces}). A possible reason is that \cite{sayre2017} use a higher depth resolution: In total, 102 depth intervals were defined and the very variable first \SI{100}{\meter} of the ocean column are subdivided into \SI{5}{\meter} steps. This higher resolution might enable a better separation of data points in feature space and thus a more precise clustering. 

Generally, the presented cluster set picked up well-known oceanographic features, like the outflow of warm, saline Mediterranean water through the Strait of Gibraltar \citep{pinardi2023} that is traceable at the \SI{2,000}{\meter} level across the North Atlantic Ocean \citep{tomczak2003}. Due to the time-averaging of data, small-scale and dynamic oceanographic features such as eddies were not sufficiently represented in this cluster set. Another well-represented feature is the subdivision of deep Atlantic waters along the north-west axis. Compared to the east, the western waters were characterized by lower silicate, nitrate and phosphate and higher oxygen concentration (Fig.~\ref{fig:deep_params}). This is in good agreement with the fact that the western part of the deep North Atlantic is more influenced by relatively young North Atlantic deep water, while there is more influence of Southern Ocean deep waters in the east \citep{johnson2008}. This emphasises that while temperature and salinity remain key parameters in defining water masses, the inclusion of additional parameters such as oxygen and nutrients is crucial for a comprehensive and detailed analysis of water mass properties and dynamics in the ocean. 

The clustering results for the case study in the Labrador Sea and Davis Strait, an area of deep-water formation, were generally in very good agreement with pertinent oceanographic literature. Those clusters that represented freshly formed deep water (particularly labels 3 and 24) were characterised by high salinity, low temperature and slightly depleted oxygen values, as shown previously e.g. by \cite{tomczak2003}.
The clustering did not always yield spatially coherent clusters, for example a cluster in the deep Atlantic near the equator (label 7). Despite fairly different temperatures, the same cluster label was also assigned to water in the Labrador Sea, because of a high similarity in the inorganic nutrient concentrations. A possible reason is the exclusion of geographic coordinates  (or proximity to coasts, such as in \cite{longhurst2007}) from the clustering process or that additional parameter(s) are required for the distinction.

% Conclusion ----------------------------------------------

\section{Conclusion and Outlook}
By comparing pre-processings, clustering methods and various validation techniques, this study found a clustering that adequately reflected the embedded data structure of North Atlantic physical and biogeochemical properties. Such a methodological approach to clustering is of high importance for quality and hence potential downstream tasks. \cite{thrun2021} formulated this precisely referring to their medical use case: 

\begin{quote}
“[...] only the combination of empirical medical knowledge and an unbiased, structure-based choice of the optimal cluster analysis method w.r.t. the data will result in precise and reproducible clustering with the potential for knowledge discovery of high clinical value.” \\
--- \cite{thrun2021} 
\end{quote}

DBSCAN applied to a dimensionality-reduced space using UMAP best reflected the data structure, outperforming k-Means and agglomerative Ward. When validating the results, it was imperative to not rely on single criteria, e.g. to compute multiple CVIs for hyperparameter tuning. The presented results moreover discourage using CVIs for the comparison between clustering methods. 

For reproducibility purposes, analysis of uncertainty is an important aspect to consider when non-deterministic algorithms are applied. The variability of the presented method was quantified using ensemble analysis revealing low variabilities of the individual methods (UMAP, DBSCAN) and slight deviations in the clustering when combined (overlap of $88.81\pm1.8\%$). By combining the clustering results over the ensemble following the NEMI framework, reproducibility and representativeness of the statistical co-variance space was further increased (uncertainty of $15.49\pm20\%$).

There were several aspects that could be further explored related to data, pre-processing, method and post-processing. Regarding the data, further parameters like dissolved (in-)organic carbon or biogeochemical tracers such as Apparent Oxygen Utilisation (AOU) could be added. \add{Future work aims} to scale the clustering up to global coverage and add the temporal component \add{to increase oceanographic utility}. For this, data sparsity could be a limiting \add{factor machine learning is especially suited to overcome}. Also, other clustering methods that are able to deal with varying densities, such as HDBSCAN, are worth exploring. Further, self-organising maps \citep{kohonen1990} could increase interpretability of results by providing meaningful maps of the classes while preserving data topology \citep{yonggang2011}. To further improve performance, other hyperparameters can be explored. For agglomerative Ward, connectivity constraints could be used to e.g. enforce spatially connected clusters \citep{dalila2020}. Moreover, distance metrics other than Euclidean could be investigated for Ward and DBSCAN clustering to ensure the optimal strategy for the given data. Oceanographically, the cluster set can be compared more extensively to existing definitions to potentially extract new knowledge.

%\appendix

%\section{}\label{}

\medskip

\textbf{Acknowledgments} \\
We would like to express our gratitude to the COMFORT project group for their invaluable contribution in providing the dataset that formed the foundation of this research. We are thankful to Claire Monteleoni and Anastase Alexandre Charantonis for their insightful discussions that helped shape the direction of this study. Also, we wish to thank Marlo Bareth for being a valuable discussion partner, whose perspectives helped refine the analysis. 
YJ was funded through the Helmholtz School for Marine Data Science (MarDATA, Grant No. HIDSS-0005). This work was supported by a fellowship of the German Academic Exchange Service (DAAD), allowing for a research stay of YJ with MS at Princeton University and University of Washington, Seattle. We acknowledge support by the Open Access publication fund of Alfred-Wegener-Institut Helmholtz-Zentrum für Polar- und Meeresforschung. MS acnowledges support from the Award NA24OARX431C0058-T1-01 from the National Oceanic and Atmospheric Administration, U.S. Department of Commerce and the Award RISE-2425906 from the U.S. National Science Foundation.

\printcredits

\medskip

\textbf{Data Availability} \\
The cluster set and auxiliary information is publicly available on Zenodo (\url{https://doi.org/10.5281/zenodo.15201767}) along with the code base used to conduct and analyse the presented experiments (\url{https://doi.org/10.5281/zenodo.15202343}) 
and the dashboard to explore the final cluster set (\url{https://doi.org/10.5281/zenodo.16742243}). 
Column descriptions can be found in Supplementary Material D. 
The COMFORT dataset is publicly available online \citep{comfortDataset2022}.

% For arXive submission, include supplementary materials files
\clearpage
% Attention! The images here are below submission resolution!

{\Large Supplementary Material}

\appendix

% Reset counters
\setcounter{figure}{0}
\setcounter{table}{0}
\setcounter{equation}{0}

% Redefine numbering
%\renewcommand{\thesection}{S} % Section labeled as "S"
\renewcommand{\thefigure}{S\arabic{figure}} % Figures: S1, S2...
\renewcommand{\theequation}{SE\arabic{equation}}
\renewcommand{\thetable}{ST\arabic{table}}

\section{Data}
\label{s:data}
\subsection{Data Preparation}
\label{s:data_preparation}
We focused on the North Atlantic and considered a rectangular area extending from \SI{-77}{} to \SI{30}{\degree} longitude, from 0 to \SI{70}{\degree} and from 0 to \SI{5000}{\meter} depth. This area had better data coverage compared to the remaining ocean. From the COMFORT dataset \citep{comfortDataset2022}, we selected the parameters temperature, salinity, oxygen, nitrate, silicate and phosphate, which had best spatial coverage and can serve biogeochemical research questions. The six parameters globally  comprised 441,054,314 measured values.

The data was filtered for quality flags provided in the COMFORT dataset. The primary quality flag 1 (PQF1, inherited from the original data sources) was restricted to $>0$ (value was checked) and primary quality flag 2 (PQF2, additional range check) was restricted to $>2$ (acceptable value), which reduced the number of measured values to 
%313,389,032.
313,387,613.

Temperature and salinity values were averaged over identical times and locations (i.e., latitude, longitude, and depth). The averaged salinity and temperature values were joined with the other parameters based on time and location. This enabled the subsequent harmonisation of different units for oxygen, nitrate, silicate and phosphate to \SI{}{\micro\mole\per\kilo\gram} (Python implementation of Gibbs SeaWater Oceanographic Toolbox of TEOS-10; gsw, version 3.6.17). 
For the unit conversion, suggestions and formulas from \cite{comfortReport2021} were applied. If temperature and salinity values were available, they were used for density computation. Otherwise, density was set to a constant value of \SI{1.025}{\kilo\gram\per\cubic\meter} as suggested by \cite{comfortReport2021}. For the oxygen conversion from \SI{}{\percent} to \SI{}{\micro\mole\per\kilo\gram}, we leveraged formula 31 from \cite{benson1984}. This conversion required temperature and salinity values wherefore three oxygen values were discarded. Silicate, nitrate, phosphate and oxygen were averaged over time and location. Temperature was converted to potential temperature (Python library gsw, version 3.6.17) making temperature values from different depths comparable. This procedure resulted in a final set of 
%313,280,300 
306,652,251
%313,338,823 before averaging over loc/time
%313,280,300 after averaging over loc/time
%306,652,251 after conversion to pott
measured values.  

%Temperature values
%New preprocessing: 124,513,005
%Old preprocessing: 131,141,054
%Reproduced preprocessing (no pott conversion): 131,141,054

The six parameters were averaged over all available times (years 1772 -- 2020) and mapped on a grid with a spatial resolution of \SI{1}{\degree} and 12 water depth intervals: 
0 -- \SI{50}{\meter}, 
50 -- \SI{100}{\meter}, 
100 -- \SI{200}{\meter}, 
200 -- \SI{300}{\meter}, 
300 -- \SI{400}{\meter}, 
400 -- \SI{500}{\meter}, 
500 -- \SI{1000}{\meter}, 
1000 -- \SI{1500}{\meter}, 
1500 -- \SI{2000}{\meter}, 
2000 -- \SI{3000}{\meter}, 
3000 -- \SI{4000}{\meter}, 
4000 -- \SI{5000}{\meter}. 
The depth intervals were roughly aligned with definitions by \cite{sayre2017} but more coarse to achieve a lower proportion of empty grid cells. We limited the gridded data to latitude in $[0, 70]$\SI{}{\degree}, 
longitude in $[-77, 30]$\SI{}{\degree} and 
depth in $[0, 5000]$\SI{}{\meter} to achieve good spatial coverage and due to the computational expense. For measurements geographically located on a grid cell boundary, we defined the cells to include values on the lower boundary while excluding values on the upper boundary. The only exceptions were cells with maximum latitude, longitude or depth for which the upper or right boundary was included. To depict a grid cell, we used the horizontal centre and the vertical upper depth boundary, e.g. a grid cell covering $[0, 1)$\SI{}{\degree} $\times [0, 1)$\SI{}{\degree} $\times [0, 50)$\SI{}{\meter} would be labelled [0.5, 0.5, 0]. 

Grid cells on land were removed from the grid using GEBCO bathymetry information \citep{gebco2022}, resulting in a total of 
%49,131 
49,030 grid cells. GEBCO has a resolution of 15 arc seconds, which we downsampled to our \SI{1}{\degree} resolution by selecting the nearest depth information to each grid cell centre. Downsampling can lead to inaccuracies especially around the coastlines. This means that even though the cell is labelled as land, it can contain a value from a water measurement. Therefore, we only dropped land cells that had no parameter value assigned to it, which also encompassed a few data points that were located further inland.

Grid cells that had values only for a subset or none of the six parameters were filled using k-Nearest Neighbours (KNN, Python library scikit-learn, version 1.5) to achieve a completely specified grid. KNN has two hyperparameters, the number of neighbours, which was set to five and the weights parameter, which was set to “distance”. Nine features served as input for training the imputer: The six oceanographic parameters scaled using MinMaxScaler (Python library scikit-learn, version 1.5), latitude, longitude and depth scaled to $[0, 1]$. 
In total, 
%\SI{3.2}{\percent} 
\SI{3.4}{\percent} 
of the grid cells were completely empty, 
%\SI{36.8}{\percent} 
\SI{36.6}{\percent} 
were partly empty. Nitrate had the highest proportion of missing values in the grid cells 
% \SI{35.5}{\percent} 
(\SI{35.3}{\percent}), followed by silicate 
%\SI{26.6}{\percent}
(\SI{26.4}{\percent}), phosphate 
%\SI{19.9}{\percent}
(\SI{19.7}{\percent}), oxygen 
%\SI{7.9}{\percent}
(\SI{7.7}{\percent}), temperature 
%\SI{4.7}{\percent}
(\SI{5.3}{\percent}) and salinity 
%\SI{3.8}{\percent}
(\SI{3.6}{\percent}). The proportion of missing values for all measurements was 
%\SI{16.9}{\percent}.
\SI{16.4}{\percent}.

\subsection{Data Description}
\label{s:data_description}
\begin{figure}[h]
    \centering
    \phantomsection
    \includegraphics[width=\textwidth]{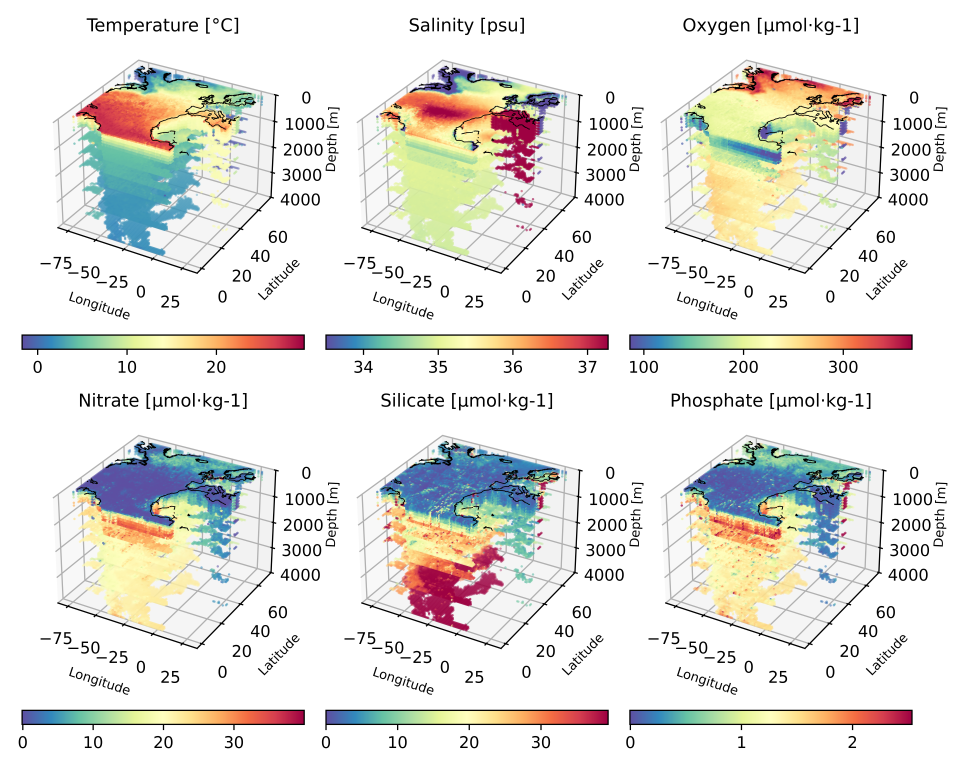}
    \caption{Spatial distribution of the \SI{1}{\degree} gridded and imputed six input parameters (temperature, salinity, oxygen, nitrate, silicate, phosphate).}
    \label{s:fig:3d}
\end{figure}

The overall average temperature of the prepared data (Fig.~\ref{s:fig:3d}, top left) was 
%\SI{9.46}{\celsius}, 
\SI{9.37}{\celsius}, 
with a minimum of 
%\SI{-1.81}{\celsius} 
\SI{-1.74}{\celsius} 
in the high latitudes and a maximum of 
%\SI{29.84}{\celsius}
\SI{29.83}{\celsius} 
at the surface of the Caribbean Sea. Temperature generally decreased with depth, with the exception of the Mediterranean Sea, where warmer waters occurred below the less saline surface layer. 

The average salinity (Fig.~\ref{s:fig:3d}, top middle) was 
%35.24, 
35.25, with highest values in the Mediterranean 
%(max: 39.5) 
(max: 39.49) and in the central area of the North Atlantic, where fluvial input and precipitation is low. Lowest values occurred at the northern coasts, which are influenced by higher precipitation and river run-off (min: 0.05). 

The average oxygen concentration (Fig.~\ref{s:fig:3d}, top right) was 
% \SI{222.5}{\micro\mole\per\kilo\gram} (min: \SI{0}{\micro\mole\per\kilo\gram}, max: \SI{419.2}{\micro\mole\per\kilo\gram}) 
\SI{222.43}{\micro\mole\per\kilo\gram} (min: \SI{0}{\micro\mole\per\kilo\gram}, max: \SI{419.2}{\micro\mole\per\kilo\gram}) 
with high values at low surface temperatures in the north (due to higher solubility and high primary production in the summer season). The Atlantic subsurface oxygen minimum zone \citep{karstensen2008} extends from the west coast of Africa towards South America (blue subsurface layer, Fig.~\ref{s:fig:3d}). 

The average nitrate concentration (Fig.~\ref{s:fig:3d}, bottom left) was \SI{13.69}{\micro\mole\per\kilo\gram} (min: \SI{0}{\micro\mole\per\kilo\gram}, max: \SI{315.43}{\micro\mole\per\kilo\gram}) and varied with depth. The surface is nitrate-poor with slightly higher values around Greenland and Iceland and a few exceptionally high values at England’s south coast and the north coasts and estuaries of France and Germany. Higher nitrate values were generally found below the surface, especially at intermediate depths between South America and Africa. 

\begin{figure}[h]
    \centering
    \includegraphics[width=0.5\textwidth]{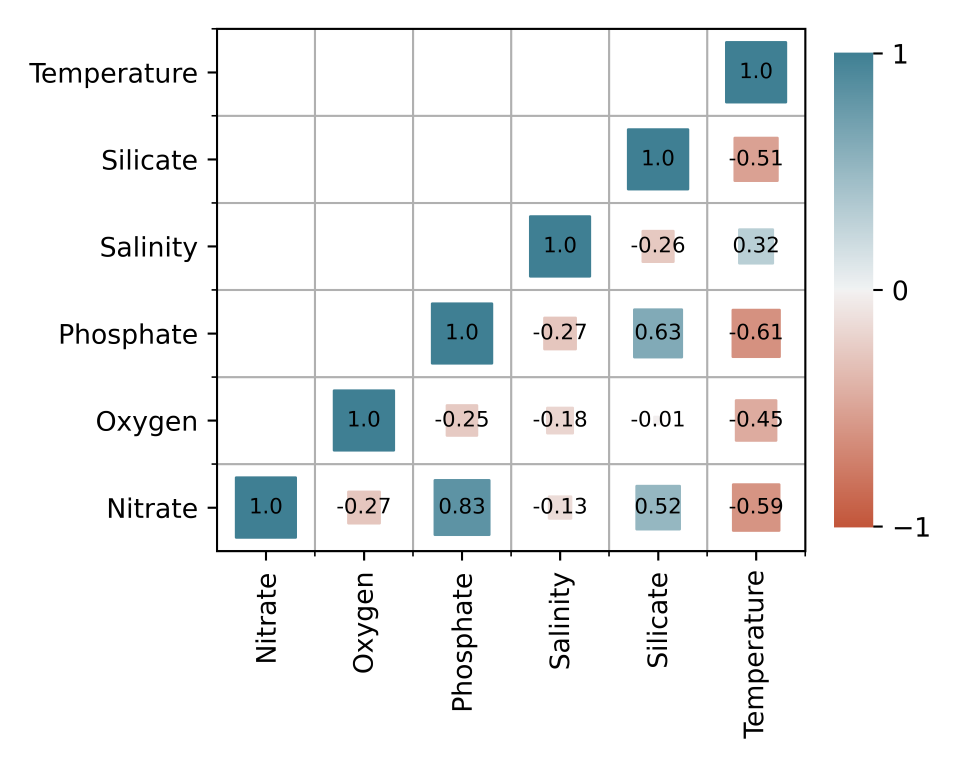}
    \caption{Pearson correlation coefficients between features. Note the moderate to strong correlation between the nutrients, as well as the moderate correlation of temperature with nutrients and oxygen.}
    \label{s:fig:correlation}
\end{figure}

Average phosphate concentrations (Fig.~\ref{s:fig:3d}, bottom right) were \SI{0.91}{\micro\mole\per\kilo\gram} (min: \SI{0}{\micro\mole\per\kilo\gram}, max: \SI{7.42}{\micro\mole\per\kilo\gram}. Exceptionally high concentrations occur at the Baltic seafloor, the Gulf of \break St. Lawrence between \SI{200}{\meter} and \SI{400}{\meter}, and in the Black Sea (from which only a small region was included in this work). Phosphate showed a strong positive linear correlation with nitrate concentration ($r = 0.83$; Fig.~\ref{s:fig:correlation}). 

Silicate (Fig.~\ref{s:fig:3d}, bottom middle) had an average concentration of 
%\SI{13.45}{\micro\mole\per\kilo\gram} 
\SI{13.44}{\micro\mole\per\kilo\gram} 
(min: \SI{0}{\micro\mole\per\kilo\gram}, max: \SI{260}{\micro\mole\per\kilo\gram}). 
The surface was characterised by low values with the main exception being the Labrador Sea and the terrestrially-influenced Baltic Sea. Similar to nitrate, values increased with depth. However, the general correlation between silicate and nitrate/phosphate was weaker ($r \in [0.5, 0.6]$; Fig.~\ref{s:fig:correlation}) compared to nitrate and phosphate. At \SI{2000}{\meter}, a large high-silicate area was present at the coasts of Western Europe and Western Africa with values increasing towards the seafloor. Another region of exceptionally high silicate values was located off the west coast of Greenland stretching from \SI{50}{\meter} down to the sea floor. 

Temperature had a Pearson correlation coefficient between \SI{-0.5}{} and \SI{-0.6}{} with the three nutrients (Fig.~\ref{s:fig:correlation}). Correlations among nutrients ranged between 0.5 and 0.8. All other Pearson coefficients were below %0.4 
0.5 highlighting that most parameters were not linearly correlated  (compare \cite{dickey2006, yang2020}).

\section{Method Details}
\label{s:method_descriptions}
\subsection{Dimensionality Reduction: Uniform Manifold Approximation and Projection (UMAP)}
\label{s:umap}
Dimensionality reduction is a technique to project data from high to lower dimensionality. It can be classified as a feature selection and feature extraction method \citep{ayesha2020}. Here we focused on feature extraction, i.e., transforming input features into a new set of features. We applied the non-linear dimensionality reduction technique Uniform Manifold Approximation and Projection (UMAP, \cite{mcinnes2018}, Python library umap-learn, version 0.5.5, \cite{umap_software}). UMAP emphasises similarities and differences between data points while keeping the topology. It first represents neighbourhoods by a fuzzy topology, i.e., a high-dimensional graph. Then a low-dimensional graph is constructed to be as similar as possible to the high-dimensional representation by optimising the cross-entropy between the two fuzzy sets (Eq.~\ref{s:eq:umap}, \cite{mcinnes2018}). The lower-dimensional representation of the data is called embedding.

\begin{equation}
    C((A, \mu), (A, \nu)) \triangleq \sum_{a \in A}\left(\mu(a)\log\left(\frac{\mu(a)}{\nu(a)}\right) + (1-\mu(a))\log\left(\frac{1-\mu(a)}{1-\nu(a)}\right) \right)
\label{s:eq:umap}
\end{equation}
\noindent where \\
\noindent $(A, \mu)$: Fuzzy set \\
\noindent $A$: Reference set \\
\noindent $\mu \mapsto [0, 1]$: Membership strength function \\

UMAP has two main hyperparameters: The number of neighbours used to compute the high-dimensional representation (\textit{n\_neighbors}) and the minimum distance between points in the low-dimensional space (\textit{min\_dist}). Fewer neighbours lead to a focus on more local structures, while higher values to a focus on more global structures. Increasing the minimum distance shifts the focus from fine to coarse topology in the target space. 

\subsection{Clustering}
A clustering is defined as a function $F$ that maps a set of $d$-dimensional data points $X=\{x_i \mid i=1, ..., n\}$ to a set of $K$ labels/clusters $\text{Cl}=\{C_k \mid k=1, ..., K\}$, namely $F: X \mapsto Cl$ \citep{jain2010}. 

\subsubsection{KMeans Clustering}
\label{s:kmeans}
KMeans was introduced independently by various researchers \citep{ikotun2023}, including \cite{steinhaus1956}, \cite{macqueen1967} and \cite{lloyd1982}.
The default implementation of KMeans (Python library scikit-learn, version 1.5) uses Lloyd’s algorithm, which minimises the within-cluster sum-of-squares and operates in three steps:
\begin{enumerate}
    \item Select $k$ cluster centroids by default using greedy KMeans++. 
    \item Assign data points to nearest centroids using Euclidean distances. 
    \item Update centroids by computing the mean of the assigned data points. 
\end{enumerate}
Steps 2-3 are repeated until the centroids do not change more than a threshold \citep{jain2010}. The main hyperparameter to tune is the number of clusters $k$.

\subsubsection{Agglomerative Clustering}
\label{s:agglomerative_clustering}
Agglomerative clustering is a type of hierarchical clustering in which the algorithm starts with each data point as a cluster. Then, it successively merges the most similar clusters until only one cluster for all data points remains. The linkage metric determines similarity and thus merging of clusters. Here, we applied Ward linkage (Eq.~\ref{s:eq:ward}). It is defined as the sum of squared distances within each cluster and distances between these values for each cluster pair \citep{ward1963}:

\begin{equation}
\Delta(C_i, C_j) = \frac{|C_i||C_j|}{|C_i| + |C_j|}\left \| \text{centroid}(C_i) - \text{centroid}(C_j) \right \|^2
    \label{s:eq:ward}
\end{equation}
\noindent where \\
\noindent $\Delta(C_i, C_j)$: Distance between two clusters $C_i$ and $C_j$ in $Cl$

The clustering proceeds by merging the clusters that are closest according to the linkage \citep{nielsen2016}. 

\subsubsection{Density-Based Spatial Clustering of Application with Noise (DBSCAN)}
Density-Based Spatial Clustering of Applications with Noise (DBSCAN, \cite{ester1996}, Python library scikit-learn, version 1.5) groups data according to density: First, neighbours for each data point are detected, i.e., the points that fall into a $d$-dimensional sphere of radius (\textit{epsilon}) drawn around the initial data point. If this set consists of less than a minimum number of neighbours (\textit{min\_samples}), the central data point is labelled as noise - a label that DBSCAN introduces for data points that it cannot assign. Otherwise, the central point is considered a core point and a new cluster is created. This cluster is expanded by all data points that are found by recursively drawing circles around the member points \citep{schubert2017}.

% \begin{figure}[ht]
%     \centering
%     \begin{subfigure}[b]{0.45\textwidth}
%         \centering
%         \includegraphics[width=\textwidth]{supplements/images_medium/labels_in_geospace_confetti.png}
%         \label{fig:confetti_geo}
%     \end{subfigure}
%     \raisebox{7\height}{ % This adjusts the vertical alignment of the middle figure (the arrow)
%         \begin{subfigure}[b]{0.05\textwidth} 
%             \centering
%             \includegraphics[width=\textwidth]{supplements/images_medium/dummy_arrow.png}
%         \end{subfigure}
%     }
%     \begin{subfigure}[b]{0.45\textwidth} 
%         \centering
%         \includegraphics[width=\textwidth]{supplements/images_medium/labels_in_umapspace_confetti.png}
%         \label{fig:confetti_umap}
%     \end{subfigure}
    
%     \caption{An example of a cluster set only consisting of \textit{confetti} in geographic (left) and in UMAP space (right). It was generated by applying DBSCAN to the embedded data with $\textit{epsilon} = 0.01$ and $\textit{min\_samples} = 2$.}
%     \label{fig:confetti}
% \end{figure}

\subsection{Internal Validation}
\label{s:internal_valdidation}
\begin{figure}[h]
    \centering
    \begin{subfigure}[b]{0.45\textwidth}
        \centering
        \includegraphics[width=0.4\textwidth]{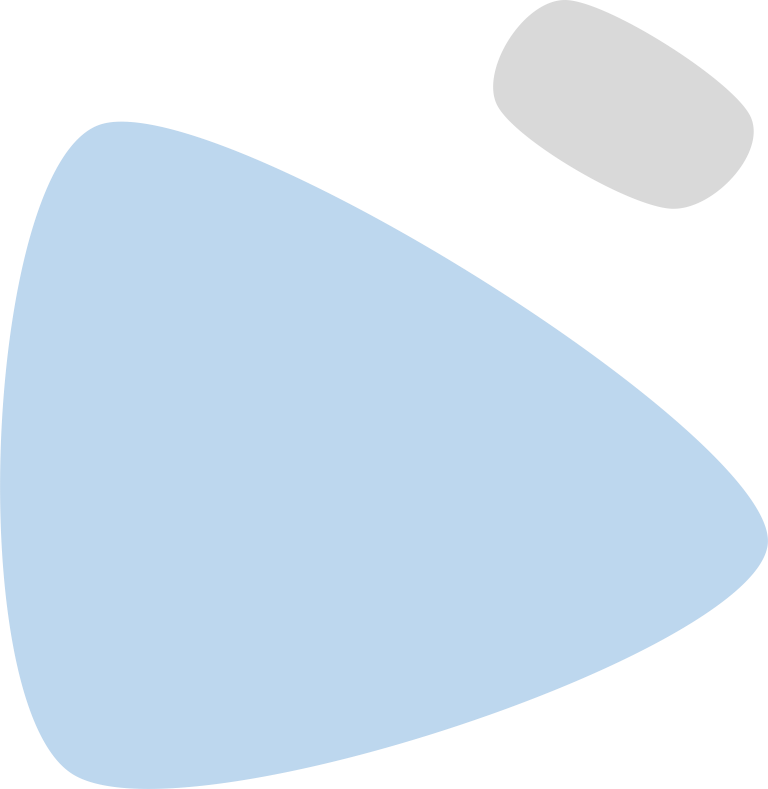}
    \end{subfigure}
    \hspace{0.01\textwidth}
    \begin{subfigure}[b]{0.45\textwidth} 
        \centering
        \includegraphics[width=0.5\textwidth]{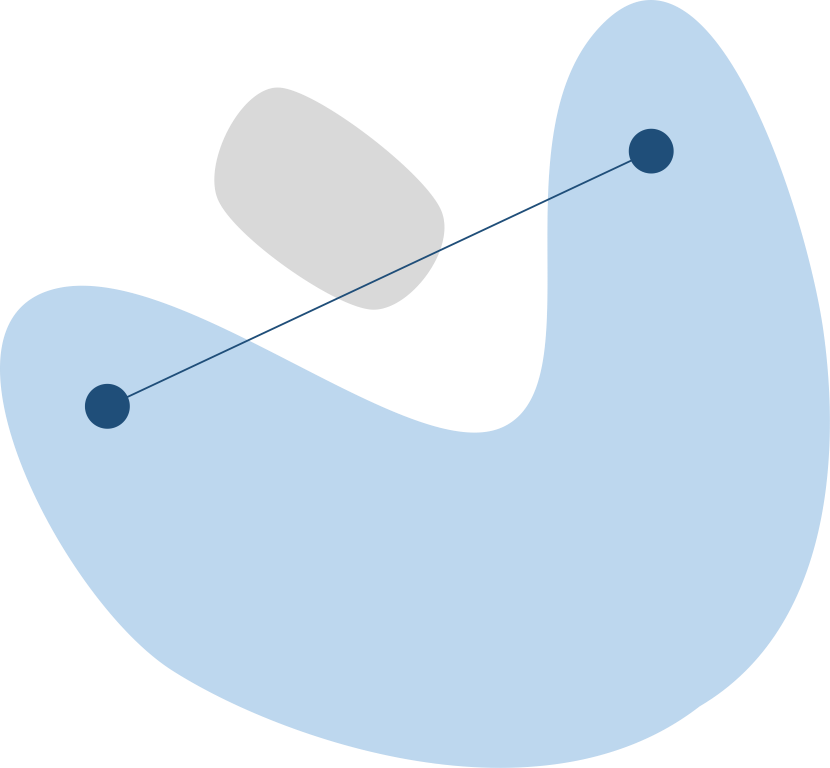}
    \end{subfigure}
    \caption{Convex (left, blue) and concave cluster (right, blue). In a concave cluster it is possible to find two points whose connecting line leaves the cluster. Three of the applied scores (Calinski-Harabasz Score (CH), Davies-Bouldin Score (DB),Silhouette Score (SH)) favour convex clusters, meaning that they would give the left two clusters (blue and grey) a better score than the right two clusters even though the separation is clear in both cases.}
    \label{s:fig:convex_concave}
\end{figure}

The scores used in this study are based on different mathematical assumptions and could hence lead to different conclusions. Moreover, Calinski-Harabasz, Davies-Bouldin and Silhouette Scores favour convex clusters. Favouring convexity means that, for example, a clustering with two perfect spheres would result in better scores than a banana-shaped cluster with a cluster in its cavity (Fig.~\ref{s:fig:convex_concave}).

\subsubsection{Calinski-Harabasz Score}
The Calinski-Harabasz Score (CH, \cite{calinski1974}) or Variance Ratio Criterion is a measure for within-cluster cohesion and between-cluster disparities. It computes distances to all cluster centres and to the overall data centre:
\begin{equation}
    \text{CH} = \frac{tr(B)}{tr(W)} \times \frac{n_E - k}{k - 1}
    \label{s:eq:ch}
\end{equation}
\noindent where $tr(W)$ is the trace of the within-cluster dispersion matrix $W$, which is defined as the sum of within-cluster distances (Eq.~\ref{s:eq:w})
\begin{equation}
    W = \sum_{q=1}^k \sum_{x \in C_q}(x - c_q)(x - c_q)^T
    \label{s:eq:w}
\end{equation}

\noindent
and $tr(B)$ is the trace of the between-cluster dispersion matrix $B$ (Eq.~\ref{s:eq:b}) that represents the weighted sum of distances between cluster centres and the overall data centre
\begin{equation}
    B = \sum_{q=1}^k n_q(c_q - c_E)(c_q - c_E)^T
    \label{s:eq:b}
\end{equation}
\noindent where \\
\noindent $k$: Number of clusters \\
\noindent $n_E$: Number of samples in data set $E$ \\
\noindent $n_q$: Number of points in cluster $q$ \\
\noindent $c_E$: Centre of data set $E$ \\
\noindent $C_q$: Set of samples in cluster $q$ \\
\noindent $c_q$: Centre of cluster $q$ \\

\cite{calinski1974} recommend selecting an absolute or local maximum in the CH curve or if not present, a comparatively steep increase to decide on a number of clusters. A decreasing coefficient is associated with a possible hierarchical structure of the points. An increasing score indicates clusters that are not well separated. A global/local maximum in the CH curve represents a good balance between within- and between-cluster dispersion, whereas a monotonically increasing CH curve suggests that no partition was better than the one into individual points \citep{calinski1974}.

\subsubsection{Davies-Bouldin Score}
Similar to the CH, the Davies-Bouldin Score (DB, \cite{davies1979}) considers both intra- and inter-cluster Euclidean distances. DB differs from CH by averaging over nearest-cluster similarities instead of taking all cluster centroids into account (Eq.~\ref{s:eq:db}):
\begin{equation}
    \text{DB} = \frac{1}{k}\sum_{i=1}^k \max_{i \neq j} \frac{s_i + s_j}{d_{ij}}
    \label{s:eq:db}
\end{equation}
\noindent where \\
\noindent $k$: Number of clusters \\
\noindent $s_i$: Mean distance between each point of cluster i and the cluster centroid \\
\noindent $d_{ij}$: Distance between the centroids of cluster i and its most similar cluster j \\

\noindent
Small DB indicates better clustering with the minimum value being 0.

\subsubsection{Silhouette Score}
The Silhouette Score (SH, \cite{rousseeuw1987}) computes the mean intra-cluster distance $a$ and the mean nearest-cluster distance $b$. Contrary to CH and DB, the SH uses pairwise distances, not centroid distances. The SH for one sample is computed as
\begin{equation}
    s = \frac{b-a}{max(a, b)}
    \label{s:eq:s}
\end{equation}

The SH for the complete clustering averages the SH from each sample. The metric to measure the distance between two data points is Euclidean by default. Besides being higher for convex clusters, this coefficient favours clusters that are dense and clearly separated. SH close to 1 is the best achievable result. Scores around 0 suggest overlapping clusters. Negative values down to \SI{-1}{} signify that the data point better fits to another cluster.

\subsubsection{Contiguous Density Region Index}
The Contiguous Density Region Index (CDR, \cite{rojasthomas2021}) bases on local density differences. Local density of a data point $x_i$ is defined as the Euclidean distance to its nearest neighbour in the same cluster. Local density variation of a cluster $C_q$, or uniformity, is then computed as

\begin{equation}
    \text{unif}(C_q) = 
        \begin{cases}
        \frac{\sum_{x_i \in C_q} \left| \text{local\_dens}(x_i) - \text{avg\_dens}(C_q)  \right|}{\text{avg\_dens}(C_q)}, & x_i \in C_q, n_q > 1 \\
            0, & n_q = 1
        \end{cases}
\end{equation}

\noindent where \\
\noindent $n_q$: Number of data points in cluster $C_q$ \\
\noindent avg\_dens$(C_q)$: Average local density of cluster $C_q$ \\

The CDR is the average cluster uniformity, normalised by the number of data points in the cluster 

\begin{equation}
    \text{CDR} = \frac{\sum_{q=1}^{l} n_q \cdot unif(C_q)}{n}
\end{equation}
\noindent where \\
\noindent $n$: Number of data points \\
\noindent $l$: Number of clusters in Cl \\

A small CDR, with 0 being the minimum, indicates a better clustering, i.e. a clustering with few density variations within each cluster.

\subsubsection{Clustering Validation Index based on Nearest Neighbours Halkidi} 
The Clustering Validation Index based on Nearest Neighbours by \cite{halkidi2015} (CVNNH) is an adaptation of CVNN \citep{liu2013} which combines measures of cluster separation and compactness

\begin{equation}
    \text{CVNN}(K) = \text{sep}(K) + \text{comp}
\end{equation}

For a cluster \(C_q\), separation is the average proportion of data point in the K-nearest neighbourhood of its member points that belong to other clusters. The overall separation of a cluster set Cl is the maximum separation across all clusters

\begin{equation}
    \text{sep}(K) = \max_{C_q \in \text{Cl}} \left(\frac{1}{n_q} \sum_{x_i \in  C_q} \frac{n_K(x_i)}{K}\right)
\end{equation}

\noindent where \\
\noindent $n_K(x_i)$: Number of data points in the K-nearest neighbourhood of data point $x_i$ that belong to a different cluster than $x_i$\\

Compactness after \cite{halkidi2015} assesses intra-cluster distances by measuring the average pairwise distances between all distinct point pairs within the same cluster

\begin{equation}
    \text{comp} = \frac{\sum_{C_q \in \text{Cl}} \sum_{\substack{x_i, x_j \in C_q \\ x_i \neq x_j}} d(x_i, x_j)}{\sum_{C_q \in \text{Cl}} n_q(n_q - 1)}
\end{equation}

\noindent
where \\
$d(x_i, x_j)$: Distance between points $x_i$ and $x_j$ \\

CVNNH was calculated using the ASCVI library \citep{schlake2024} which applies Euclidean distances for $d$. A low CVNNH indicates good clustering, with the optimal value being 0, corresponding to perfectly compact and well-separated clusters.

\subsubsection{K-Density-Based Cluster Validation Index} 
The k-Density-Based Cluster Validation Index (k-DBCV, \cite{hammer2024}, in review) is an efficient implementation of DBCV \citep{moulavi2014} which evaluates cluster quality using density separation between clusters and intra-cluster density sparseness

\begin{equation}
    \text{k-DBCV} = \frac{1}{n} \sum_{i=1}^{k} n_i \frac{\min(sep(C_i, C_{j \neq i})) - spars(C_i)}{\max(sep(C_i, C_{j \neq i}), spars(C_i))}
\end{equation}

\noindent where \\
\noindent $n$: Number of data points \\
\noindent $n_i$: Number of data points in cluster $C_i$ \\
\noindent $k$: Number of clusters \\

Both, separation and sparseness, are derived from mutual reachability distances (MRDs). The MRD between two distinct points $x_i$ and $x_j$ is

\begin{equation}
    \text{MRD}(x_i, x_j) = \max(\text{coredist}(x_i), \text{coredist}(x_j), d(x_i, x_j))
\end{equation}

where $d(x_i, x_j)$ is the pairwise-distance between points and coredist is the all-points-core-distance describing inverse density in a local neighbourhood of $x_i$

\begin{equation}
    \text{coredist}(x_i) = \left( \frac{\sum_{j=2}^{n_q} ( \frac{1}{\text{KNN}(x_i, j)} )^f }{n_q - 1} \right)^{-\frac{1}{f}}
\end{equation}

\noindent where \\
\noindent $f$:  Number of features \\
\noindent KNN$(x_i, j)$: Distance between data point $x_i$ and its $j^{th}$ nearest neighbour \\

An MRD-weighted minimum spanning tree (MST) is constructed per cluster. Sparseness is defined as the maximum edge weight of the MST of the respective cluster.
Separation between two clusters $C_i$ and $C_j$ is the minimum MRD between any point in $C_i$ and any point in $C_j$.

The score ranges from –1 to 1, where values close to 1 indicate well-connected, well-separated clusters, values near 0 suggest weak or ambiguous structure, and negative values reflect poorly separated clusters or noise.

\subsection{Elbow Method}
The elbow or knee method is a well-established subjective heuristic method to balance cost and benefit/performance when altering a system parameter \citep{satopaa2011}. 
For example, a Cluster Validity Index (CVI) can be plotted over the number of clusters detected by KMeans, resulting in an L-shaped curve. The elbow of the L-shape indicates a tradeoff between complexity and performance. 
% An automatic algorithm for the elbow detection \citep{satopaa2011} was applied to tune hyperparameters (Python, kneed, version 0.8.5).
The elbow concept is transferable to 2d for methods that require two hyperparameters, here for DBSCAN by replacing the elbow curve by a heatmap, in which the elbow was detected as the steepest gradient in the heatmap by computing differences between neighbouring cells.

\section{Figures for Results and Discussion Sections}
\begin{figure}[h]
    \centering
    \includegraphics[width=0.7\textwidth]{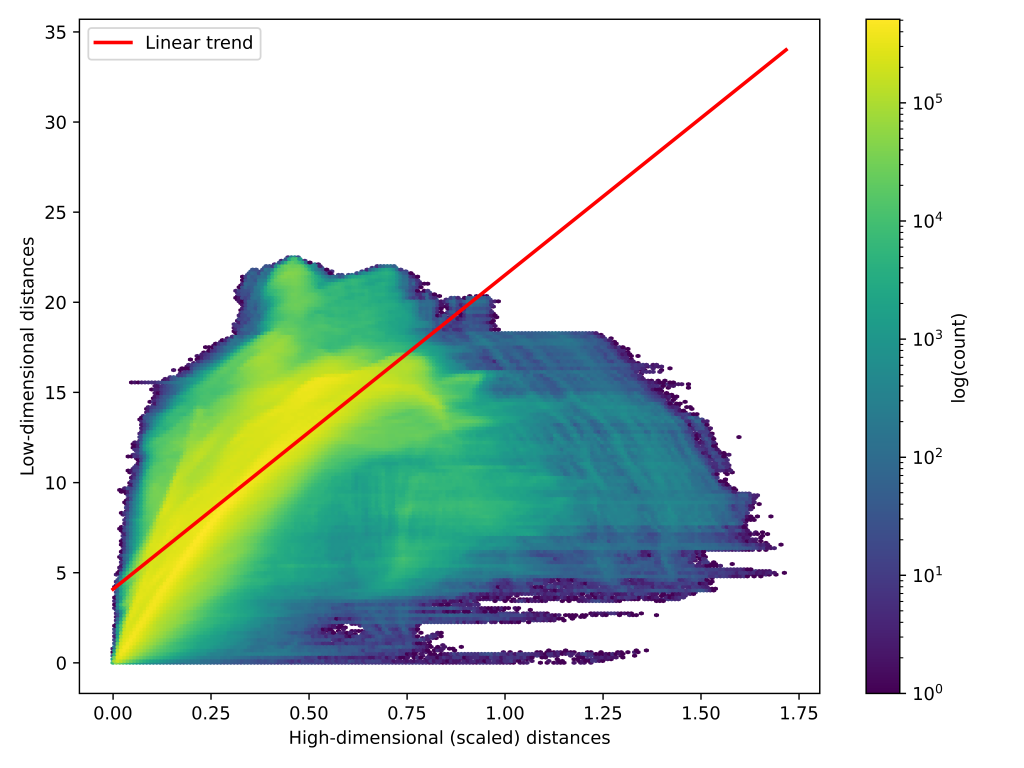}
    \caption{Shepard plot of a UMAP embedding (\textit{n\_neighbors}=20, \textit{min\_dist}=0, \textit{metric}='euclidean', \textit{n\_components}=3) computed on the min-max scaled data. While local distances are well preserved, global structures are more distorted.}
    \label{s:fig:shepard}
\end{figure}

\begin{figure}[h]
    \centering
    \includegraphics[width=\textwidth]{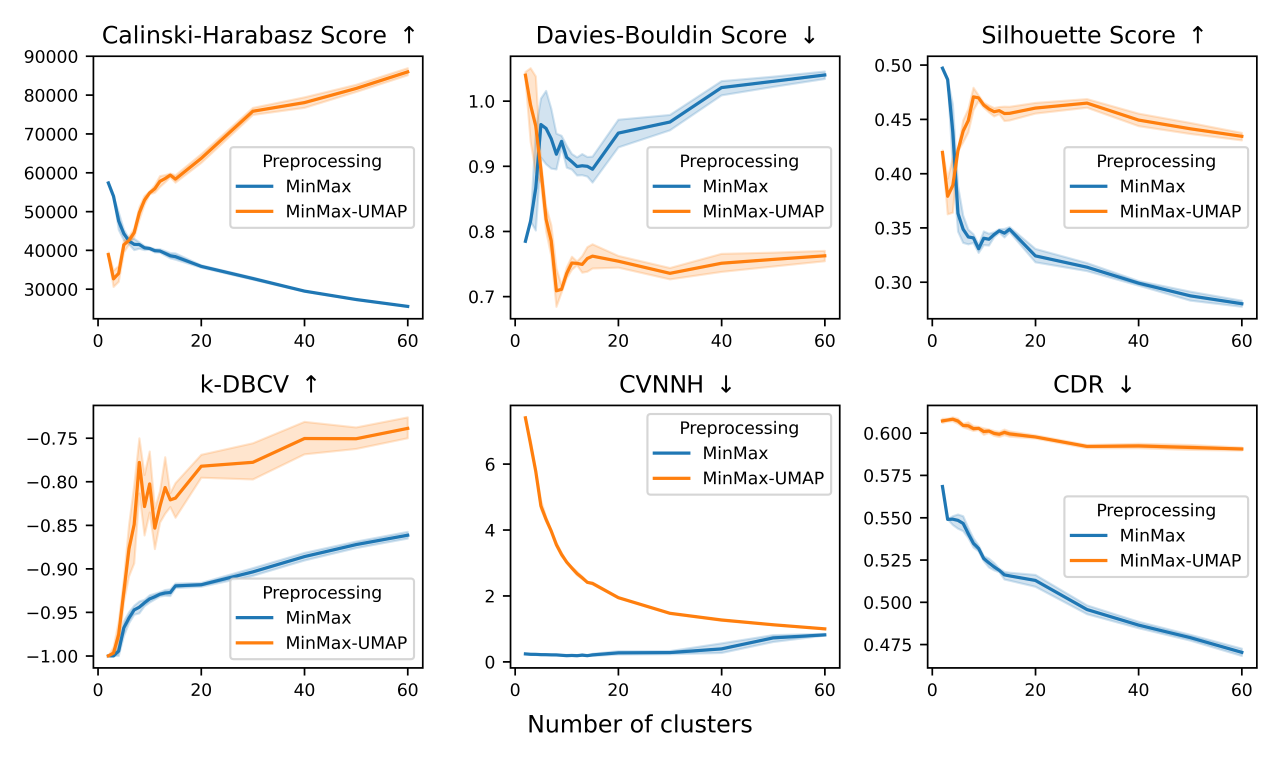}
    \caption{Internal validation of KMeans. For clustering the original data (blue), Calinski-Harabasz, Davies-Bouldin and Silhouette scores suggest two clusters as the optimal number of clusters, while k-Density Based Clustering Validation (k-DBCV) and Contiguous Density Region (CDR) prefer the maximum tested value of 60 clusters. k-DBCV remains negative. Clustering Validation Index Based on Nearest Neighbours Halkidi (CVNNH) reaches its optimum for 12 clusters. For clustering the embedded data (orange), four out of six scores improved. SH and DB favour eight clusters, the other scores prefer 60 clusters.}
    \label{s:fig:kmeans_hyps}
\end{figure}

\begin{figure}[h]
    \centering
    \begin{subfigure}[b]{0.45\textwidth}
        \centering
        \includegraphics[width=\textwidth]{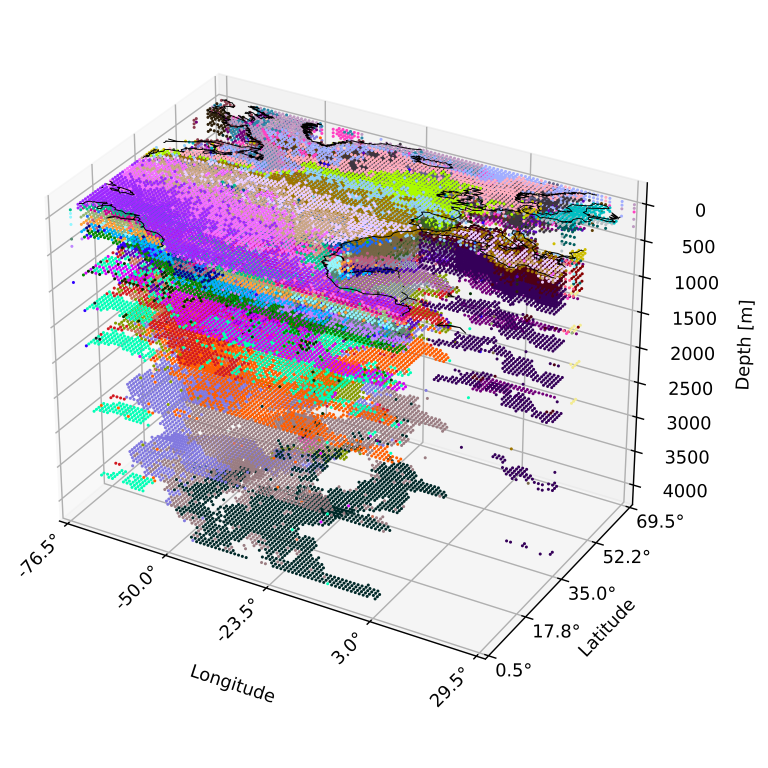}
    \end{subfigure}
    \raisebox{7\height}{ % This adjusts the vertical alignment of the middle figure (the arrow)
        \begin{subfigure}[b]{0.05\textwidth} 
            \centering
            \includegraphics[width=\textwidth]{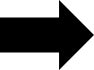}
        \end{subfigure}
    }
    \begin{subfigure}[b]{0.45\textwidth} 
        \centering
        \includegraphics[width=\textwidth]{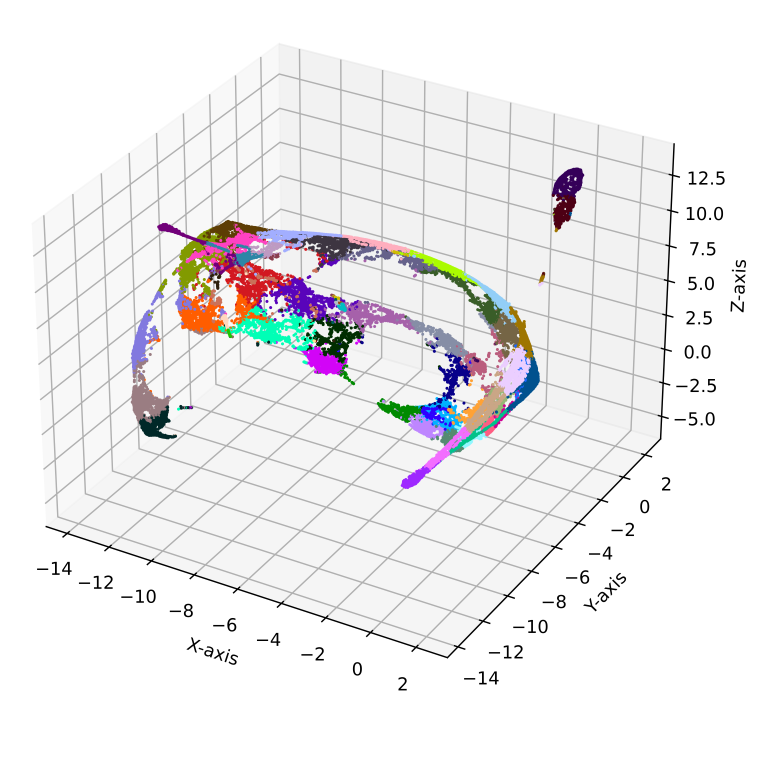}
    \end{subfigure}
    \caption{KMeans on original data with the number of clusters set to 60 (left) and its projection into the embedded space (right). A tendency to form convex clusters in UMAP space is visible, clearly not suiting the given data structure.}
    \label{s:fig:kmeans_original_60}
\end{figure}

\begin{figure}[h]
    \centering
    \begin{subfigure}[b]{0.45\textwidth}
        \centering
        \includegraphics[width=\textwidth]{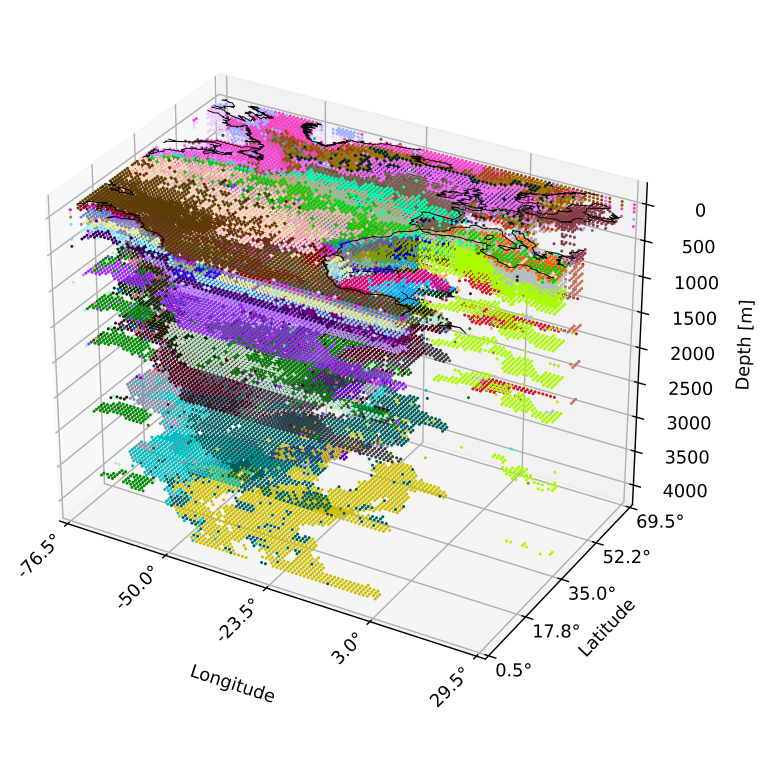}
    \end{subfigure}
    \raisebox{7\height}{ % This adjusts the vertical alignment of the middle figure (the arrow)
        \begin{subfigure}[b]{0.05\textwidth} 
            \centering
            \includegraphics[width=\textwidth]{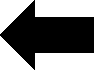}
        \end{subfigure}
    }
    \begin{subfigure}[b]{0.45\textwidth} 
        \centering
        \includegraphics[width=\textwidth]{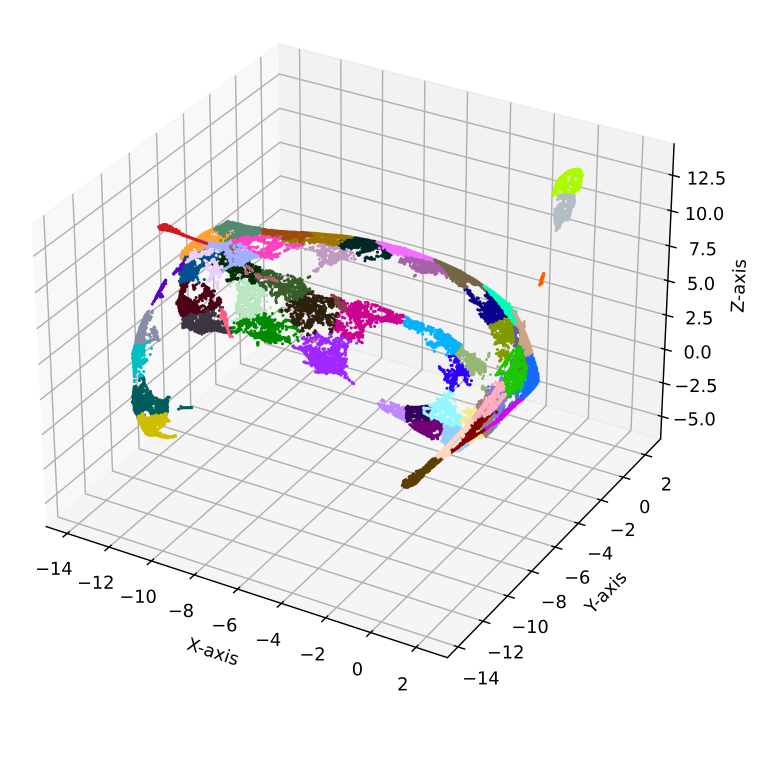}
    \end{subfigure}
    \caption{KMeans on embedded data with number of clusters set to 60 (right) and its projection into geographic space (left). The tendency to form convex clusters in UMAP space is even more clear than for KMeans on original data, clearly not suiting the given data structure.}
    \label{s:fig:kmeans_embedding_60}
\end{figure}

\begin{figure}[h]
    \centering
    \includegraphics[width=\textwidth]{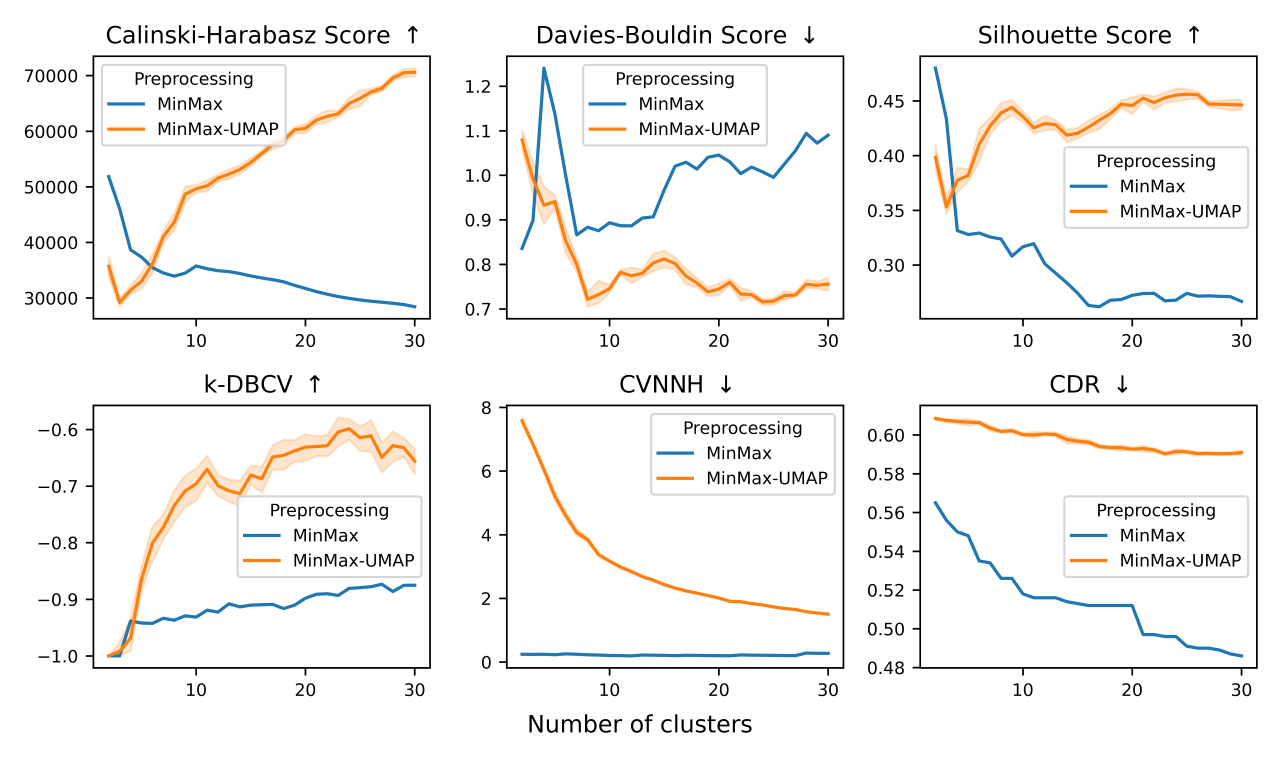}
    \caption{Internal validation of agglomerative Ward. For clustering the original data (blue), all classical scores (Calinski-Harabasz, Davies-Bouldin and Silhouette) suggested two clusters as optimal. Structural scores (k-Density Based Clustering Validation (k-DBCV), Clustering Validation Index Based on Nearest Neighbours Halkidi (CVNNH) and Contiguous Density Region (CDR)) instead favoured more fine-grained solutions with 27, 12, and 30 clusters respectively. For clustering the embedded data (orange), the scores gave a more clear answer but did not agree on an optimal cluster number. For clustering the embedded data (orange), the scores showed clearer trends but did not agree on a single optimal cluster number.}
    \label{s:fig:ward_hyps}
\end{figure}

\begin{figure}[htb]
    \centering
    \includegraphics[width=\textwidth]{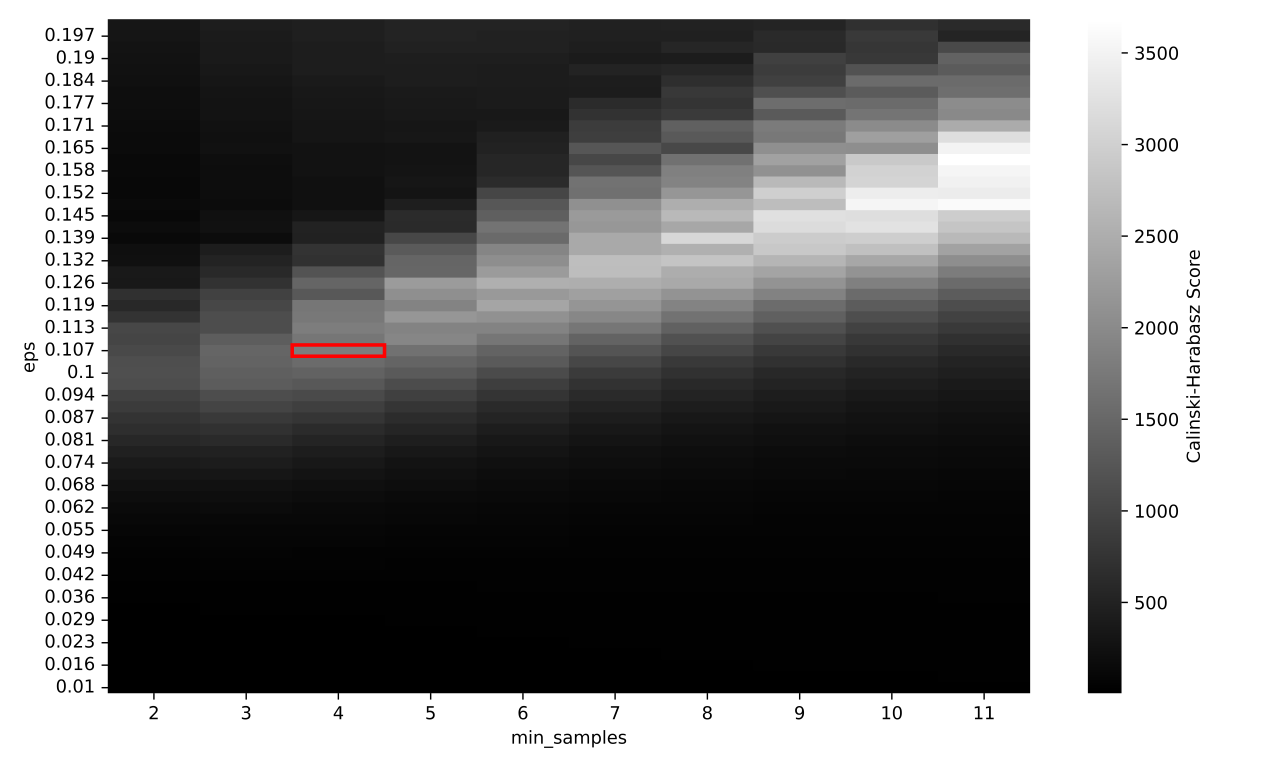}
    \caption{Heatmap showing the Calinski-Harabasz Score (CH) for DBSCAN clustering applied to the embedded data space. Each grid cell represents one cluster experiment carried out with a given combination of the hyperparameters values \textit{epsilon} and \textit{min\_samples}. The final hyperparameter selection is highlighted by a red square.}
    \label{s:fig:dbscan_ch}
\end{figure}

\clearpage
\begin{sidewaystable}
\centering
\caption{Suggested hyperparameter combinations for clustering with DBSCAN according to Calinski-Harabasz Score (CH), Davies-Bouldin Score (DB), Silhouette Score (SH), k-Density Based Clustering Validation (k-DBCV), Clustering Validation Index Based on Nearest Neighbours Halkidi (CVNNH) and Contiguous Density Region (CDR). High CH, SH and k-DBCV as well as low DB, CVNNH and CDR indicate a well separated clustering.}
\label{s:tab:dbscan_hyps}
\begin{tabularx}{\textwidth}{X X X X X X X X X X X X}
%\begin{tabular*}{\textwidth}{@{\extracolsep{\fill}}p{1.6cm}p{1.8cm}p{1.6cm}p{1.2cm}p{0.9cm}p{0.9cm}p{0.9cm}p{1.6cm}p{1.6cm}@{}}
\toprule
\textbf{Da\-ta} & 
\textbf{Se\-lec\-tion cri\-te\-ri\-on} & 
\textbf{$epsilon$} & 
\textbf{$min\_ \allowbreak samples$} & 
\textbf{CH $\uparrow$} & 
\textbf{DB $\downarrow$} & 
\textbf{SH $\uparrow$} & 
\textbf{k-DBCV $\uparrow$} & 
\textbf{CVNNH $\downarrow$} & 
\textbf{CDR $\downarrow$} & 
\textbf{Num\-ber of clus\-ters} & 
\textbf{Pro\-por\-tion noise grid cells [\%]} \\
\midrule
Original & CH & 0.11949153 & 11 & 800 & 1.44 & 0.55 & -1 & 0.98 & 0.54 & 4 & 0.3 \\
Original & DB & 0.10983051 & 3 & 120 & 1.08 & 0.38 & -1 & 1.31 & 0.55 & 11 & 0.1 \\
Original & SH & 0.2 & 9 & 337 & 1.34 & 0.58 & -1 & 0.96 & 0.56 & 3 & 0.1 \\
Original & k-DBCV & 0.01 & 10 & 375 & 1.98 & -0.62 & -0.18 & 1.91 & 0.79 & 71 & 58.5 \\
Original & CVNNH & 0.17423729 & 3 & 334 & 1.35 & 0.57 & -1 & 0.95 & 0.57 & 3 & 0.1 \\
Original & CDR & 0.01322034 & 2 & 8 & 1.45 & -0.75 & -1 & 1.53 & 0.43 & 1455 & 21 \\
Em\-bed\-ded & CH & 0.15491525 & 11 & 3747 & 1.45 & -0.2 & -0.35 & 3.58 & 0.59 & 128 & 4.2 \\
Em\-bed\-ded & DB & 0.02610169 & 10 & 25 & 0.95 & -0.77 & 0.05 & 7.6 & 7.77 &  262 & 93.3 \\
Em\-bed\-ded & SH & 0.04542373 & 2 & 45 & 1.11 & 0.03 & 0.17 & 1.47 & 0.45 & 4969 & 18.9 \\
Em\-bed\-ded & k-DBCV & 0.02932203 & 3 & 16 & 1.06 & -0.25 & 0.23 & 1.86 & 0.73 & 3281 & 51 \\
Em\-bed\-ded & CVNNH & 0.03576271 & 2 & 21 & 1.08 & 0.00 & 0.22 & 1.47 & 0.45 & 6058 & 28.5 \\
Em\-bed\-ded & CDR & 0.03898305 & 2 & 27 & 1.09 & 0.02 & 0.21 & 1.47 & 0.45 & 5734 & 24.8 \\
\bottomrule
%\end{tabular*}
\end{tabularx}
\end{sidewaystable}
\clearpage

\begin{figure}[h]
    \centering
    \includegraphics[width=0.32\textwidth]{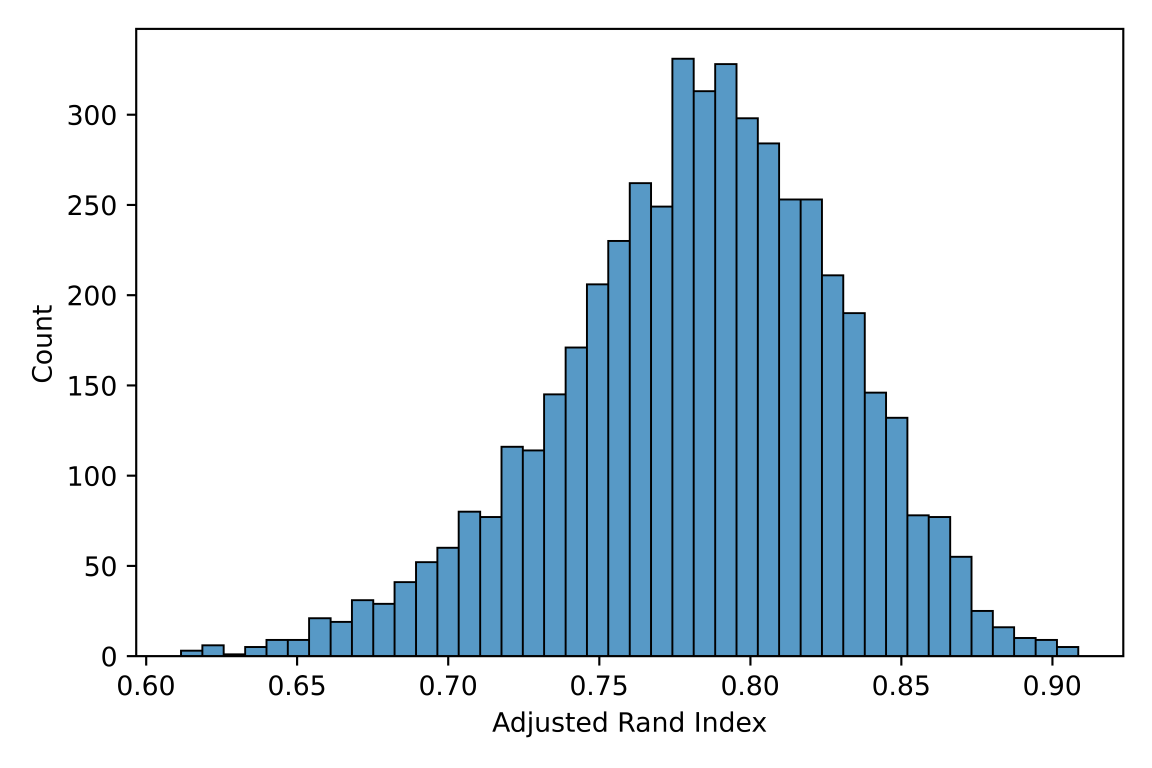}
    \includegraphics[width=0.32\textwidth]{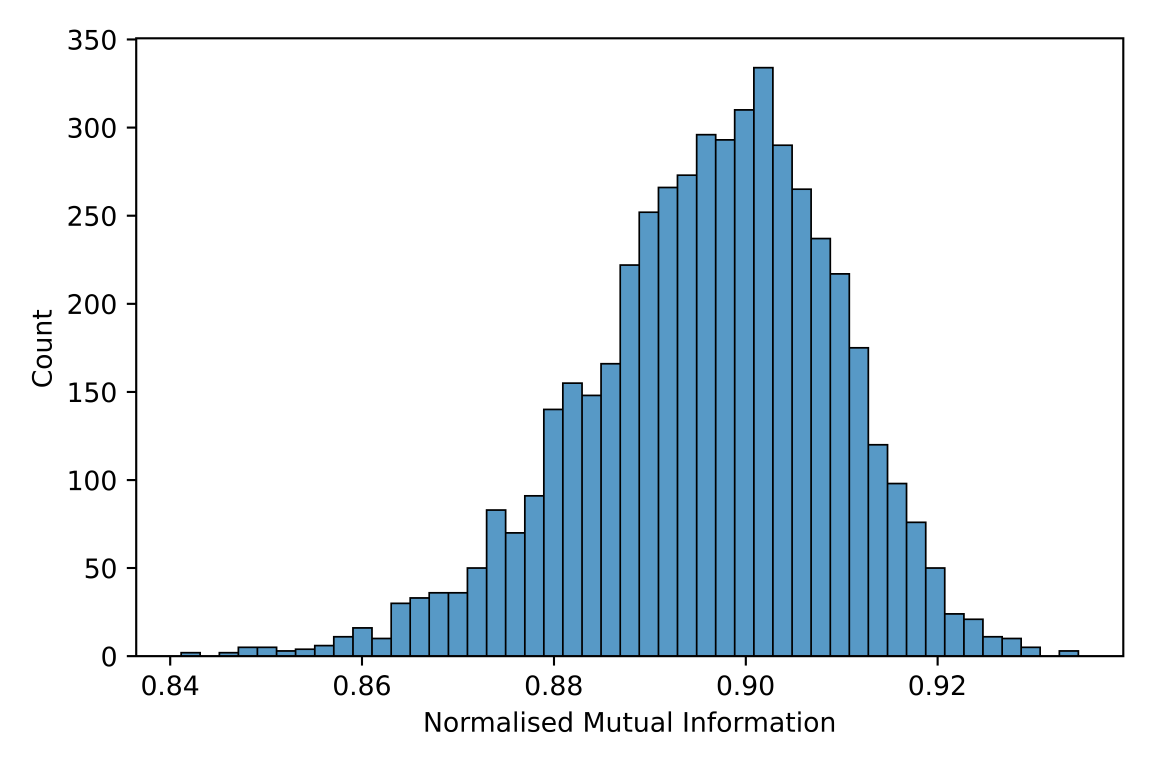}
    \includegraphics[width=0.32\textwidth]{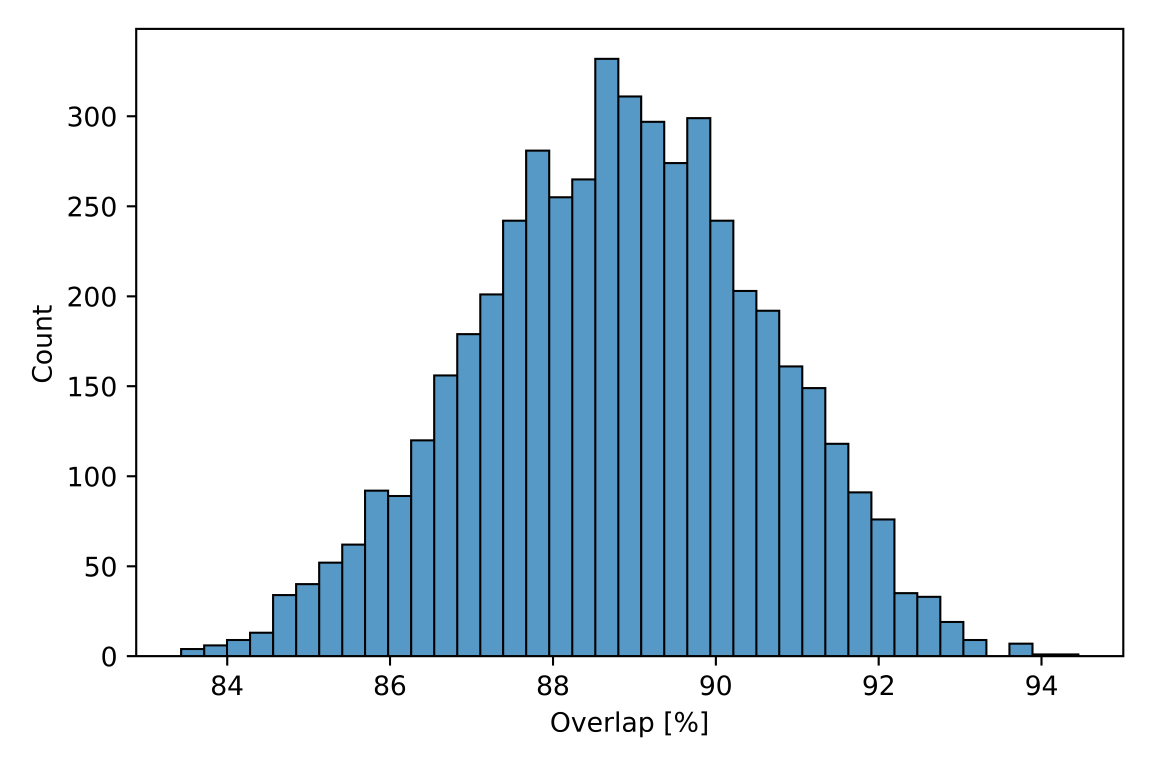}
    \caption{Similarity metrics computed for every pair of cluster sets resulting from re-running UMAP-DBSCAN 100 times. On average, the Adjusted Rand Index was $0.78\pm0.05$, Normalised Mutual Information was $0.91\pm0.01$ and overlap was $88.81\pm1.8\%$.}
    \label{s:fig:overlap}
\end{figure}

% Uncertainty plots
\begin{figure}[h]
    \centering
    \begin{subfigure}[c]{0.3\textwidth} % Increased width to allow for side-by-side placement
        \centering
        \includegraphics[width=\textwidth]{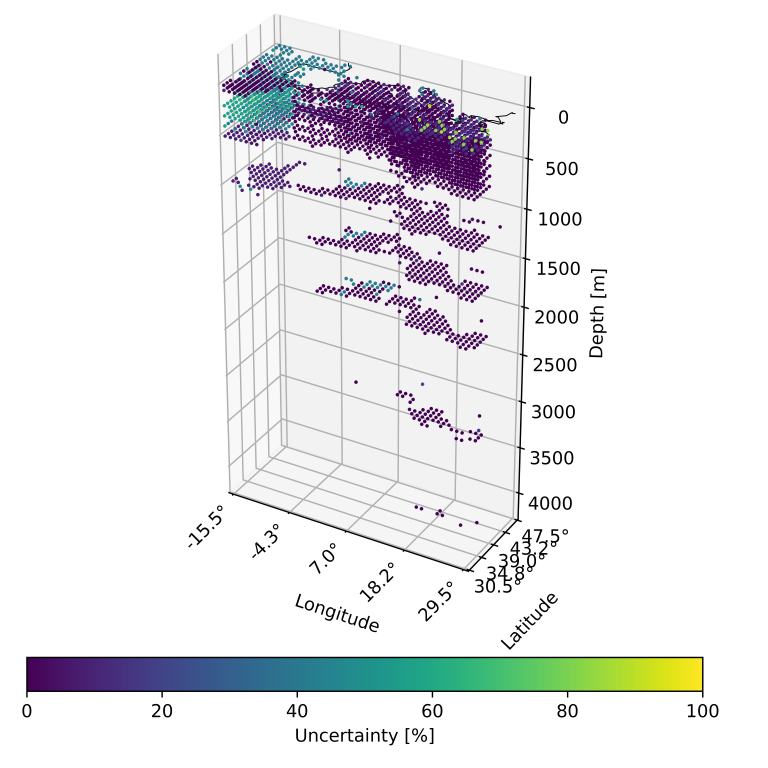}
    \end{subfigure}
    \hspace{0.05\textwidth} % Adds horizontal space between the subfigures
    \begin{subfigure}[c]{0.3\textwidth} % Increased width to allow for side-by-side placement
        \centering
        \includegraphics[width=\textwidth]{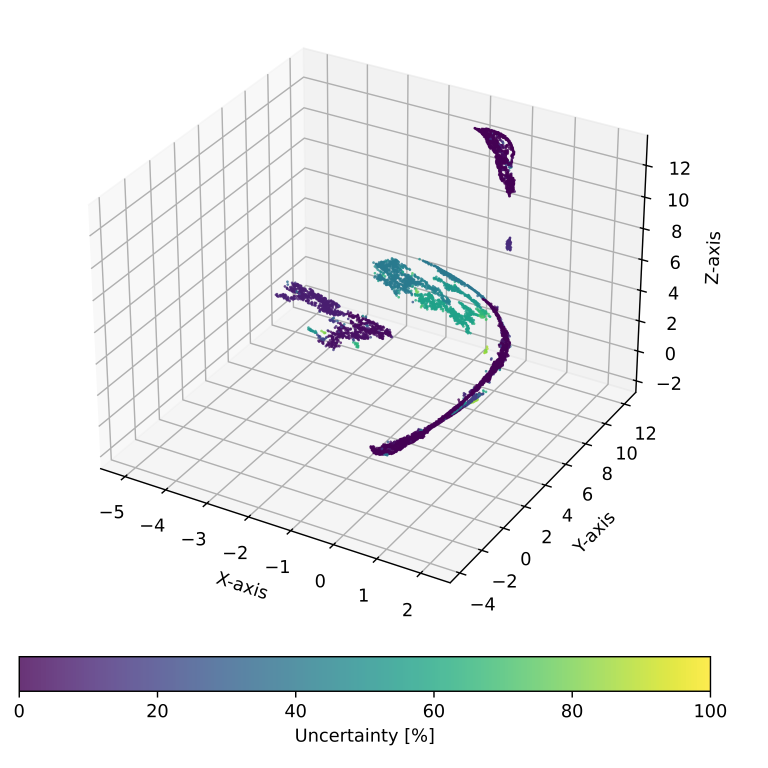}
    \end{subfigure}
    \begin{subfigure}[c]{0.3\textwidth} 
        \centering
        \includegraphics[width=\textwidth]{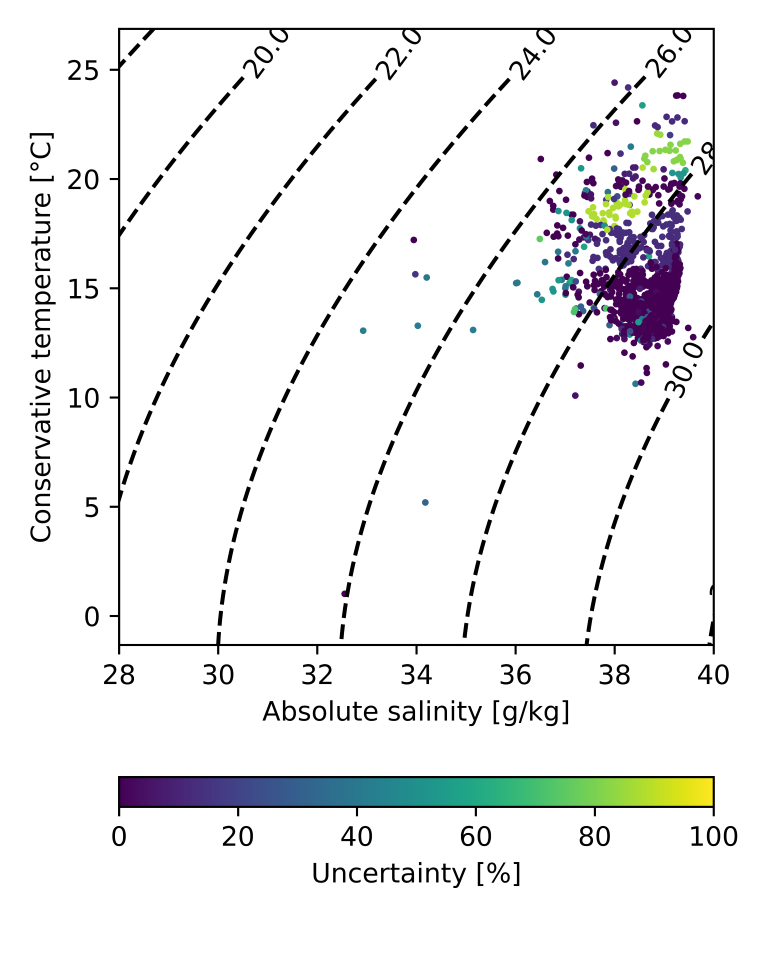}
    \end{subfigure}
    
    \caption{NEMI uncertainty, generated from the variability of the combination of 100 UMAP-DBSCAN runs, of selected  clusters in the Mediterranean Sea (labels 6, 32, 35, 138) and its related in-/outflow regions in the North Atlantic (labels 0, 10, 19, 28) in geographic (left, zoomed into the geographic area of the Mediterranean Sea), embedded (middle) and temperature-salinity (TS) space (right). The average uncertainty was $13.08\pm20.7\%$.}
    \label{s:fig:medi_uncertainty}
\end{figure}

\begin{figure}[h]
    \centering
    \begin{subfigure}[c]{0.3\textwidth} % Increased width to allow for side-by-side placement
        \centering
        \includegraphics[width=\textwidth]{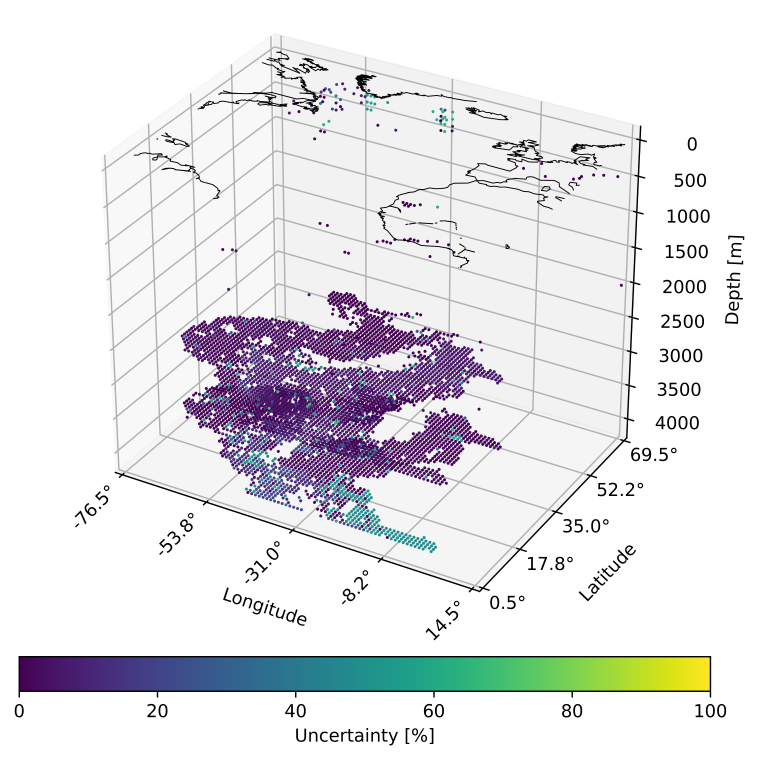}
    \end{subfigure}
    \hspace{0.05\textwidth} % Adds horizontal space between the subfigures
    \begin{subfigure}[c]{0.3\textwidth} % Increased width to allow for side-by-side placement
        \centering
        \includegraphics[width=\textwidth]{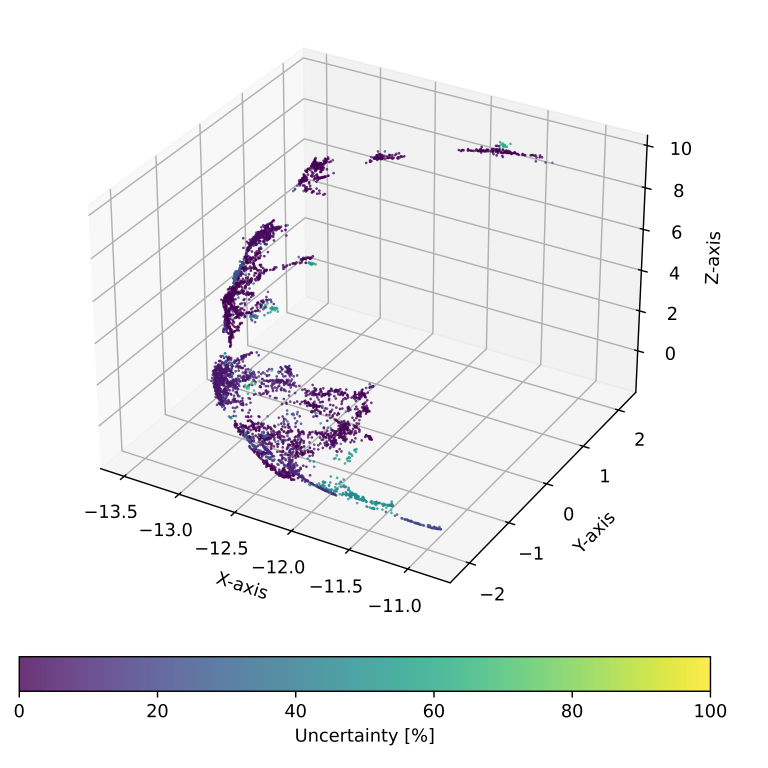}
    \end{subfigure}
    \begin{subfigure}[c]{0.3\textwidth} 
        \centering
        \includegraphics[width=\textwidth]{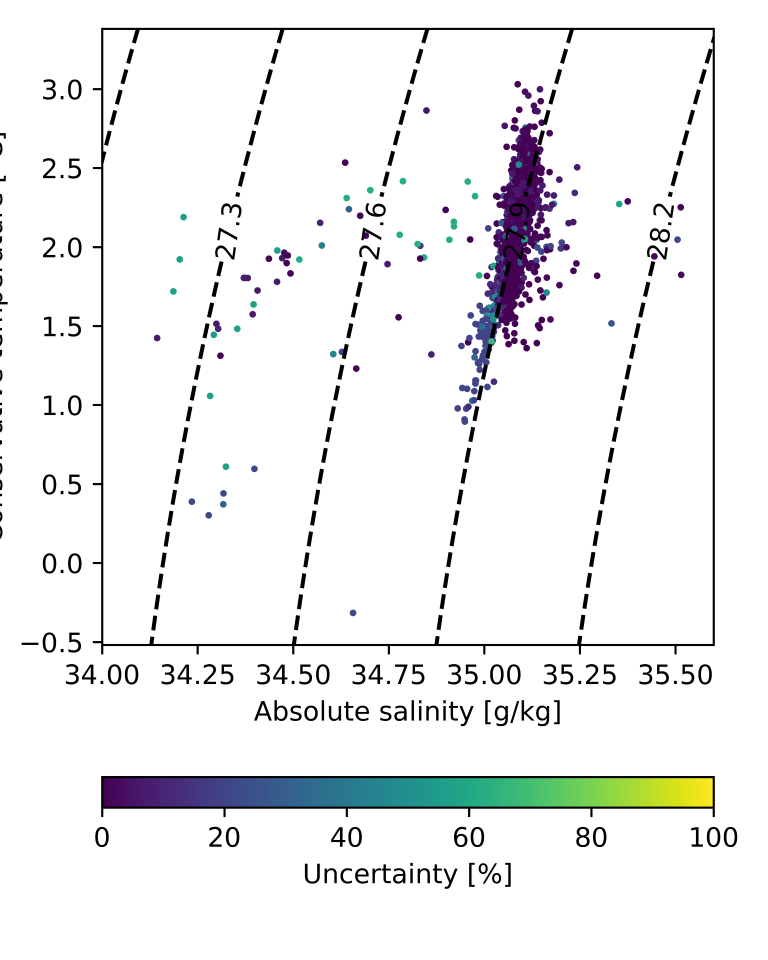}
    \end{subfigure}
    
    \caption{NEMI uncertainty, generated from the variability of the combination of 100 UMAP-DBSCAN runs, of selected clusters in the deep North Atlantic (labels 2, 11, 31, 42, 51, 52) in geographic (left), embedded (middle) and temperature-salinity (TS) space (right). The average uncertainty was $7.74\pm13.4\%$.}
    \label{s:fig:dna_uncertainty}
\end{figure}

\begin{figure}[h]
    \centering
    \begin{subfigure}[c]{0.3\textwidth} % Increased width to allow for side-by-side placement
        \centering
        \includegraphics[width=\textwidth]{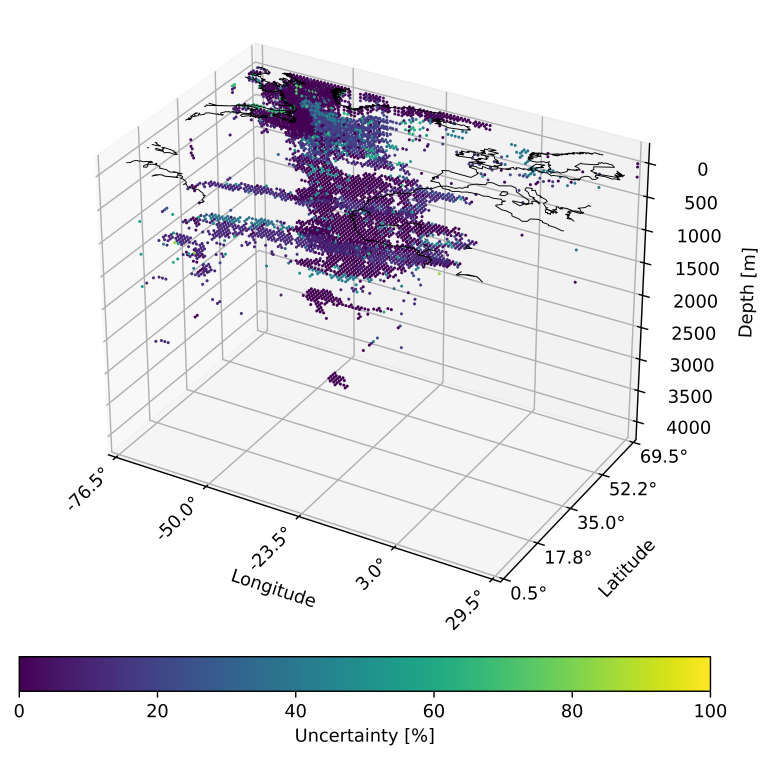}
    \end{subfigure}
    \hspace{0.05\textwidth} % Adds horizontal space between the subfigures
    \begin{subfigure}[c]{0.3\textwidth} % Increased width to allow for side-by-side placement
        \centering
        \includegraphics[width=\textwidth]{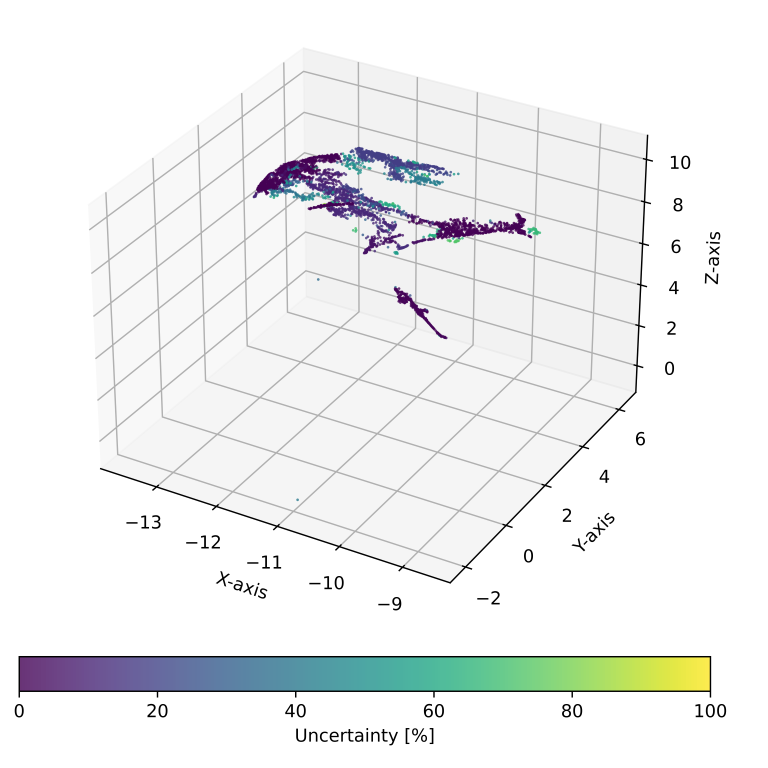}
    \end{subfigure}
    \begin{subfigure}[c]{0.3\textwidth} 
        \centering
        \includegraphics[width=\textwidth]{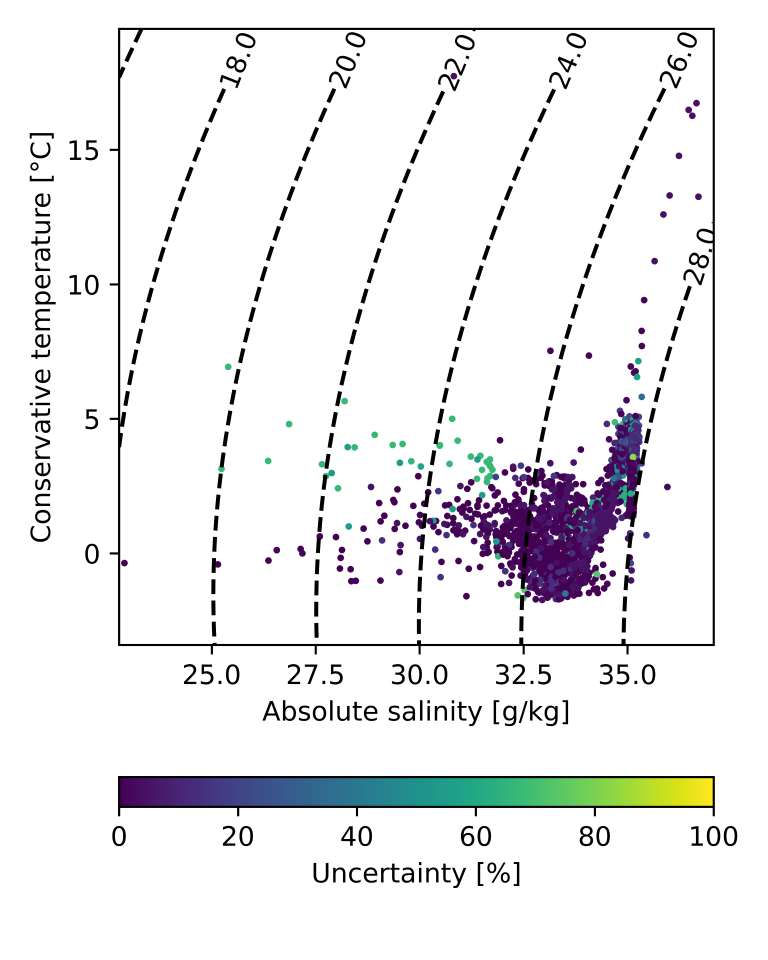}
    \end{subfigure}
    
    \caption{NEMI uncertainty, generated from the variability of the combination of 100 UMAP-DBSCAN runs, of selected clusters in the Labrador Sea and Davis Strait (labels 3, 17, 18, 24, 30, 42) in geographic (left), embedded (middle) and temperature-salinity (TS) space (right). The average uncertainty was $10.73\pm15.4\%$.}
    \label{s:fig:lab_uncertainty}
\end{figure}

\begin{figure}[h]
    \centering
    \includegraphics[width=0.7\textwidth]{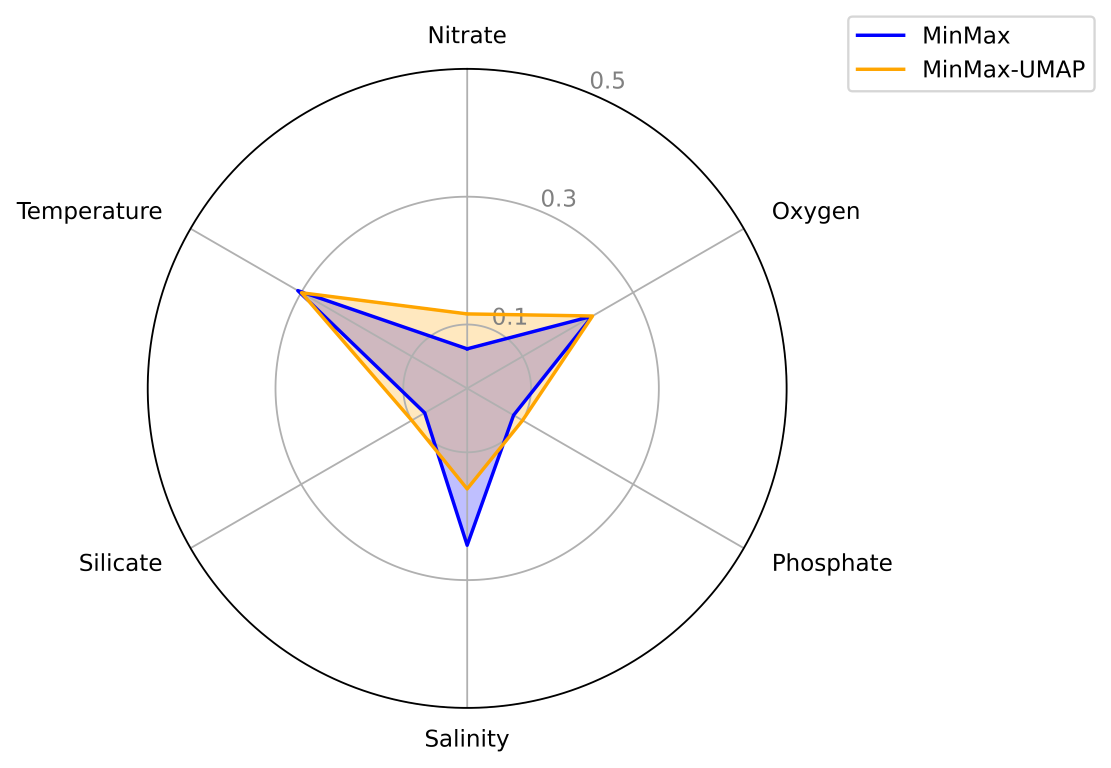}
    \caption{Feature importance of the six input variables averaged over all six clustering experiments revealed temperature, salinity and oxygen as strongest drivers. Clustering embedded data shows a more balanced parameter influence than clustering original data.}
    \label{s:fig:feature_importance}
\end{figure}

\begin{figure}[h]
    \centering
    \includegraphics[width=0.9\textwidth, trim=0 120 0 100, clip]{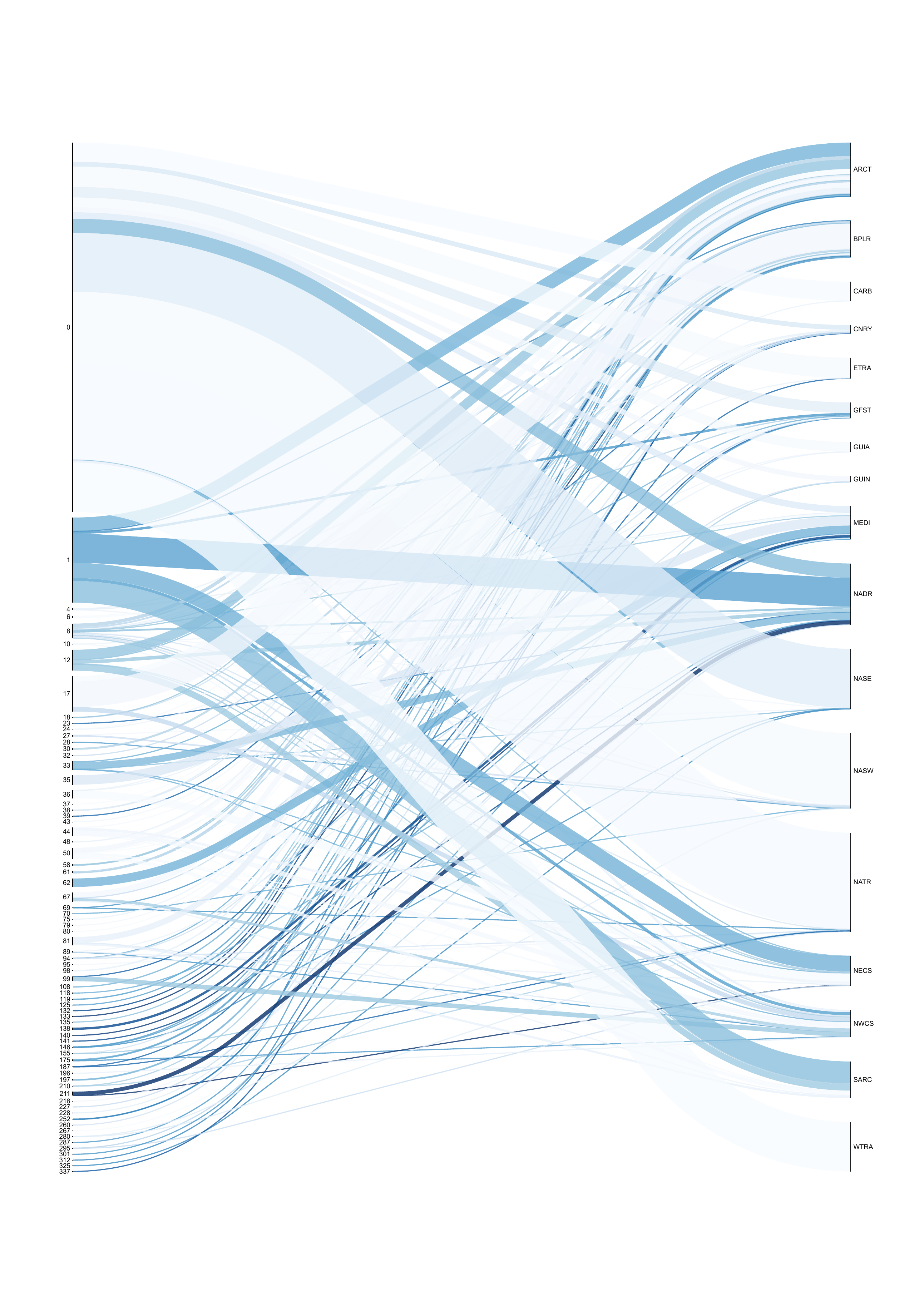}
    \caption{Sankey plot (Python library floweaver, version 2.0.0, \cite{lupton2017}) illustrating the distribution of geographic sampling points across the presented cluster set (left) and the Longhurst provinces (right, \cite{longhurst2007}). The width of each flow corresponds to the number of shared points between clusters and provinces. Flow colour indicate the Native Emergent Manifold Interrogation (NEMI) uncertainty of this study's clusters. The plot shows that most of our clusters provide a finer subdivision of geographic points compared to Longhurst provinces, except for clusters 0 and 1, which are further subdivided by the Longhurst scheme.}
    \label{s:fig:sankey_longhurst}
\end{figure}

\begin{figure}[h]
    \centering
    \includegraphics[width=0.9\textwidth, trim=0 120 0 100, clip]{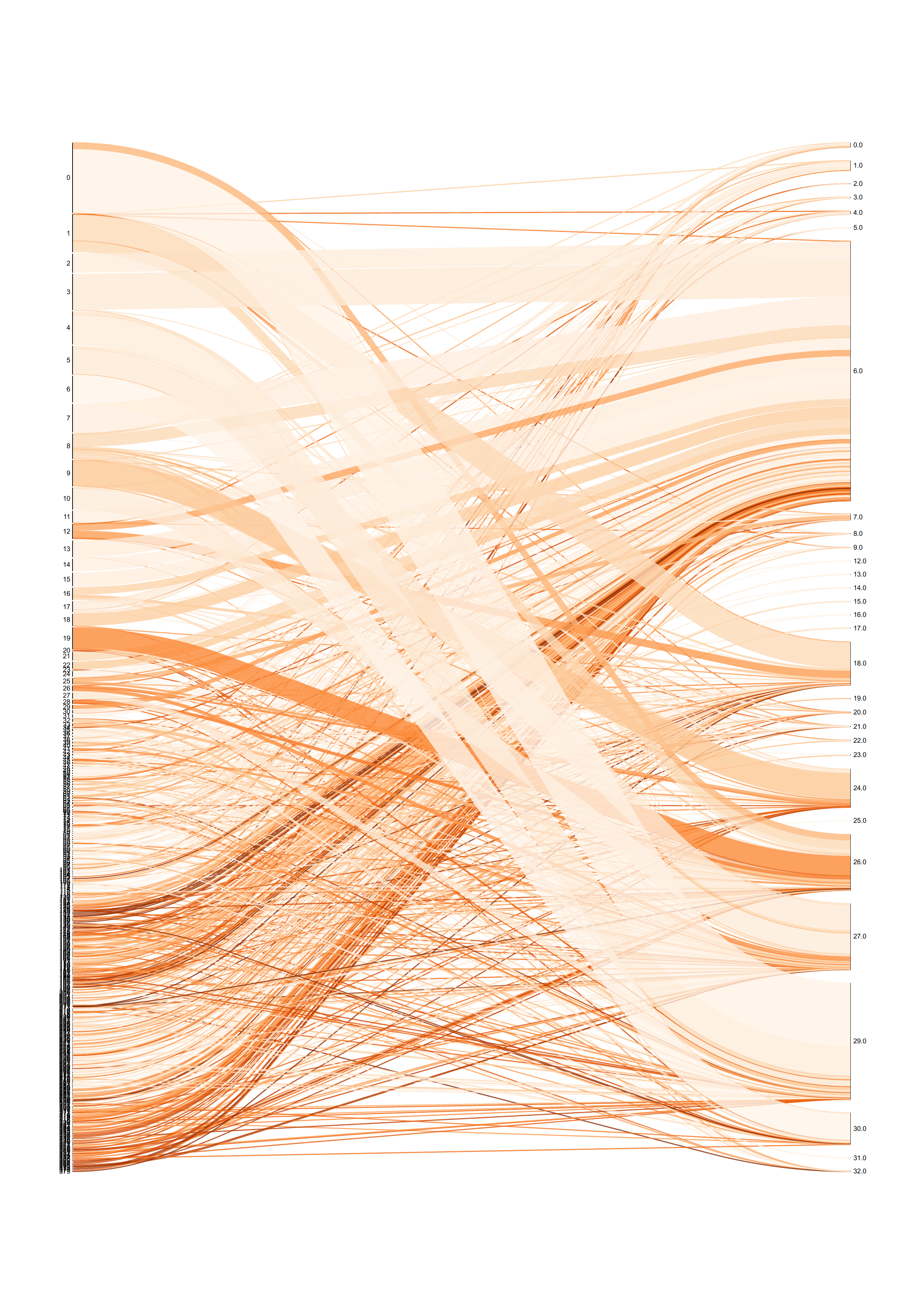}
    \caption{Sankey plot (Python library floweaver, version 2.0.0, \cite{lupton2017}) illustrating the distribution of geographic sampling points across the presented cluster set (left) and the Ecological Marine Units (EMUs, right, \cite{sayre2017}). The width of each flow corresponds to the number of shared points between clusters and EMUs. Flow colour indicate the Native Emergent Manifold Interrogation (NEMI) uncertainty of this study's clusters. The plot shows good agreement for several cluster-EMU combinations including 0-29, 1-18, 6-30 and 9-24. The largest difference is EMU 6 which aggregates many clusters of various sizes.}
    \label{s:fig:sankey_emus}
\end{figure}
\newpage

\section{Cluster Dataset}
The final produced clustering (100 UMAP-DBSCAN runs combined into one cluster set using NEMI) is publicly available online (cluster\_set.csv). Cluster label eight represents noise as assigned by DBSCAN. The provided file has 18 columns: 

\begin{enumerate}
    \item LEV\_M: Upper grid cell depth in \SI{}{\meter}
    \item LATITUDE: Geographic latitude of the grid cell centre in decimal degrees
    \item LONGITUDE: Geographic longitude of the grid cell centre in decimal degrees
    \item P\_TEMPERATURE: Average potential temperature of the grid cell in \SI{}{\celsius}
    \item P\_SALINITY: Average salinity of the grid cell in psu
    \item P\_OXYGEN: Average oxygen concentration of the grid cell in \SI{}{\micro\mole\per\kilo\gram}
    \item P\_NITRATE: Average nitrate concentration of the grid cell in \SI{}{\micro\mole\per\kilo\gram}
    \item P\_SILICATE: Average silicate concentration of the grid cell in \SI{}{\micro\mole\per\kilo\gram}
    \item P\_PHOSPHATE: Average phosphate concentration of the grid cell in \SI{}{\micro\mole\per\kilo\gram}
    \item e0: First reduced feature from UMAP dimensionality reduction
    \item e1: Second reduced feature from UMAP dimensionality reduction
    \item e2: Third reduced feature from UMAP dimensionality reduction
    \item volume: Volume of the grid cell in \SI{}{\cubic\meter}
    \item label: Cluster label assigned to the grid cell
    \item uncertainty: NEMI uncertainty of the cluster label of the grid cell
    \item color: Hexadecimal color assigned to the cluster label of the grid cell
    \item water: Binary flag denoting if the grid cell is located on land or in water according to GEBCO bathymetry information \citep{gebco2022}
    \item imputed: Percentage of parameters that needed to be imputed for the grid cell, e.g. 50 denotes that \SI{50}{\percent} = 3 of the six parameters were imputed using KNN
\end{enumerate}

\bibliographystyle{cas-model2-names}

\bibliography{cas-refs}



\end{document}